\documentclass[sigplan,screen]{acmart}
\usepackage[]{hyperref}
\usepackage{amsmath}
\usepackage{subcaption}
\usepackage{graphicx}
\usepackage{xspace}
\usepackage{ifthen}
\usepackage{xcolor}
\usepackage{array}
\usepackage{cleveref}
\usepackage{xparse}
\usepackage{multirow}
\usepackage{booktabs}
\usepackage{arydshln}
\usepackage{enumitem}
\usepackage{algorithm}
\usepackage{algpseudocode}
\usepackage{fancyvrb}
\usepackage{spverbatim}
\usepackage{verbatim}

\newcommand{\pa}{PagedAttention\xspace}
\newcommand{\vllm}{vLLM\xspace}
\newcommand{\flashinfer}{FlashInfer\xspace}
\newcommand{\flashattention}{FlashAttention-2\xspace}
\newcommand{\fapaged}{FA2\_Paged\xspace}
\newcommand{\fipaged}{FI\_Paged\xspace}
\newcommand{\sysname}{vAttention\xspace}
\newcommand{\favattention}{FA2\_vAttention\xspace}
\newcommand{\fivattention}{FI\_vAttention\xspace}

\newcommand{\yismall}{Yi-6B\xspace}
\newcommand{\llamasmall}{Llama-3-8B\xspace}
\newcommand{\yimedium}{Yi-34B\xspace}

\newcommand{\cudamalloc}{\texttt{cudaMalloc}\xspace}
\newcommand{\cumemreserve}{\texttt{cuMemAddressReserve}\xspace}
\newcommand{\cumemcreate}{\texttt{cuMemCreate}\xspace}
\newcommand{\cumemmap}{\texttt{cuMemMap}\xspace}
\newcommand{\cumemsetaccess}{\texttt{cuMemSetAccess}\xspace}
\newcommand{\cumemunmap}{\texttt{cuMemUnmap}\xspace}
\newcommand{\cumemrelease}{\texttt{cuMemRelease}\xspace}
\newcommand{\cumemfree}{\texttt{cuMemAddressFree}\xspace}
\newcommand{\cudamallocmanaged}{\texttt{cudaMallocManaged}\xspace}

\newcommand{\vmemreserve}{\texttt{vMemReserve}\xspace}
\newcommand{\vmemcreate}{\texttt{vMemCreate}\xspace}
\newcommand{\vmemmap}{\texttt{vMemMap}\xspace}
\newcommand{\vmemrelease}{\texttt{vMemRelease}\xspace}
\newcommand{\vmemfree}{\texttt{vMemFree}\xspace}

\newcommand{\torchempty}{\texttt{torch.empty}\xspace}

\newcommand{\vattninit}{\texttt{init}\xspace}
\newcommand{\vattnallocid}{\texttt{alloc\_reqid}\xspace}
\newcommand{\vattnstep}{\texttt{step}\xspace}
\newcommand{\vattnfreeid}{\texttt{free\_reqid}\xspace}

\newcommand{\kvcache}{KV cache\xspace}
\newcommand{\kcache}{K cache\xspace}
\newcommand{\vcache}{V cache\xspace}
\newcommand{\reqid}{\texttt{reqId}\xspace}

\newcommand{\arxivsummarization}{arXiv-Summarization\xspace}

\algrenewcommand\algorithmiccomment[1]{\hfill// \textnormal{#1}}

\makeatletter
\renewcommand{\sectionautorefname}{\S\@gobble}
\renewcommand{\subsectionautorefname}{\S\@gobble}
\renewcommand{\subsubsectionautorefname}{\S\@gobble}
\makeatother

\usepackage{ifthen}
\newboolean{publicversion}
\setboolean{publicversion}{true}

\ifthenelse{\boolean{publicversion}}{
	\newcommand{\grumbler}[3]{}
        \newcommand{\jm}[1]{}
        \newcommand{\ap}[1]{}
        \newcommand{\rp}[1]{}
}
{
\newcommand{\grumbler}[3]{\xspace\textcolor{#3}{\bf #1: #2}}

\newcommand{\jm}[1]{\grumbler{Jayashree}{#1}{magenta}}
\newcommand{\ap}[1]{\grumbler{Ashish}{#1}{violet}}
\newcommand{\rp}[1]{\grumbler{Ramya}{#1}{blue}}

}

\AtBeginDocument{%
  }

\copyrightyear{2025}
\acmYear{2025}
\setcopyright{rightsretained}
\acmConference[ASPLOS '25]{Proceedings of the 30th ACM International Conference on Architectural Support for Programming Languages and Operating Systems, Volume 1}{March 30--April 3, 2025}{Rotterdam, Netherlands}
\acmBooktitle{Proceedings of the 30th ACM International Conference on Architectural Support for Programming Languages and Operating Systems, Volume 1 (ASPLOS '25), March 30--April 3, 2025, Rotterdam, Netherlands}
\acmDOI{10.1145/3669940.3707256}
\acmISBN{979-8-4007-0698-1/25/03}

\makeatletter

\gdef\@copyrightpermission{
  \begin{minipage}{0.95\columnwidth}
   \href{https://creativecommons.org/licenses/by/4.0/}{This work is licensed under a Creative Commons Attribution International 4.0 License.}
  \end{minipage}
  \vspace{12pt}
}

\makeatother
\begin{document}

\title[vAttention: Dynamic Memory Management for Serving LLMs]{\sysname: Dynamic Memory Management for \\ Serving LLMs without PagedAttention}

\author{Ramya Prabhu}
\affiliation{%
  \institution{Microsoft Research}
  \city{Bengaluru}
  \country{India}
}

\author{Ajay Nayak}
\authornote{Contributed to this work as an intern at Microsoft Research India.}
\affiliation{%
  \institution{Indian Institute of Science}
  \city{Bengaluru}
  \country{India}
}

\author{Jayashree Mohan}
\affiliation{%
  \institution{Microsoft Research}
  \city{Bengaluru}
  \country{India}
}

\author{Ramchandran Ramjee}
\affiliation{%
  \institution{Microsoft Research}
  \city{Bengaluru}
  \country{India}
}

\author{Ashish Panwar}
\affiliation{%
  \institution{Microsoft Research}
  \city{Bengaluru}
  \country{India}
}

\renewcommand{\shortauthors}{Ramya Prabhu, Ajay Nayak, Jayashree Mohan, Ramchandran Ramjee, \& Ashish Panwar}

\begin{abstract}

PagedAttention is a popular approach for dynamic memory allocation in LLM serving systems. It enables on-demand allocation of GPU memory to mitigate \kvcache fragmentation --- a phenomenon that crippled the batch size (and consequently throughput) in prior systems. However, in trying to allocate \textbf{physical} memory at runtime, \pa ends up changing the \textbf{virtual} memory layout of the \kvcache from contiguous to non-contiguous. Such a design leads to non-trivial programming and performance overheads.

We present \sysname{} --- an approach that mitigates fragmentation in physical memory while retaining the virtual memory contiguity of the \kvcache. We achieve this by decoupling the allocation of virtual and physical memory using CUDA virtual memory management APIs. We also introduce various LLM-specific optimizations to address the limitations of CUDA virtual memory support. Overall, \sysname is a simpler, portable, and performant alternative to \pa: it supports various attention kernels out-of-the-box and improves LLM serving throughput by up to $1.23\times$ compared to the use of \pa-based kernels of \flashattention and \flashinfer.

\end{abstract}




\maketitle
\section{Introduction}
\label{sec:introduction}

\begin{table}[t!]
\scalebox{0.95}{
\begin{tabular}{p{8.5cm}}
\textbf{System/library and issues related to \pa} \\ \toprule

\textbf{\vllm~\cite{vllmsosp}:} Pioneered \pa. Despite being in an actively maintained code repository, \vllm's \pa kernel is up to $2.8\times$\ slower than \flashattention (\autoref{table:eval:decode-attn}). Furthermore, changing block size changes the execution time of the kernel by as much as $1.9\times$ (\autoref{fig:eval:vllm:blocksize}). \\ \hline

\textbf{FlashAttention-2~\cite{flashattention2}:} PagedAttention-based prefill kernel is up to $37\%$ slower than the non-paged kernel (\autoref{fig:motivation:paged-vs-nonpaged}) while the decode kernel is up to $12\%$ slower. Initial attempts to add paging support failed unit tests~\cite{fa-paged-crash}. \\ \hline

\textbf{FlashAttention-3~\cite{flash-attention-3} / SDPA in cuDNN-9~\cite{cudnn-9}:} State-of-the-art attention kernels for NVIDIA Hopper architecture did not support \pa when released. \\ \hline

\textbf{TensorRT-LLM~\cite{trtllmgithub}:} Serving throughput dropped by more than $10\%$ in a Python front-end~\cite{tensorrt-perf-decay}. Recommends using the C++ front-end. Even with C++, we observe up to $5\%$ higher latency in some cases with \pa. \\ \hline

\textbf{\flashinfer~\cite{flashinfer}:} PagedAttention-based prefill kernel is up to $42\%$ slower than the non-paged kernel (\autoref{fig:motivation:paged-vs-nonpaged}). \\ \bottomrule
\end{tabular}}
\caption{The \pa approach requires an application to explicitly manage dynamically allocated physical memory, including re-writing of attention kernels. These examples highlight the complexity, performance and maintenance challenges associated with this approach.}
\label{table:intro:banner}
\end{table}

Large Language Models (LLMs) are being deployed in a wide range of applications e.g., chat bots, search engines and coding assistants~\cite{openai2022gpt4techreport,chowdhery2022Palm,bingai,bard,githubcopilot,amazoncodewhisperer,replitghostwriter}. Given the size and scale of modern LLM deployments, optimizing inference has become extremely important ~\cite{sarathi2023,splitfuse2024,patel2023splitwise,distserve2024,tetriinfer,vllmsosp,orca,pod-attn-2024}. 

Batching is a powerful technique to boost LLM serving throughput~\cite{orca,vllmsosp,patel2023splitwise,sarathiserve2024}. However, achieving a large batch size requires careful allocation of GPU memory. For each request, the serving framework stores the activations of all the tokens processed so far in GPU memory and reuses them for generating subsequent tokens. This is called the \kvcache~\cite{orca,sarathi2023,patel2023splitwise} which accounts for a majority of GPU memory usage during inference. Efficiently allocating GPU memory for the \kvcache is challenging for two reasons. First, the per-request \kvcache grows slowly (one token per iteration), and second, a request's decode length (or its total \kvcache size) is not known ahead of time.

Prior systems like Orca~\cite{orca} and FasterTransformer~\cite{fastertransformer} allocate memory for each request based on the maximum context length supported by the model (e.g., \yimedium model supports context length of up to 200K~\cite{yi-34b-200k-hf}). However, the number of decode tokens generated are far less in practice, e.g., the average decode length for the chat-based sharegpt dataset is 415 tokens~\cite{sarathiserve2024}. Therefore, static memory allocation could create severe internal fragmentation, limiting batch size and serving throughput.

Inspired by demand paging in OS-based virtual memory systems, \vllm introduced \pa~\cite{vllmsosp} that allocates small blocks of GPU memory on demand i.e., when previously allocated blocks are fully utilized and the model continues to generate more tokens. This approach provides a near-perfect solution for mitigating fragmentation and hence, \pa has become the de facto standard for dynamic memory allocation in LLM serving systems, e.g., TensorRT-LLM, HuggingFace TGI,  LightLLM~\cite{trtllmgithub, hftgi, lightllm:github} etc.

However, we show that \pa faces a fundamental consequence of dynamic memory allocation: \textit{dynamically allocated objects are not guaranteed to be contiguous}. Note that user-level objects are allocated in virtual memory. Therefore, in trying to enable dynamic allocation of \textbf{physical} memory, \pa ends up changing the \textbf{virtual} memory layout of \kvcache from contiguous to non-contiguous. We argue that this approach has several pitfalls (\autoref{sec:motivation}). First, it requires rewriting attention kernels, i.e., to enable de-referencing all tokens of the non-contiguous \kvcache. Second, it forces developers to implement a memory manager in the serving framework, i.e., to stitch together dynamically allocated virtual memory blocks. Third, it adds runtime overhead in the critical path of both CPU and GPU execution. \autoref{table:intro:banner} provides empirical evidence and real-world experiences to support these arguments.


The fundamental issue with \pa and prior systems is that they rely on the reservation-based memory allocation method exposed by the GPU runtime. In this method (used by \cudamalloc), the runtime allocates both virtual and physical memory on the GPU meaning that \textit{physical memory is allocated even if the corresponding virtual memory is not accessed}. This is in stark contrast to OS-based demand paging~\cite{ingens,hawkeye}. We show that separating the allocation of virtual and physical memory allows for more effective \kvcache memory management. To support our claim, we introduce \sysname (\autoref{sec:design}) --- an approach that stores \kvcache in contiguous virtual memory without committing physical memory ahead-of-time. \sysname decouples the allocation of virtual and physical memory using the CUDA virtual memory management (VMM) APIs~\cite{cudavirtualmemory}.


In building \sysname, we find that using CUDA VMM support for \kvcache management poses two key efficiency challenges for an LLM serving system (\autoref{sec:design:optimizations}). First, memory allocation using CUDA VMM APIs incurs high latency because each allocation involves a round-trip to the OS kernel. We tackle latency issues with several LLM-specific optimizations such as overlapping memory allocation with compute, opportunistically allocating pages ahead of time, and deferring memory reclamation. Second, CUDA supports memory allocation only at the granularity of large pages, i.e., in multiples of 2MB. Use of large pages can create significant fragmentation. We address this challenge by modifying the open-source CUDA unified virtual memory driver, adding support for smaller 64KB pages. Our evaluation shows that use of 64KB pages has no negative impact on the performance of attention kernels, i.e., we do not find any evidence of TLB thrashing. Together, these optimizations mitigate fragmentation while hiding the latency cost of on demand memory allocation, making  \sysname a simpler, portable and performant alternative to \pa.

Overall, we make the following contributions:
\begin{itemize}
\item We present \sysname{} -- a memory management approach that retains the virtual contiguity of \kvcache while enabling dynamic allocation of physical memory. Our implementation of \sysname in \vllm seamlessly adds dynamic memory allocation support to various unmodified attention kernels.


\item We compare \sysname against \pa-based alternatives of \vllm, \flashattention and \flashinfer on \yismall, \llamasmall and \yimedium with 1-2 A100 GPUs. Using \flashattention's non-paged attention kernel, \sysname outperforms \vllm by up to $1.99\times$ in decode throughput. In long-context scenarios, it also improves the end-to-end LLM serving throughput by up to $1.18\times$ and $1.23\times$ over \pa based kernels of \flashattention and \flashinfer 

\item We demonstrate the portability benefit of \sysname with the recently launched FlashAttention-3 kernel (FA3~\cite{flash-attention-3}). FA3 is optimized for the NVIDIA Hopper architecture and was not released with \pa support. \sysname supports FA3 out-of-the-box, leading to $1.26-1.5\times$ higher throughout over \pa based \flashattention.

\end{itemize}

\section{Background}
\label{sec:background}

\subsection{Large Language Models}

Given an input sequence, an LLM predicts the probability of an output sequence: a sequence is a set of tokens~\cite{vllmsosp}. Each inference request begins with a prefill phase that processes prompt tokens in parallel and produces the first output token of the request. Thereafter, the decode phase iteratively processes the output token generated in the previous step and produces the next output token in every iteration~\cite{sarathi2023}. 

LLMs are built atop one of the variants of the transformer architecture~\cite{attentionpaper}; an LLM consists of multiple transformer blocks. Internally, a transformer block contains two types of operators: position-wise and sequence-wise~\cite{orca}. The former category includes feed-forward network, layer normalization, activation, embedding layer, output sampling layer, and residual connections  whereas \textit{attention} is a sequence-level operator. We primarily focus on attention since it is the primary consumer of GPU memory in LLM inference. \autoref{table:notation} summarizes the notations used in the paper.

For the attention operator, the model computes the query, key and value vectors from a given sequence of tokens
$(x_1, x_2,...., x_K)\in\mathbb{R}^{K\times E}$
where E represents the embedding size of the model. For each $x_i$, query, key and value vectors are computed as follows:

\begin{equation}
    q_{i} = W_{q}x_{i},\;\;\; k_{i} = W_{k}x_{i}, \;\;\; v_{i} = W_{v}x_{i} 
\end{equation}

The resulting $k_{i}$ and $v_{i}$ are appended to the key and value vectors of the prior tokens of the corresponding request, producing two matrices $K, V \in \mathbb R^{L'\times (H\times D)}$ where $L'$ represents the context length of the request seen so far, H is the number of KV heads of the model on a worker and D is the dimension of each KV head. Then, attention is computed as follows:

\begin{equation}
Attention(q_{i}, K, V) = softmax(\frac{q_{i}K^T}{scale})V    
\label{eq:attention}
\end{equation}

\begin{table}[t!]
\centering
\scalebox{0.9}{
\begin{tabular}{p{0.15\linewidth}p{0.75\linewidth}}
\toprule
\textbf{Symbol} & \textbf{Definition} \\ \midrule
\( N \) & Number of layers on a particular worker \\
\( H \) & Number of KV heads on a particular worker \\
\( D \) & Dimension of each attention head \\
\( P \) & Number of bytes needed to store one element  \\
\( B \) & Maximum batch size \\
\( L \) & Maximum context length supported by the model \\
\( L' \) & Context length of a request seen thus far \\
\bottomrule
\end{tabular}}
\caption{Notations used in the paper.}
\vspace{-1em}
\label{table:notation}
\end{table}

The attention score is computed separately for each request in the batch. A request executes until the model generates a special end-of-sequence token or reaches the maximum context length for the request. Note that in each iteration of a request, all its preceding $k_{i}$ and $v_{i}$ are needed to compute attention. Hence, an inference engine stores the $k_{i}$ and $v_{i}$ vectors in memory to reuse them across iterations. We refer to this state of all layers collectively as \kvcache. In systems prior to \pa, the \kcache (or \vcache) at each layer of a worker is typically allocated as a 4D tensor of shape $[B, L, H, D]$ where $B$ refers to batch size and $L$ refers to the maximum possible context length of a request.

\subsection{Fragmentation and PagedAttention}
Serving LLMs with high throughput requires careful allocation of GPU memory. This is challenging because the total context length of a request is not known in advance. Serving systems worked around this challenge by pre-reserving \kvcache space assuming that each context is as long as the maximum length supported by the model (e.g., 200K for \yimedium-200K). \vllm shows that this strategy is prone to severe internal fragmentation. In fact, \vllm showed that prior reservation is suboptimal even if the context lengths are known in advance. This is because the per-request \kvcache grows one token at a time and hence prior reservation wastes memory over the entire lifetime of a request. 

Inspired by the OS-based virtual memory systems, \vllm proposed \pa to mitigate fragmentation by dynamically allocating memory for the \kvcache. \pa splits \kvcache into fixed-sized blocks and allocates memory for one block at a time. This way, \vllm allocates only as much memory as a request needs, and only when required -- not ahead-of-time.

\noindent
\textbf{GPUs and page sizes:} NVIDIA GPUs support multiple page sizes in the hardware~\cite{mcmgpus,gpu-mismanaged,gpu-reverse-engg-tlb1,pascalmmu}. A single call to a CUDA VMM API can allocate one or more physical pages, which we refer to as a page-group. We use page-groups to support multiple allocation granularities for the \kvcache, similar to how multiple page sizes are commonly used in conventional OS-based virtual memory systems~\cite{ingens,hawkeye}.

\section{Issues with the \pa Approach}
\label{sec:motivation}

Despite being inspired by demand paging, the \pa approach is different from it: \textit{\pa implements demand paging in user space whereas conventional demand paging is transparent to applications}. This section elaborates on issues that arise with such an approach.

\subsection{Requires Re-writing the Attention Kernel}
\label{subsec:rewriting}

Conventional implementations of the attention operator assume that the two input tensors K and V (\autoref{eq:attention}) are stored in contiguous memory. By departing from the conventional memory layout, \pa requires an implementation of the attention operator to be modified so as to compute attention scores over non-contiguous \kvcache blocks. Writing correct and performant GPU kernels can be challenging for most programmers~\cite{fa-paged-crash}.

Being a fundamental building block of the transformer architecture, the attention operator has witnessed a tremendous pace of innovation in the systems and ML communities for performance optimizations~\cite{flashattention, flashattention2, flashdecoding, flashdecoding++, flashinfer, h2o2023, flashattnhopper, groupedqueryattention, multiqueryattention, generatinglongsequence2019, longformer, pod-attn-2024}, and this trend is likely to continue. In the \pa model, keeping up with new research requires continued efforts in porting new optimizations to a \pa-aware implementation. Production systems can therefore easily fall behind research, potentially losing performance and competitive advantage. To provide an example,~\autoref{table:eval:decode-attn} shows that the paged kernel of \vllm is already up to $2.8\times$ slower than the \flashattention kernel~\cite{groupedqueryattention}.

\begin{figure}[t!]
    \centering
    \includegraphics[width=0.95\columnwidth]{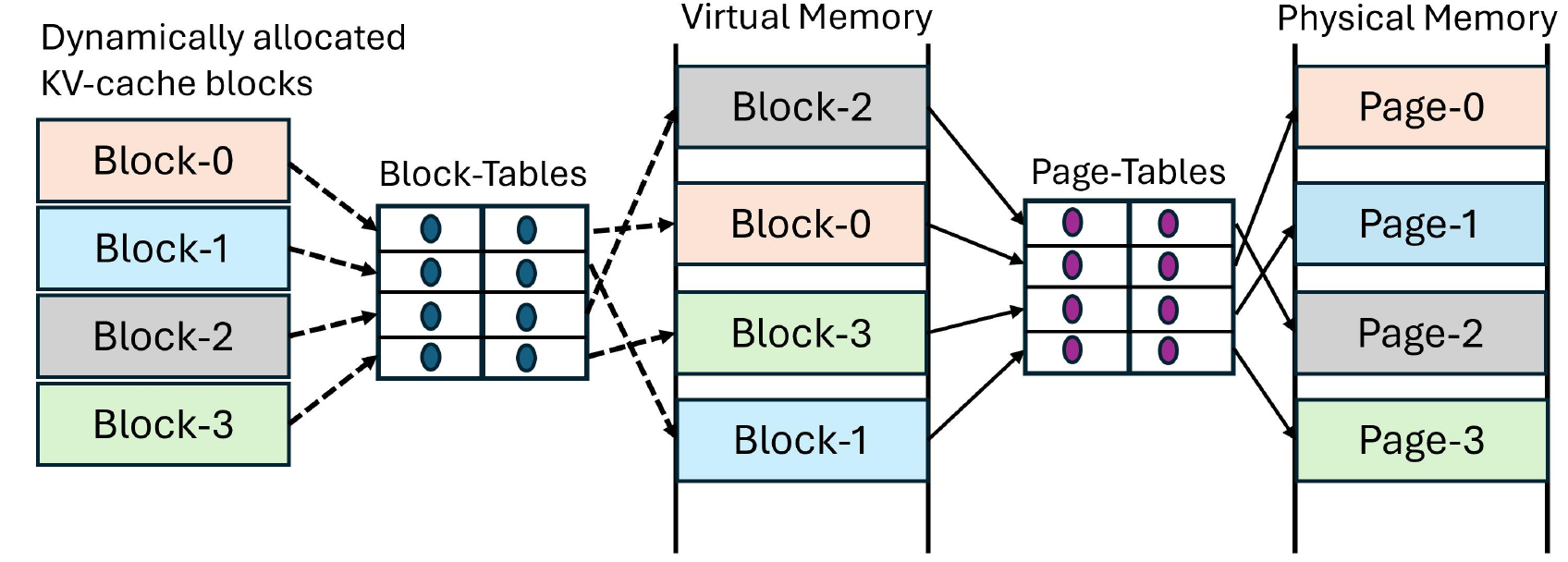}
    \caption{\pa involves two layers of memory management: one in user space and one in OS kernel space.}
    \label{fig:motivation:pamemlayout}
\end{figure}

\subsection{Adds Redundancy in the Serving Framework}
\label{subsec:blocktable}

\pa makes an LLM serving system responsible for managing the mappings between \kvcache and dynamically allocated memory blocks.
For example, consider a request that allocates four \kvcache blocks over time (left half of~\autoref{fig:motivation:pamemlayout}). These blocks are usually non-contiguous in virtual memory. During the computation of~\autoref{eq:attention}, \pa kernel needs to access all the elements of the four \kvcache blocks. To facilitate this, the serving system needs to track the virtual memory addresses of \kvcache blocks and pass them to the attention kernel at runtime. This approach effectively requires duplicating what the operating system already does for enabling virtual-to-physical address translation (right half in~\autoref{fig:motivation:pamemlayout}).

\subsection{Performance Overhead}
\label{sec:motivation:perfoverhead}

\subsubsection{Runtime overhead on the GPU}
\label{subsec:gpuperfoverhead}
\begin{figure}[t!]
    \centering
\includegraphics[width=0.9\columnwidth]{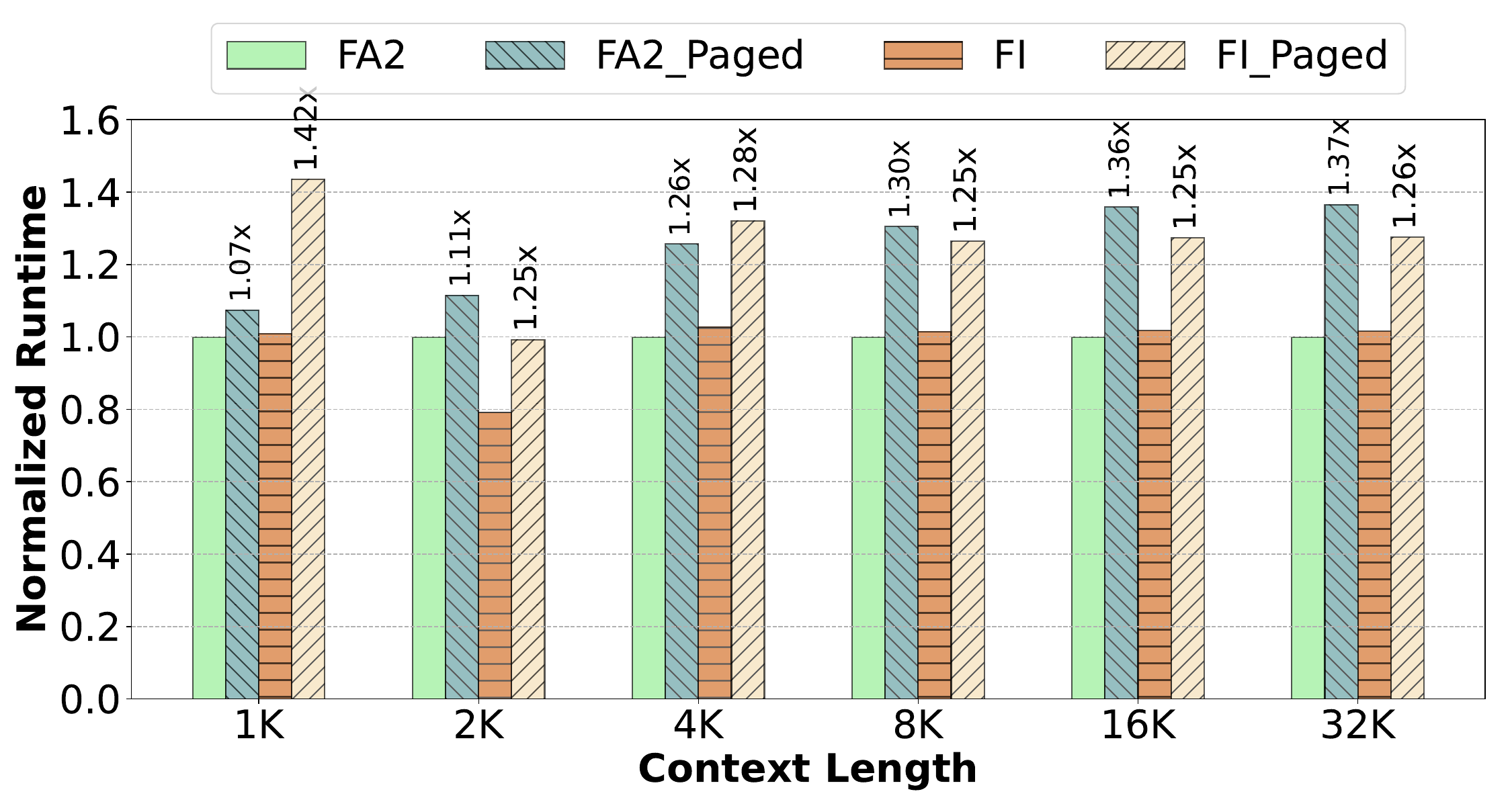}
    \caption{Overhead of \pa in prefill kernels (model: \llamasmall, one A100 GPU). Numbers on top show overhead over the corresponding non-paged implementation of \flashattention (FA2) and \flashinfer (FI).}
    \label{fig:motivation:paged-vs-nonpaged}
\end{figure}

\pa slows down attention computation by adding extra code in the critical path of execution. For example, the \vllm paper acknowledges that the \pa-based implementation was $20-26\%$ slower than the corresponding none-paged FasterTransformer kernel, primarily due to the overhead of looking up Block-Tables and executing extra branches (see Figure 18a in ~\cite{vllmsosp}). In addition, \autoref{fig:motivation:paged-vs-nonpaged} shows that incorporating \pa has also added a significant performance overhead in other state-of-the-art kernel libraries. For example, \pa based prefill kernels of \flashattention and \flashinfer are up to 37\% and 42\% slower than the non-paged kernels in the corresponding libraries. Our analysis reveals that the number of instructions executed in \pa kernels is $7-13\%$ higher than the non-paged kernels. Caching page indices also increases register pressure, causing register spilling.

\begin{figure}[t!]
    \centering
    \includegraphics[width=0.9\columnwidth]{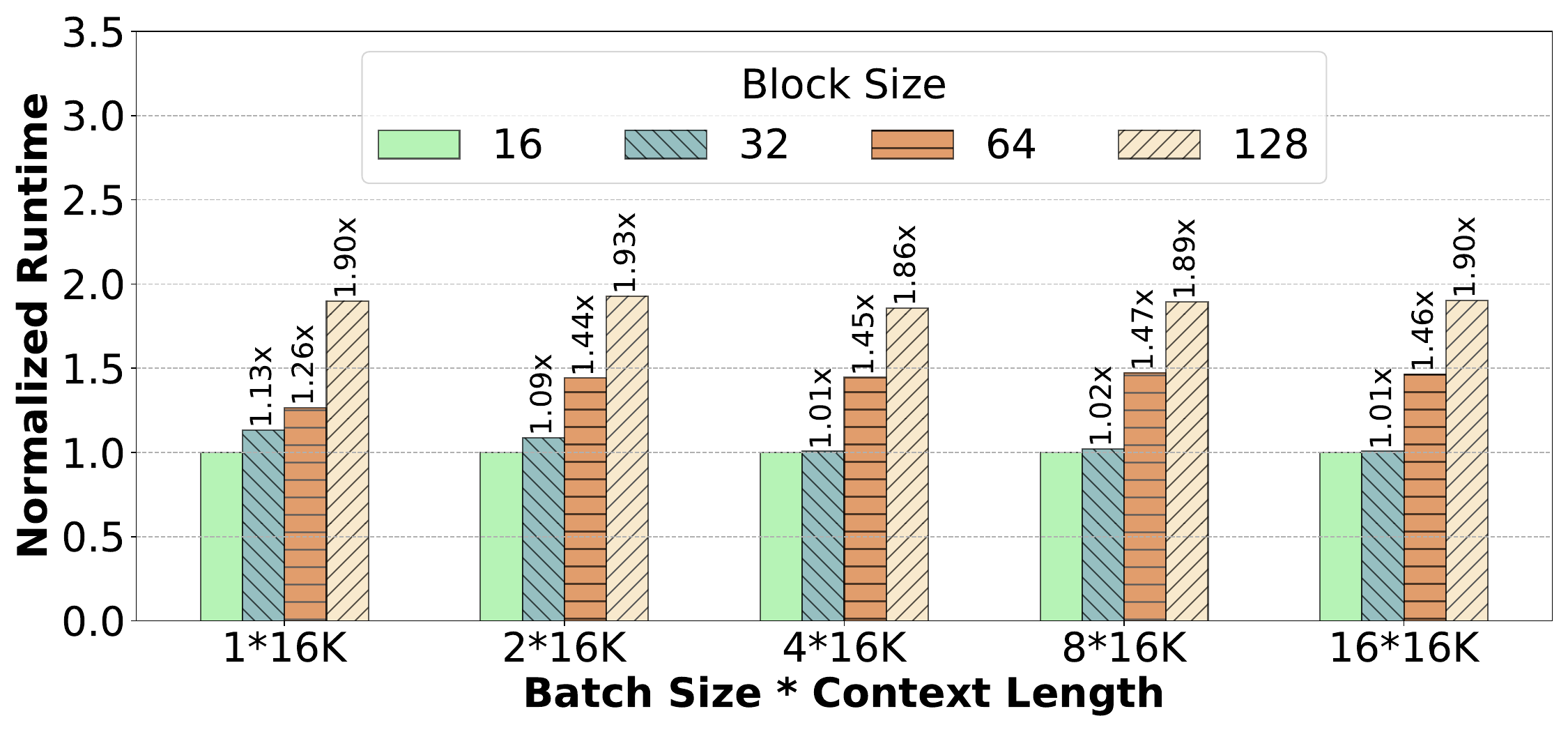}
    \caption{Latency of \vllm's paged decode kernel is sensitive to block size (model: \llamasmall, one A100 GPU).}
    \label{fig:eval:vllm:blocksize}
\end{figure}

To highlight another example of difficulty involved in writing an efficient paged kernel, \autoref{fig:eval:vllm:blocksize} shows that the performance of \vllm's paged decode kernel is significantly worse with large block sizes of 64 and 128. Our analysis indicates that this is likely due to L1 cache efficiency: smaller blocks have a higher memory bandwidth utilization due to higher hit rates in the L1 cache.

\subsubsection{Runtime overhead on the CPU}
\label{subsec:cpuperfoverhead}
Implementing an additional memory manager can add performance issues in the CPU runtime of the serving system. We refer to a few real-world examples and our own observations on \vllm to corroborate this argument.

To enable \pa, a serving system needs to supply Block-Tables to the attention kernel. In \vllm, the latency of preparing a Block-Table depends on batch composition and grows  proportional to \texttt{max\_num\_blocks} $\times$ \texttt{batch\_size} where \texttt{max\_num\_blocks} refers to the number of \kvcache blocks in the longest request of the batch. This is because \vllm manages a Block-Table as a 2D tensor and aligns the number of \kvcache blocks in each request by padding unoccupied slots with zeros. If a batch contains a few long and many short requests, such padding results in a significant overhead. In our earlier experiments, we observed that Block-Table preparation in \vllm was contributing 30\% latency in decode iterations. While a recent fix~\cite{vllm-blocktable-issue} has mitigated some of this overhead, we find that it can still be as high as $10\%$. High overhead of \pa has also been found in TensorRT-LLM, degrading throughput by 11\%~\cite{tensorrt-perf-decay}. This issue was attributed to the Python runtime of TensorRT-LLM and moving to a C++ runtime can mitigate the CPU overhead. However, doing so requires non-trivial programming effort.

Overall, this section shows that \pa adds a significant programming burden while also being inefficient. In \sysname, we propose a more principled approach to  dynamic \kvcache memory management. However, before delving into \sysname, we first highlight some of the fundamental characteristics of LLM serving workloads from a memory management perspective.

\section{Insights into LLM Serving Systems}
\label{sec:analysis}

To understand the memory allocation pattern of LLM serving systems, we experiment with  \yismall running on a single A100 GPU, and \llamasmall and \yimedium running on two A100 GPUs with tensor-parallelism (TP). We set the initial context length of each request to 1K tokens, vary the batch size from 1 to 320 and measure the throughput and memory requirement of the decode phase (see \autoref{sec:design:optimizations} for our discussion and optimizations for the prefill phase).

\noindent
\textbf{Observation-1:} \textit{\kvcache memory requirement is predictable on a per-iteration basis}. Due to auto-regressive decoding, once a request enters the decode phase, its \kvcache size increases uniformly by one token per iteration. This allows the serving system to determine in advance if additional memory will be required during the iteration's execution.\footnote{Note that this prediction is possible only at the granularity of individual iterations; the total \kvcache requirement of a request remains unknown as it depends on the total number of output tokens in the request.}

\begin{figure}[t]
    \centering
    \begin{subfigure}[b]{0.9\columnwidth}
    \centering
        \includegraphics[trim={20 25 0 0}, clip=True, width=\textwidth]{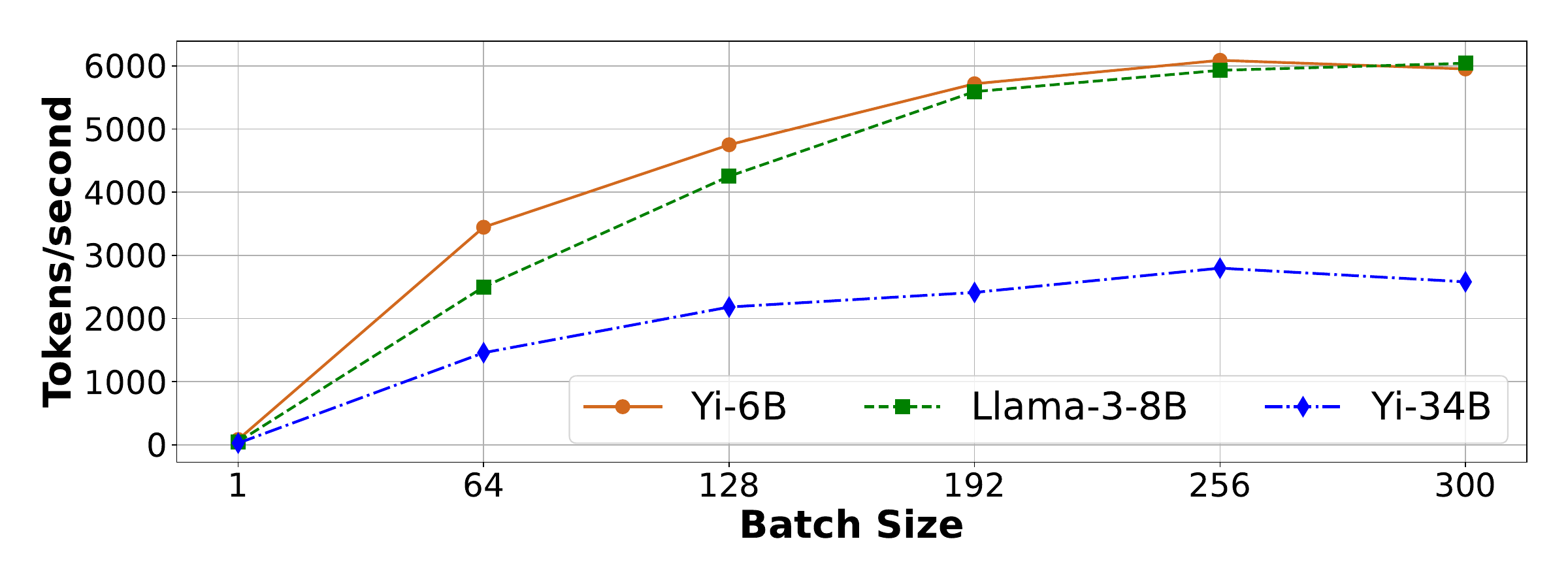}
        \caption{Decode throughput.}
        \label{fig:analysis:llm:memory:top}
    \end{subfigure}
    \\
    \begin{subfigure}[b]{0.9\columnwidth}
    \centering
        \includegraphics[trim={20 25 0 0}, clip=True, width=\textwidth]{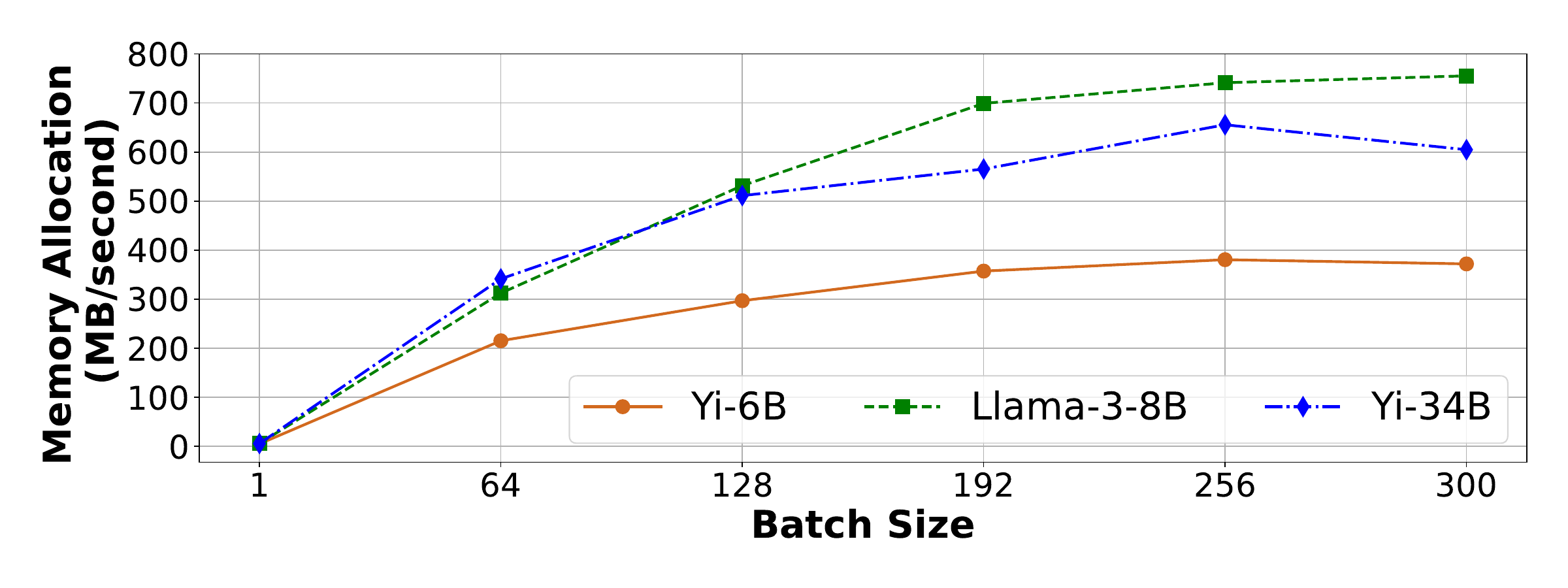}
        \caption{Rate of memory allocation.}
        \label{fig:analysis:llm:memory:bottom}
    \end{subfigure}
    \caption{Decode throughput (top) and the rate of physical memory allocation (bottom) saturate at large batch sizes.}
    \label{fig:analysis:llm:memory}
\end{figure}

\noindent
\textbf{Observation-2:} \textit{\kvcache does not require high memory allocation bandwidth.} The memory footprint of a single token across all layers is typically few 10s-100s of kilobytes of memory. For example, the per-token memory footprint of \yismall, \llamasmall and \yimedium is 64KB, 128KB and 240KB, respectively. Further, each iteration runs for 10s-100s of milliseconds implying that a request requires at most a few megabytes of memory per second. While batching improves system throughput~\cite{sarathi2023,sarathiserve2024,patel2023splitwise,distserve2024}, the number of tokens generated per second plateaus beyond a certain batch size (\autoref{fig:analysis:llm:memory:top}). This implies that the memory allocation bandwidth requirement also saturates at large batch sizes (e.g., at 256 for \yimedium). For all the models we studied, we observe that the highest memory allocation rate is at most 750MB per second (\autoref{fig:analysis:llm:memory:bottom}). \sysname leverages these observations to optimize \kvcache memory management.

\begin{figure*}[t!]
    \centering
    \includegraphics[trim={0 315 0 0}, clip=True, width=0.9\textwidth]{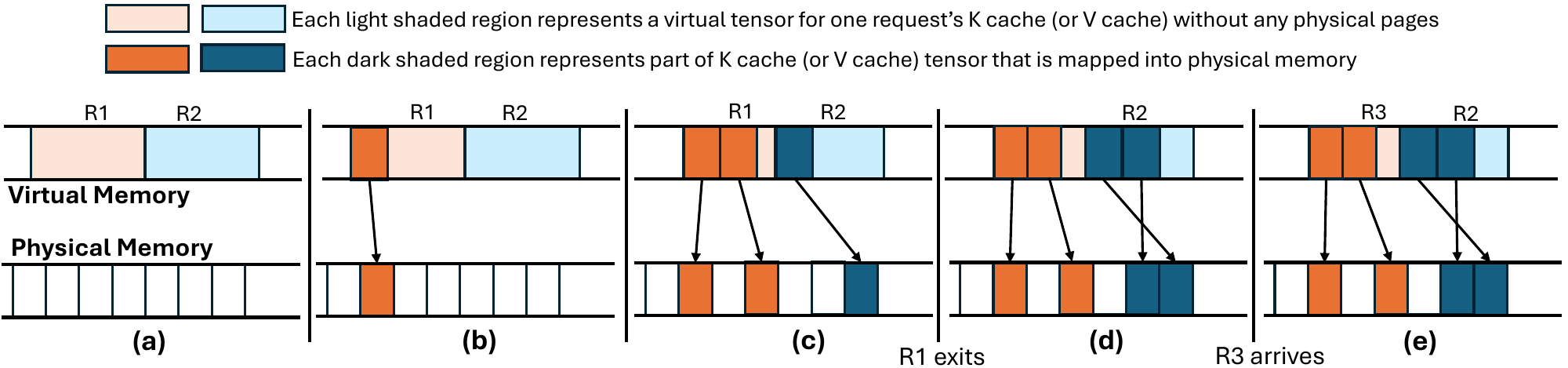}
    \caption{Dynamic memory management in \sysname for a single \kcache (or \vcache) tensor. (a) shows a virtual tensor for a batch of two requests with no physical memory allocation yet. (b) \texttt{R1} is allocated one physical page. (c) \texttt{R1} is allocated two pages and \texttt{R2} is allocated one page. (d) \texttt{R1} has completed but \sysname does not reclaim its memory (deferred reclamation). (e) when \texttt{R3} arrives, \sysname assigns \texttt{R1}'s tensor to it which is already backed by physical memory. }
    \label{fig:design:vattention}
\end{figure*}


\section{\sysname: Design and Implementation}
\label{sec:design}

Our primary observation is that \textit{physical memory fragmentation can be avoided without making \kvcache non-contiguous in virtual memory.} To realize this, \sysname decouples the allocation of virtual memory from physical memory by leveraging system support for demand paging (instead of implementing demand paging in user space, as in \pa). 


\subsection{Design Overview}
\label{sec:design:overview}

\sysname employs distinct allocation policies for virtual and physical memory. Specifically, we allocate a large contiguous buffer for the \kvcache in virtual memory ahead-of-time (similar to systems prior to \pa) while deferring the allocation of physical memory to runtime (similar to \pa). This design preserves virtual contiguity of \kvcache without fragmenting physical memory. Note that this approach could fragment and waste virtual memory. However, this is not an issue since virtual memory is abundant, e.g., modern 64-bit systems provide a 128TB user-addressable virtual memory per  process.\footnote{64-bit systems use only 48 bits for virtual addresses today, providing a per-process virtual memory space of 256TB which is divided equally between the user space and (OS) kernel space.}

\subsubsection{Pre-reserving virtual memory}
Since virtual memory is abundant, we pre-allocate it in size that is large enough to hold the \kvcache of the maximum batch size (configurable) that needs to be supported. In doing so, we assume that each request's context length is same as the maximum  supported by the model.

\subsubsection{Number of virtual memory buffers}
A serving framework maintains separate K and V tensors for each layer of the model. Therefore, we reserve $2\times N$ buffers on a worker where $N$ is the number of layers managed by that worker. 


\begin{table*}[t!]
    \centering
    \scalebox{0.85}{
    \begin{tabular}{l|l|l|c|c|c|c}
    & & & \multicolumn{4}{c}{\textbf{Latency (microseconds)}} \\ 
       \textbf{CUDA VM APIs} & \textbf{\sysname VM APIs} & \textbf{Description} & \textbf{64KB} & \textbf{128KB} & \textbf{256KB} & \textbf{2MB} \\ \toprule
      \cumemreserve*  & \vmemreserve*    & Allocate a buffer in virtual memory          & 18 & 17 & 16 & 2     \\
      \cumemcreate*   & \vmemcreate*     & Allocate a handle in physical memory         & 1.7 & 2  & 2.1 & 29      \\
      \cumemmap       & \vmemmap         & Map a physical handle to a virtual buffer      & 8 & 8.5 & 9 & 2      \\
      \cumemsetaccess & -                    & Enable access to a virtual buffer         & - & - & - & 38       \\ \hdashline 
      \cumemunmap     & -                    & Unmap physical handle from a virtual buffer  & - & - & - & 34       \\
      \cumemrelease*  & \vmemrelease*    & Free physical pages of a handle             & 2 & 3 & 4 & 23       \\
      \cumemfree*     & \vmemfree*       & Free a virtual memory buffer                 & 35 & 35 & 35 & 1        \\
      \bottomrule
    \end{tabular}}
    \caption{CUDA VMM APIs. * represents APIs that we use once while instantiating or terminating the serving framework. Rest of the APIs are used for (un)mapping physical memory pages at runtime. CUDA APIs (prefixed with cu) support only 2MB pages, whereas our CUDA extension APIs (prefixed with v) support fine-grained allocations.}
    \label{tab:design:cudaapis}
\end{table*}

\subsubsection{Size of a virtual memory buffer}
\label{sec:design:buffersize}
The maximum size of a buffer is $BS=B\times S$ where B is the maximum batch size and $S$ is the maximum size of a single request's per-layer \kcache (or \vcache) on a worker. Further, $S=L\times H\times D\times P$, where $L$ is the maximum context length supported by the model, $H$ is the number of KV heads on a worker, $D$ is the dimension of each KV head and $P$ is the number of bytes based on model precision (e.g., P=2 for FP16/BF16).  As an example, consider \yimedium with FP16 and two-way tensor-parallelism (TP-2). In this case, $N\!=\!60, H=4, D=128, P=2$ (8 KV heads of \yimedium are split evenly on two GPUs), and maximum supported context length $L=200K$. For this configuration, $S=200MB$ $(200K*4*128*2)$. Assuming $B=500$, the maximum size of each buffer per-worker is $BS=100GB$ ($500\times 200MB$). Therefore, the total virtual memory requirement for 60 layers is 120 buffers of 100GB each (12TB total). Note that the size of virtual address space available grows with the number of workers, e.g., with two TP workers, the total size of user-addressable virtual address space is 256TB. Therefore, virtual memory is always plentiful to satisfy large allocations. \autoref{fig:design:vattention} shows an example of how \sysname allocates physical memory pages dynamically.

\subsection{Leveraging CUDA Virtual Memory Support}

The standard GPU memory allocation interface \cudamalloc does not support demand paging, i.e., it allocates virtual memory and physical memory at the same time. However, recent CUDA versions provide programmers a fine-grained control over managing virtual and physical memory, including support for decoupling their allocations~\cite{cudavirtualmemory, gmlakeasplos24}. We leverage these low-level APIs.

\subsubsection{CUDA virtual memory APIs}~\autoref{tab:design:cudaapis} provides an overview of CUDA VMM APIs that allow decoupling the allocation of virtual memory from physical memory. The allocation granularity depends on the page size used by the GPU. Further, the size of a virtual memory buffer or a physical memory handle must be a multiple of the physical memory allocation granularity.  Physical memory pages can be allocated to (or de-allocated from) sub-regions in a virtual memory buffer independently of other sub-regions. 

\begin{figure}
    \centering
    \includegraphics[trim={250 225 250 110}, clip=true, width=0.8\columnwidth]{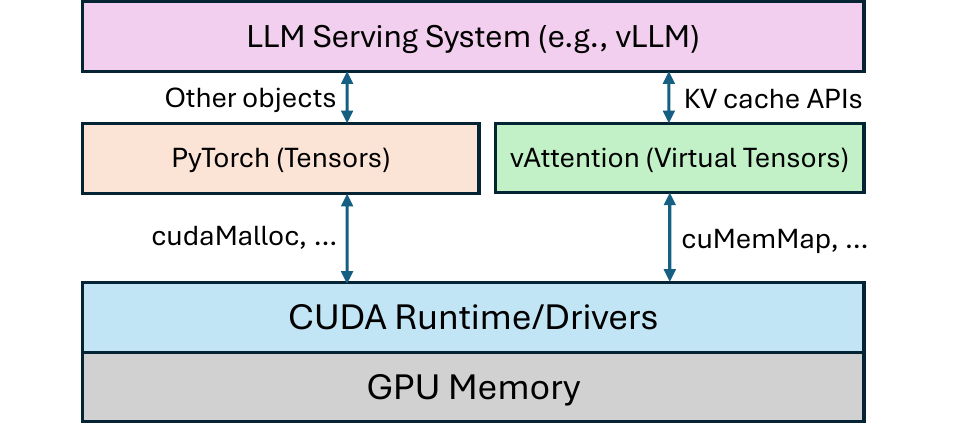}
    \caption{An LLM serving system interacts with \sysname for \kvcache management with a set of simple APIs listed in~\autoref{tab:design:vattn-apis}. All other memory objects (e.g., activations) are allocated by the standard PyTorch caching allocator.}
    \label{fig:design:vattn-usage}
\end{figure}

\subsubsection{Extending PyTorch caching allocator} \kvcache is a collection of tensors. In current deep learning frameworks such as PyTorch, a tensor allocated via APIs such as \torchempty comes with pre-allocated physical memory. This is because the PyTorch caching allocator relies on the \cudamalloc interface (\autoref{fig:design:vattn-usage}). Relying on the low-level API support from CUDA, we extend the PyTorch caching allocator to allow an application to reserve a virtual memory buffer for a tensor without committing physical memory ahead-of-time. We refer to tensors allocated via these APIs as virtual tensors.

\subsubsection{Request-level \kvcache indexing} A virtual tensor represents the \kcache (or \vcache) of a layer for the maximum batch size B. In these tensors, different requests occupy different non-overlapping sub-regions (say sub-tensors). We locate the sub-tensor of a request with a unique integer identifier \reqid that lies in the range of $0$ to $B-1$ (note that at most $B$ requests run simultaneously). The \kcache (or \vcache) offset of a request's sub-tensor in the virtual tensor of the entire batch is $\reqid\times S$ where $S$ is the maximum size of per-layer \kcache (or \vcache) of a request on a worker. The request identifier \reqid is allocated by \sysname.

\begin{table}[t!]
    \centering
    \scalebox{0.85}{
    \begin{tabular}{r|p{7cm}}
       \textbf{APIs}  & \textbf{Description} \\ \toprule
      \vattninit  & Initializes \sysname  with model parameters. \\ 
      & arguments: $N, B, L, H, D, P$, \texttt{page-group-size}. \\
      & return value: a list of \kvcache tensors.   \\ \hline
      \vattnallocid & Allocates an unused \reqid and marks it active \\
      & arguments: None \\
      & return value: an integer \reqid \\ \hline
      \vattnfreeid &  Frees a \reqid and marks it inactive \\
      & arguments: an integer \reqid \\
      & return value: None \\ \hline
      \vattnstep & Ensures \kvcache tensors are backed by physical pages up to the current context length of each active request  \\
      & arguments: an array of size B containing sequence lengths of each \reqid \\
      & return value: 0 (success), -1 (failure).\\
       \bottomrule
    \end{tabular}}
    \caption{Key APIs that \sysname exposes to a serving framework for dynamic \kvcache memory management.}
    \label{tab:design:vattn-apis}
\end{table}

\subsection{Serving LLMs with \sysname}
\label{sec:design:serving-llms}

We build \sysname as a Python library that internally uses a CUDA/C++ extension for interacting with CUDA drivers. Our library exposes a set of simple APIs to the serving framework (shown in~\autoref{fig:design:vattn-usage},~\autoref{tab:design:vattn-apis} and Algorithm~\autoref{alg:design:integration}). For simplicity, we discuss the use of these APIs from a single worker's perspective; all workers behave the same.

\subsubsection{Initial setup} When the serving framework starts, each model worker loads the \sysname library and configures it with model parameters $N, H, D, P$, $B$ and a preferred {\it page-group size} (\autoref{sec:opt:frag}) via the \vattninit API (line 4 in Algorithm~\autoref{alg:design:integration}). Internally,  \sysname reserves $2\times N$ virtual tensors on the worker, as shown in \autoref{fig:design:vattention}(a),  where $N$ is the number of layers hosted by the worker. These virtual tensors are reserved for the lifetime of the serving application. In addition, \sysname also pre-allocates physical memory pages at each worker during initialization. However, these pages are not mapped into the \kvcache at this point.

\begin{algorithm}[t]
\caption{Using \sysname in a serving framework.}
\label{alg:design:integration}
\begin{algorithmic}[1]
\State \textnormal{max\_batch\_size $\gets$ B}
\State \textnormal{cache\_seq\_len $\gets$ [0]*B}
\State \textnormal{req\_batch\_idx $\gets$ dict()}
\State \textnormal{\underline{vattention.init(config\_params)}}
\While{\textnormal{!request\_pool.is\_empty()}}
    \For{\textnormal{$R_i$ in new\_requests}}
        \If{\textnormal{can\_schedule($R_i$)}}
            \State \textnormal{idx $\gets$ \underline{vattention.alloc\_reqid()}}
            \State \textnormal{req\_batch\_idx[$R_i$] $\gets$ idx}
            \State \textnormal{cache\_seq\_len[idx] $\gets$ prompt\_len($R_i$)}
        \EndIf
    \EndFor
    \State \textnormal{\underline{vattention.step(cache\_seq\_len)}}
    \State \textnormal{model.forward()}
    \For{\textnormal{$R_i$ in active\_requests}}
        \State \textnormal{idx $\gets$ req\_batch\_idx[$R_i$]}
        \If{\textnormal{is\_complete($R_i$)}}
            \State \textnormal{cache\_seq\_len[idx] $\gets$ 0}
            \State \textnormal{\underline{vattention.free\_reqid(idx)}}
        \Else
            \State \textnormal{cache\_seq\_len[idx] $+=$ 1}
        \EndIf
    \EndFor
\EndWhile
\end{algorithmic}
\end{algorithm}

\subsubsection{Scheduling a new request} When a new request is scheduled for the first time, the serving framework obtains a new \reqid from \sysname via \vattnallocid (line 8). All subsequent memory management operations of the request are tagged with this \reqid.

\subsubsection{Model execution} Before scheduling a batch for execution, the framework needs to ensure that the \kvcache sub-tensors of each active request are backed by physical memory (\autoref{fig:design:vattention}(b) and (c)). For this purpose, before dispatching the first kernel of an iteration to the GPU, the framework invokes the \vattnstep API (line 13), specifying the current context length of each request (context length is set to 0 for each inactive \reqid). Internally, \sysname ensures that enough physical pages are mapped for each active \reqid before returning execution back to the framework. If \sysname cannot satisfy the memory demand, it returns with a failure in response to which a serving framework can preempt one or more requests to allow forward progress (this is similar to \vllm's default behavior). We leave more sophisticated policies such as swapping out \kvcache to CPU memory as future work.

Depending on whether a request is in the prefill phase or decode phase, different amount of physical memory may need to be mapped for a given iteration. The prefill phase processes the input tokens of a given prompt in parallel. Therefore, the amount of physical memory needed to be mapped depends on the number of prompt tokens being scheduled. If the total \kcache size of all prompt tokens at one layer of the model is $s$ and page-group size is $t$, then each worker needs to ensure that at least $(s+t-1)/t$ page-groups are mapped in each of the $2\times N$ \kvcache sub-tensors of the given \reqid.

For a request in the decode phase, the number of new page-groups required is at most one per virtual tensor. This is because each iteration produces only one output token for a request. \sysname internally tracks the number of page-groups mapped for each request and maps new page-groups only when prior page-groups are about to be exhausted.

\subsubsection{Request completion} A request terminates when it reaches user specified or the maximum context length supported by the model, or when the model produces a special end-of-sequence token. The framework notifies \sysname  of a request's completion with \vattnfreeid (line 19). Internally, \sysname may unmap the physical pages of a completed request or defer them to be freed later (\autoref{sec:opt:defer}).

\noindent
\textbf{Supporting continuous batching:} Continuous batching poses one challenge in computing attention in our design. When a request from somewhere in the middle of a batch exits, it creates an unused hole in the virtual tensors of \kvcache. This layout is not supported by implementations that expect the query (Q) and \kvcache to be of the same size in batch ($B$) dimension. However, FlashAttention provides rich API support to address this issue; its argument \texttt{cache\_batch\_idx} allows Q and \kvcache to have different batch sizes and also be arranged in arbitrary order (i.e., batch index 0 in Q can be mapped to batch index 1 in \kvcache). \sysname benefits from this API support in terms of both performance and ease of programming; when the request composition of a batch changes, we update \texttt{cache\_batch\_idx} of running requests such that their Q tensors map to their respective \kvcache based on their \reqid.

\section{Optimizations}
\label{sec:design:optimizations}

There are two challenges in using CUDA virtual memory support for serving LLMs. First, invoking CUDA VMM APIs at runtime incurs high latency. Second, \cumemcreate currently allocates memory only at the granularity of large pages, i.e., multiples of 2MB. Use of large pages can waste physical memory due to internal fragmentation.  This section details a set of simple-yet-effective optimizations that we introduce to overcome these challenges.

\subsection{Hiding Latency of Memory Allocation}
The serving framework invokes the \vattnstep API in every iteration. The latency of \vattnstep depends on the number of page-groups that need to be mapped into \kvcache. Consider, for example, that the \kvcache of one request needs to be extended for \yimedium which has 60 layers. This requires 120 calls to \cumemmap + \cumemsetaccess each of which takes about 40 microseconds. Therefore, growing the \kvcache of one request by new page-groups (two per layer) adds about 5 millisecond latency to the corresponding iteration. The latency overhead grows proportional to the number of requests that need new page-groups in a given iteration. We propose the following optimizations to hide this latency:

\subsubsection{Overlapping memory allocation with compute (decode phase)}
We leverage the predictability of memory demand to overlap memory allocation with computation. In particular, note that each iteration produces a single output token for every decode request. Therefore, memory demand for a decode iteration is known ahead-of-time. Further, in the decode phase, a request requires at most one new page-group for each of its virtual tensors.  \sysname keeps track of the current context length and how much physical memory is already mapped for each request. Using this information, it determines when a request would need more memory and uses a background thread to allocate new page-groups when the preceding iteration is executing. For example, consider that a request \texttt{R1} would require more physical memory in iteration \texttt{i}. When the serving framework invokes \vattnstep API in iteration \texttt{i-1}, \sysname launches a background thread that maps page-groups for iteration \texttt{i}. Since per-iteration latency is typically in the range of 10s-100s of milliseconds, the background thread has enough time to prepare physical memory mappings for an iteration before it starts executing. This way, \sysname hides the latency of CUDA APIs by performing memory allocations out of the critical path. 

\subsubsection{Deferred reclamation + eager allocation (prefill phase)}
\label{sec:opt:defer}
We observe that allocating physical memory for the prefill phase can be avoided in many cases. Consider that a request \texttt{R1} completed in iteration \texttt{i} and a new request \texttt{R2} joins the running batch in iteration \texttt{i+1}. To avoid allocating page-groups to \texttt{R2} from scratch, \sysname simply defers the reclamation of \texttt{R1}'s page-groups (\autoref{fig:design:vattention}(d)) and assigns \texttt{R1}'s \reqid to \texttt{R2}. This way, \texttt{R2} uses the same tensors for its \kvcache that \texttt{R1} was using -- which are already backed by physical pages (\autoref{fig:design:vattention}(e)). Therefore, new allocations are required only if \texttt{R2}'s context length is higher than \texttt{R1}.

We further optimize the prefill phase by proactively allocating a small number of page-groups ahead of time. For this purpose, we try to keep a certain number of page-groups mapped into the virtual tensors of one of the inactive \reqid. When a new request arrives, we allocate this \reqid. At the same time, we identify a new \reqid to be allocated next and eagerly map physical page-groups for it. In most cases, these eager allocations obviate the need to allocate physical memory in the critical path of prefill execution. Finally, we trigger memory reclamation only when the number of page-groups cached in \sysname falls below a certain threshold (e.g., less than 10\% of GPU memory). We delegate both deferred reclamation and eager allocation to the background thread that the \vattnstep API spawns.

\begin{figure*}[t!]
    \centering
    \begin{subfigure}{0.33\textwidth}
    \includegraphics[trim={0 25 0 0}, clip=True, width=0.99\columnwidth]{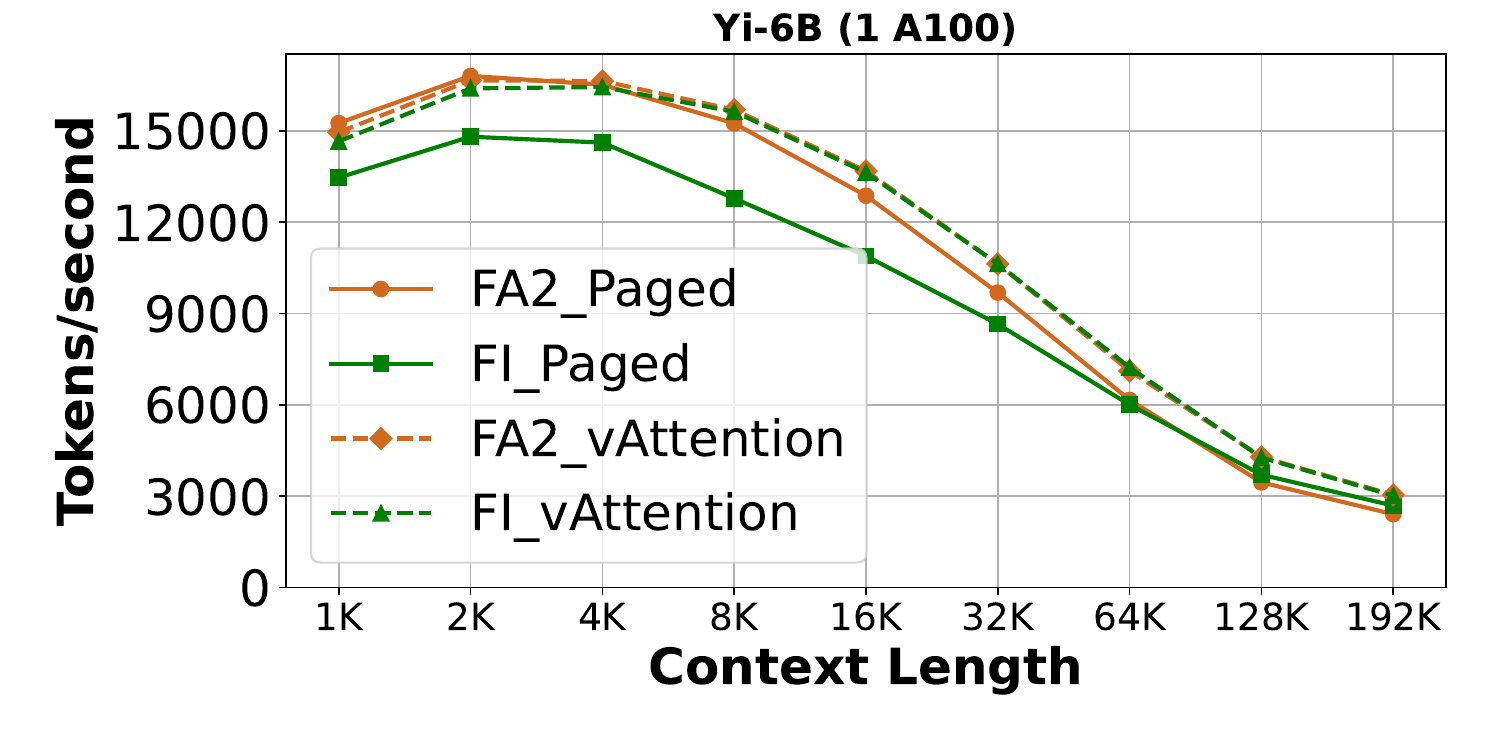} 
    \end{subfigure}
    \begin{subfigure}{0.33\textwidth}
    \includegraphics[trim={0 25 0 0}, clip=True, width=0.99\columnwidth]{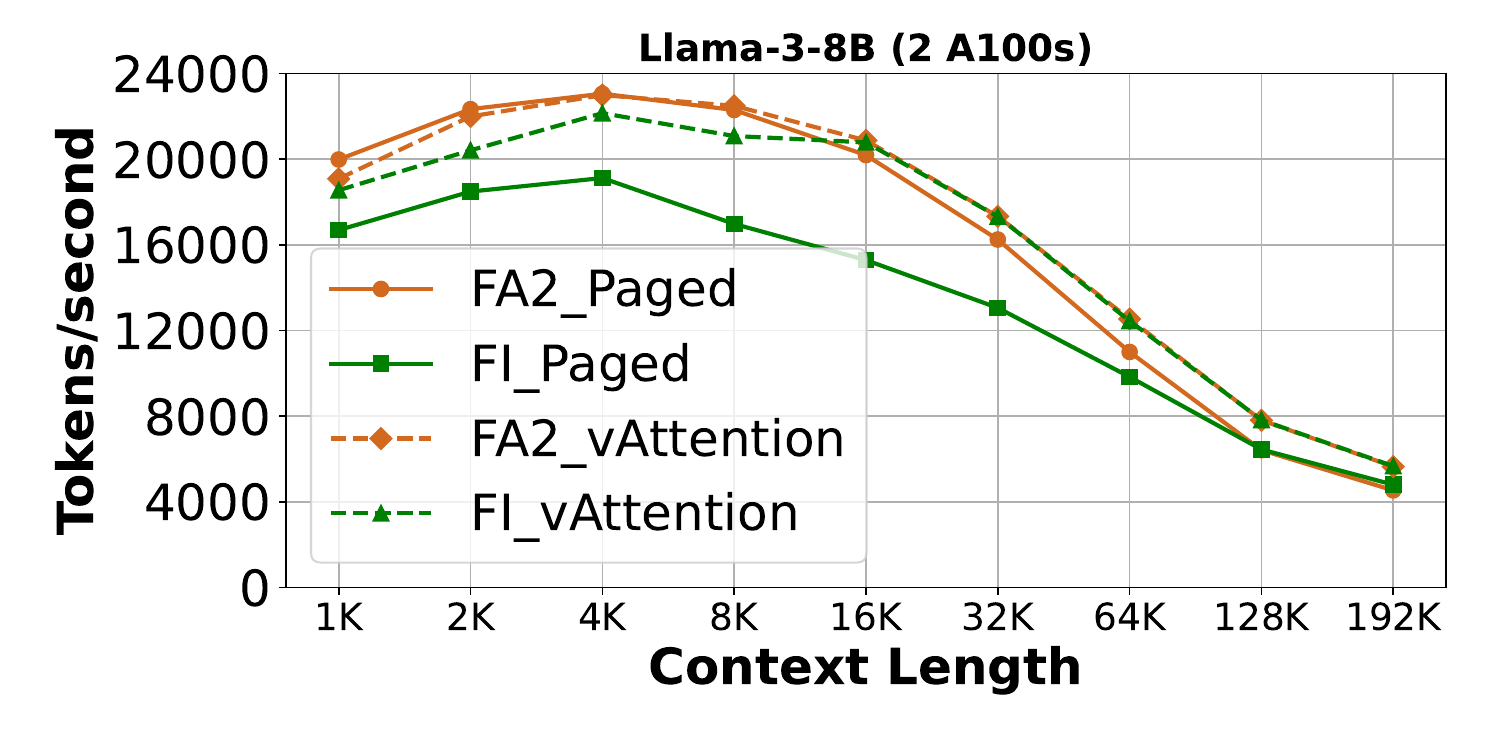}
    \end{subfigure}
    \begin{subfigure}{0.33\textwidth}
    \includegraphics[trim={0 25 0 0}, clip=True, width=0.99\columnwidth]{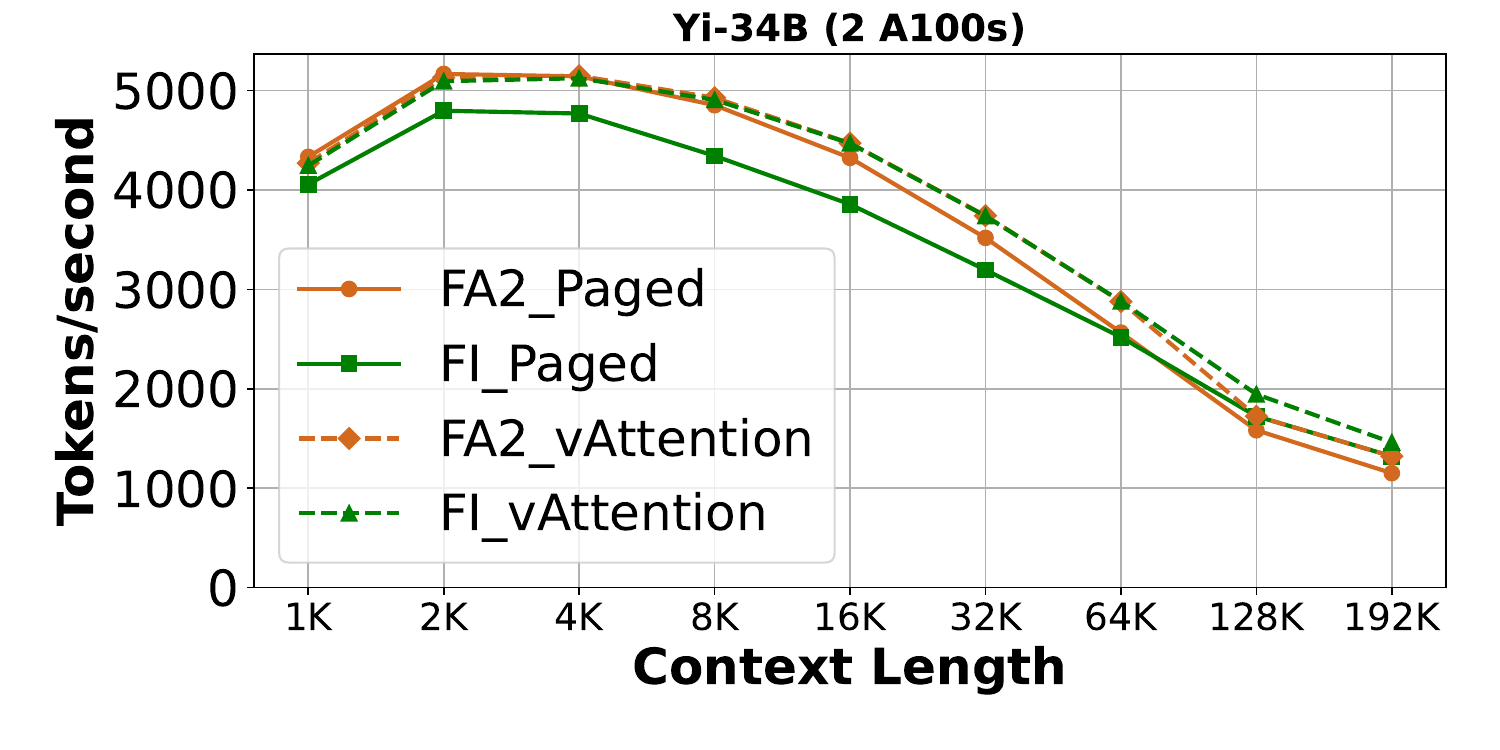}
    \end{subfigure}
    \caption{Prefill throughput. \sysname backed systems outperform the paged counterparts of both \flashattention and \flashinfer. Throughput for longer contexts is lower due to the quadratic complexity of prefill attention.}
    \label{fig:eval:prefill-throughput}
\end{figure*}

\subsection{Mitigating Internal Fragmentation}
\label{sec:opt:frag}

We mitigate internal fragmentation  by reducing the granularity of physical memory allocation. NVIDIA GPUs natively support at least three page sizes: 4KB, 64KB and 2MB~\cite{mcmgpus,gpu-mismanaged,gpu-reverse-engg-tlb1,pascalmmu}. Therefore, in principal, physical memory can be allocated in any multiple of 4KB sizes. The simplest way to achieve this would be to extend the existing CUDA VMM APIs (listed in \autoref{tab:design:cudaapis}) to also support allocating smaller pages (similar to how \texttt{mmap} in Linux supports multiple page sizes~\cite{ingens, hawkeye}). Unfortunately, the CUDA VMM APIs are implemented in the closed-source NVIDIA drivers which makes it impossible for us to modify their implementation. 

Fortunately, some part of NVIDIA drivers (particularly related to unified memory management) is open-source. Therefore, we implement a new set of APIs in the open-source NVIDIA drivers to mimic the same functionality that existing CUDA APIs provide but with support for multiple page sizes. The second column in \autoref{tab:design:cudaapis} shows our new APIs: most of our APIs have a one-to-one relationship with existing CUDA APIs except for \vmemmap that combines the functionality of \cumemmap and \cumemsetaccess, and \vmemrelease that combines the functionality of \cumemunmap and \cumemrelease for simplicity. In contrast to CUDA VMM APIs, our APIs allocate physical memory in 64KB, 128KB and 256KB sized page-groups. A serving framework can configure a desired page-group size in \sysname while initializing it (we use the standard 2MB pages if the configured page-group size is 2MB). \autoref{tab:design:cudaapis} shows the latency of each API with different page-group sizes.
\section{Evaluation}
\label{sec:evaluation}

Our evaluation answers the following questions:
\begin{itemize}
    \item How does \sysname impact the performance of attention kernels and the overall performance of prefill and decode phases (\autoref{sec:eval:prefill}, \autoref{sec:eval:decode})?
    \item How does \sysname impact the end-to-end LLM serving throughput (\autoref{sec:eval:makespan}, \autoref{sec:eval:online}, \autoref{sec:eval:fa3})?
    \item What is the effect of each of our optimizations (\autoref{sec:eval:ablation})?
\end{itemize}

\noindent
\textbf{Models and hardware:} We evaluate three models \yismall~\cite{yi-6b-200k-hf},
\llamasmall~\cite{llama-8b-hf} and \yimedium~\cite{yi-34b-200k-hf}. We conduct most of our evaluation on a single NVIDIA A100 GPU for \yismall, and two NVLink-connected A100 GPUs for \llamasmall and \yimedium (see~\autoref{tab:eval:models}). Each GPU has 80GB physical memory. We use tensor-parallelism degree of two (TP-2) for both \llamasmall and \yimedium. To demonstrate the portability benefit of \sysname, we use 1--2 H100 GPUs.

\noindent
\textbf{Evaluation methodology:} The computation and memory allocation pattern of the prefill and decode phases is substantially different~\cite{sarathiserve2024,splitfuse2024,distserve2024}. Attention kernels used for these two phases are also different and hence we evaluate them separately. The prefill phase requires one time memory allocation potentially spanning multiple pages. In comparison, the decode phase requires incremental memory allocation over the lifetime of a request~\cite{vllmsosp}. We define prefill throughput as the number of prompt tokens processed per second, and decode throughput as the number of tokens generated per second.

\begin{table}[t!]
    \centering
    \scalebox{0.85}{
    \begin{tabular}{l|c|c|c|c}
     Model & Hardware & $\#$ Q Heads & $\#$ KV Heads & $\#$ Layers \\ \toprule
    \yismall & 1 A100 & 32 & 4 & 32 \\
    \llamasmall & 2 A100s & 32 & 8 & 32 \\
    \yimedium & 2 A100s & 56 & 8 & 60 \\ \bottomrule
    \end{tabular}}
    \caption{Models and hardware used for evaluation.}
    \label{tab:eval:models}
\end{table}

\noindent
\textbf{Serving framework:} For a fair comparison, we use \vllm v0.2.7 as a common serving framework in all our experiments. We integrated state-of-the-art kernel libraries of both \flashattention v2.5.9~\cite{flashattention,flashattention2,fagithub} and \flashinfer v0.4.0~\cite{figithub} as attention back-ends into \vllm, and further added support for dynamic memory allocation via \sysname to their non-paged kernels. While \flashattention and \flashinfer are both based on the same underlying techniques (e.g., FlashDecoding~\cite{flashdecoding}), they use different Block-Table formats; the former use a simple lookup table whereas the latter uses a compressed Block-Table to optimize lookups.

\noindent
\textbf{Baselines:}  We compare performance obtained by using the non-paged attention kernels (backed by \sysname for dynamic memory allocation), and their paged counterparts. In addition, we also compare against \vllm's decode kernel (note that \vllm does not have a paged prefill kernel). We also profiled each system to find its best performing configuration. Accordingly, we set \kvcache block size to 16 for both \vllm and \flashinfer, and 256 for \flashattention. Using a higher block size for \vllm increases its kernel latency by up to $1.9\times$ as shown in~\autoref{fig:eval:vllm:blocksize}, and using a smaller block size for \flashattention paged kernel increases its latency by up to $9\%$. We find that the choice of page size does not affect \sysname as long as fragmentation is not a concern.

\begin{table}[t!]
\scalebox{0.75}{
\begin{tabular}{l|l|rr|ll}
\multirow{2}{*}{Model} & \multirow{2}{*}{\begin{tabular}[c]{@{}l@{}}Context\\ Length\end{tabular}} & \multicolumn{2}{c|}{FlashAttention-2} & \multicolumn{2}{c}{FlashInfer} \\
 &  & Paged & vAttention & Paged & vAttention \\ \toprule
\multirow{3}{*}{\yismall} & 64K & 10.6 (7.0) & 9.1 (5.5)  & 10.9 (6.0) & 9.1 (5.4) \\ 
 & 128K & 37.9 (30.3) & 30.5 (23.1) & 35.4 (25.4) & 30.7 (23.3) \\ 
 & 192K & 81.5 (70.0) & 64.6 (53.6) & 73.0 (58.3) & 65.1 (53.6)  \\ \hdashline
Llama-3 & 64K & 6.0 (3.4) & 5.2 (2.7) & 6.7 (3.0) & 5.3 (2.8) \\
-8B & 128K & 20.4 (15.4)  & 16.8 (11.6) & 20.3 (12.8) & 16.8 (11.4) \\
& 192K & 43.3 (35.6) & 34.8 (26.9) & 40.9 (29.7) & 34.7 (26.7) \\ \hdashline
 & 64K & 25.5 (13.2) & 22.8 (10.3) & 26.0 (11.2) & 22.7 (10.1) \\
\yimedium & 128K & 82.8 (56.9) & 68.4 (43.2) & 76.0 (46.7)  & 67.4 (42.5) \\
& 192K & 170.7 (131.8) & 136.9 (98.8) & 148.8 (104.7) & 134.6 (96.5) \\ \bottomrule
\end{tabular}}
\caption{Prefill completion and attention (in parenthesis) time with different attention back-ends (unit: seconds).}
\label{table:eval:ttft}
\end{table}

\subsection{Prefill Evaluation}
\label{sec:eval:prefill}

We evaluate 4 configurations for prefill: \fapaged, \fipaged, \favattention and \fivattention. Configurations with the ``\_Paged'' suffix represent the \pa-based kernel and those with ``\_\sysname'' use the non-paged kernels of the respective library. \autoref{fig:eval:prefill-throughput} shows the prefill throughput for our models. We summarize  key findings below.

\noindent
\textbf{Small contexts:}  For small contexts, prefill cost is dominated by the linear operators, i.e., attention's contribution is relatively low~\cite{sarathi2023}. Hence, even though \sysname helps speed up attention computation, the throughput of both paged and \sysname back-ends is nearly identical in \flashattention. However, for \flashinfer, we find that using \sysname helps improve prefill throughput even for small contexts. This is because \flashinfer incurs various other sources of overhead in the paged version. First, appending a new K or V tensor to the \kvcache requires a single tensor copy operation in \sysname, whereas in a paged implementation, it requires appending one block at a time (the copy operation has been optimized for \flashattention by \vllm~\cite{vllm-copy-kernels}). Second, \flashinfer involves creation and deletion of a few objects for its compressed Block-Tables in every iteration. \sysname avoids such overheads because it maintains \kvcache's virtual contiguity, eliminating the need for a Block-Table. 

\begin{figure*}[t!]
    \centering
    \begin{subfigure}{0.33\textwidth}
    \includegraphics[trim={0 25 0 0}, clip=True, width=0.99\columnwidth]{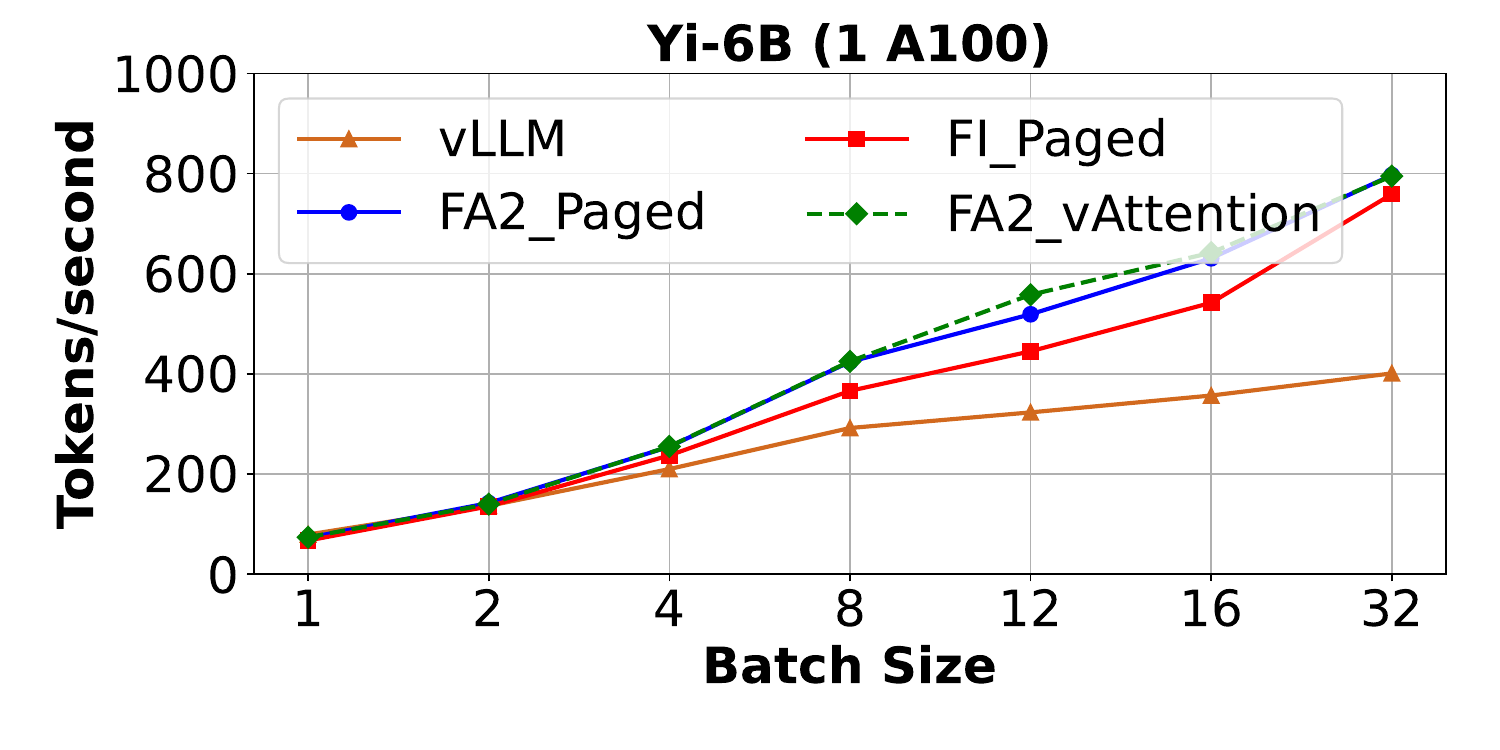} 
    \end{subfigure}
    \begin{subfigure}{0.33\textwidth}
    \includegraphics[trim={0 25 0 0}, clip=True, width=0.99\columnwidth]{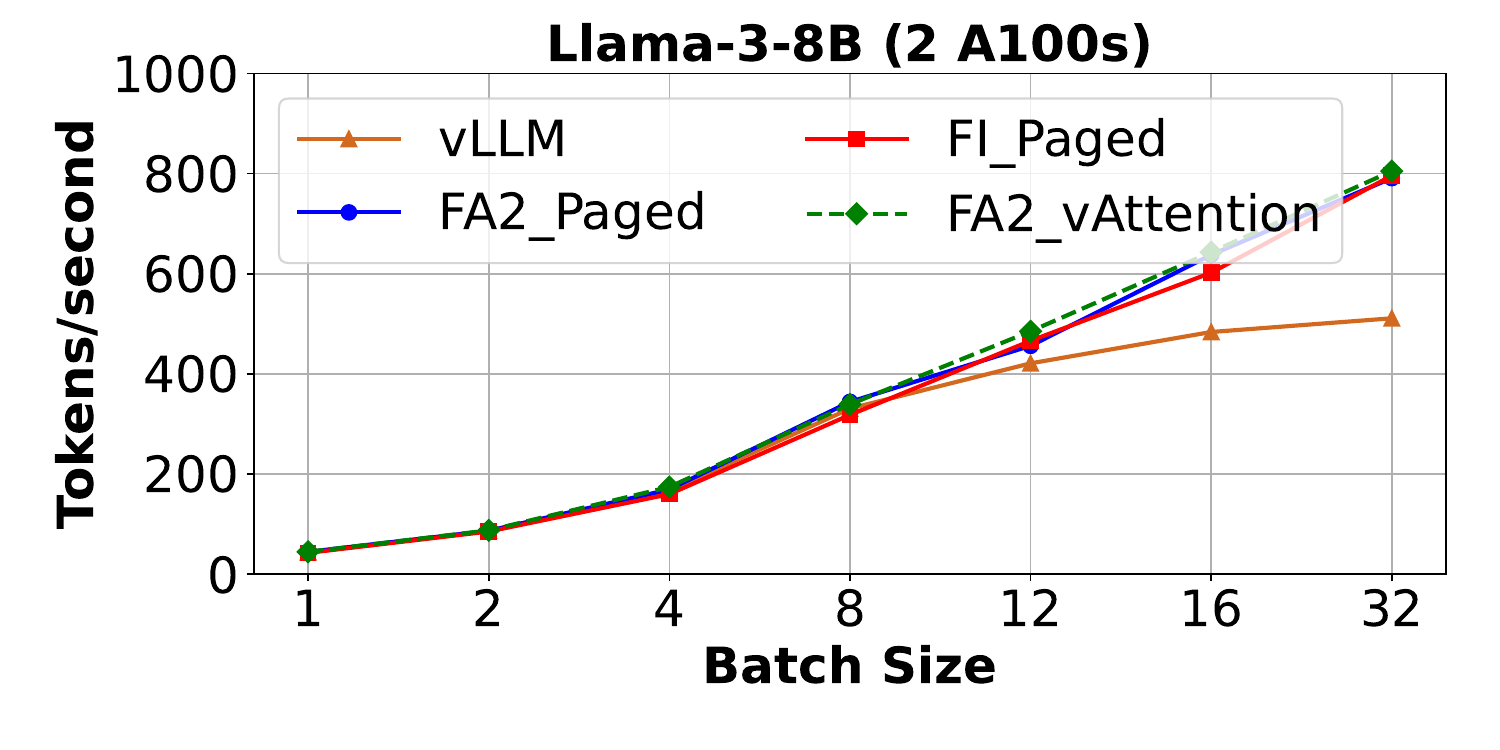}
    \end{subfigure}
    \begin{subfigure}{0.33\textwidth}
    \includegraphics[trim={0 25 0 0}, clip=True, width=0.99\columnwidth]{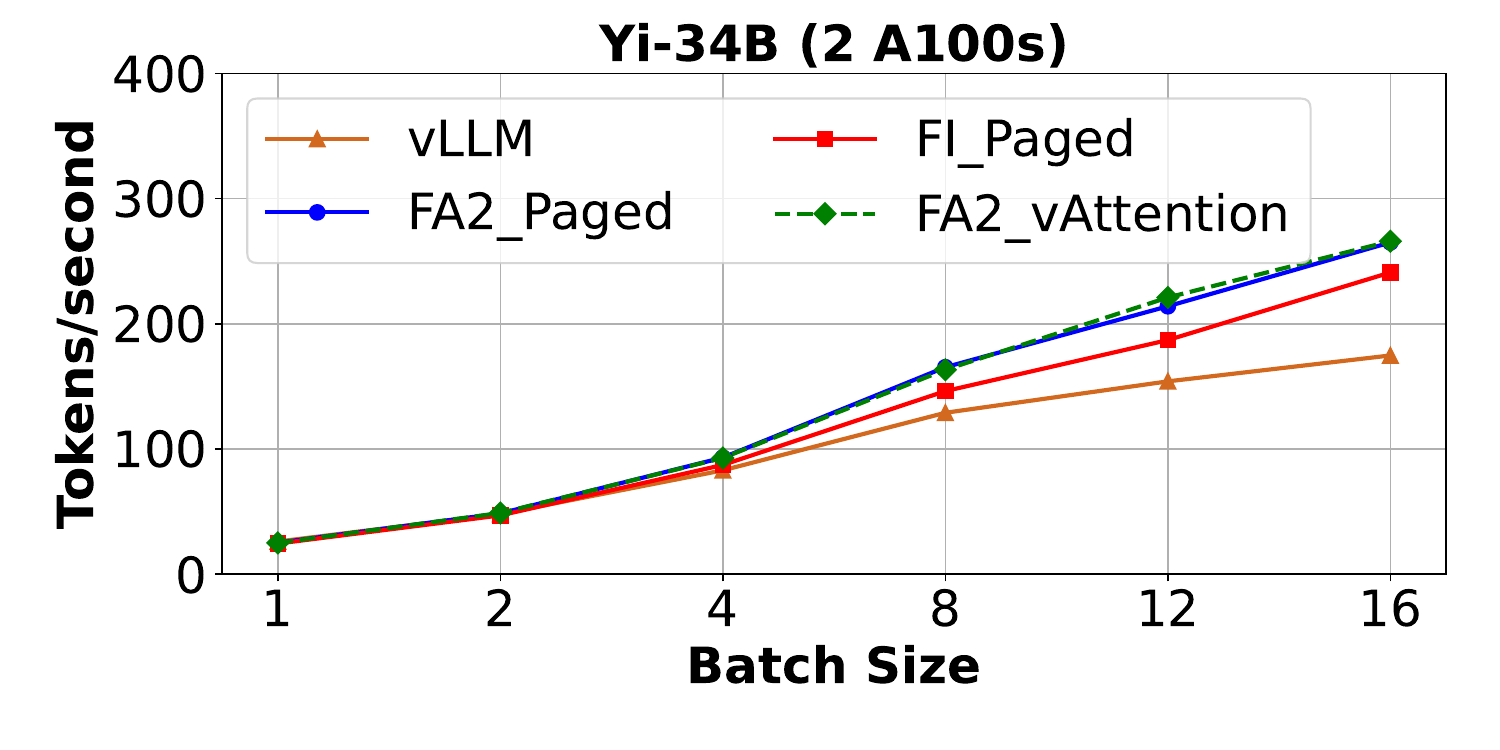}
    \end{subfigure}
    \caption{Decode throughput. \favattention is on par with \fapaged (note the overlapping lines) which is the best among all \pa based alternatives, while outperforming \fipaged and \vllm.}
    \label{fig:eval:decode-throughput}
\end{figure*}

\noindent
\textbf{Long contexts:} The contribution of attention computation becomes significant at 16K and higher context lengths in our experiments. Therefore, even for \flashattention back-end, \sysname outperforms the paged counterpart. For example, at context length 192K, \favattention outperforms \fapaged by $1.24\times$, $1.26\times$ and $1.24\times$  for \yismall, \llamasmall and \yimedium, respectively. Similarly, for \flashinfer back-ends, \fivattention improves prefill throughput by up to $1.25\times$ and $1.36\times$ for \yismall and \llamasmall (context length 16K), and  $1.17\times$ for \yimedium (context length 32K).

\noindent
\textbf{Attention time:} \sysname's improvement in prefill throughput is primarily due to faster attention kernels enabled by a (virtually) contiguous \kvcache. This is because the prefill phase of a long prompt is primarily dominated by attention computation, as can be observed by comparing the numbers inside parenthesis with total prefill completion time in~\autoref{table:eval:ttft}. For \flashattention, nearly all the gains of \sysname are due to faster attention kernels, e.g., prefill gains of 1.5 seconds, 7.4 seconds and 16.9 seconds for \yismall are all due to gains in attention computation. \sysname enabled kernels also help with the \flashinfer back-end. In addition, \fipaged also has other sources of overheads, e.g., in the 14 seconds of total savings (\yimedium, 192K context), only 7 seconds is due to attention and the rest is due to other sources.

\begin{table}[t]
\scalebox{0.85}{
\begin{tabular}{llcccc}
Model & BS & vLLM & \fapaged & \fipaged & \favattention \\ \toprule
\multirow{2}{*}{\yismall} & 16 & 32.3 & 11.5  & 15.2 & 11.3 \\
 & 32 & 64.1 & 25.5 & 25.4 & 25.3 \\ \hdashline
\multirow{2}{*}{\begin{tabular}[c]{@{}l@{}}Llama-3\\ -8B\end{tabular}} & 16 & 17.8 &  11.9 & 12.1 & 11.8 \\
 & 32 & 35.3 & 25.4 & 23.23 & 25.3 \\ \hdashline
 \multirow{2}{*}{\yimedium} & 12 & 41.4 & 17.4 & 24.1  &  17.4 \\
 & 16 & 55.1 & 21.7 & 28.8 & 21.8 \\ \bottomrule
\end{tabular}}
\caption{Total latency of attention kernel (sum of all layers) per decode iteration (in milliseconds, BS = batch size).}
\label{table:eval:decode-attn}
\end{table}

\subsection{Decode Evaluation}
\label{sec:eval:decode}

For decodes, in addition to \fapaged and \fipaged, we also evaluate the throughput obtained with \vllm's decode kernel (the first ever kernel to support \pa). For \sysname, we use \flashattention's non-paged kernel. Unfortunately \flashinfer's non-paged decode kernel has significantly higher latency (up to $14.6\times$) compared to all these other kernels. Hence, while \sysname can support dynamic memory allocation for \flashinfer's non-paged decode kernel, we omit it for evaluation in this section. 

 \autoref{fig:eval:decode-throughput} shows the decode throughput of \yismall, \llamasmall and \yimedium with varying batch size up to 32 (except for \yimedium which runs out of memory for batch size 32). We set the initial context length of each request to 16K tokens and calculate decode throughput based on the mean latency of 400 decode iterations.

First, \sysname is on par with the best of \pa as shown by \fapaged and \favattention in~\autoref{fig:eval:decode-throughput}. In comparison, \fipaged has somewhat lower throughput and \vllm is the worst for all models and configurations. For example, \fapaged and \favattention outperform \vllm by up to $1.99\times$, $1.58\times$ and $1.53\times$ for \yismall, \llamasmall and \yimedium, respectively. The primary reason is that \vllm's decode kernel has significantly higher latency than the \flashattention based kernels; while \flashattention has continuously adopted new optimizations (e.g., FlashDecoding~\cite{flashdecoding}), \vllm has lagged behind. For example,~\autoref{table:eval:decode-attn} shows that \vllm's \pa kernel incurs up to $2.8\times$, $1.5\times$, and $2.5\times$ higher latency than \flashattention kernels for \yismall, \llamasmall and \yimedium. This is despite \vllm being in an actively maintained open-source serving stack, as well as being used by various companies for serving LLMs. This is an important result that underlines the importance of low software complexity and portability.

Second, relative gains of \favattention and \fapaged increase over \vllm with the batch size, e.g., as batch size increases from 4 to 32 for \llamasmall, relative gains increase from $1.05\times$ to $1.58\times$. This is because the latency of a decode attention kernel is proportional to the total number of tokens in the batch~\cite{vidur}. Therefore, the contribution of attention kernel in the overall latency -- hence gains with a more efficient kernel -- increase with the batch size. Further, for the same reasons as discussed in~\autoref{sec:eval:prefill} (faster attention and lower CPU overhead), \favattention delivers up to $1.23\times$ higher throughput than \fipaged (\yismall, batch size 12).

Finally, note that \sysname is only as good as the state-of-the-art \pa for decode throughput, as compared to prefills where it outperforms \pa. This is because in case of decode attention, paged and non-paged kernels have similar latency as shown in~\autoref{table:eval:decode-attn}. We believe this is due to memory bound nature of decode attention, i.e., memory stalls make it possible to hide the effect of additional compute that paging support requires.  However, hiding compute overhead in a prefill attention kernel is hard because it is already compute bound (see~\autoref{table:eval:ttft} for \sysname's gains in the prefill kernel).

\begin{figure}[t!]
    \centering
    \includegraphics[trim={0 35 0 0}, clip=True, width=0.95\columnwidth]{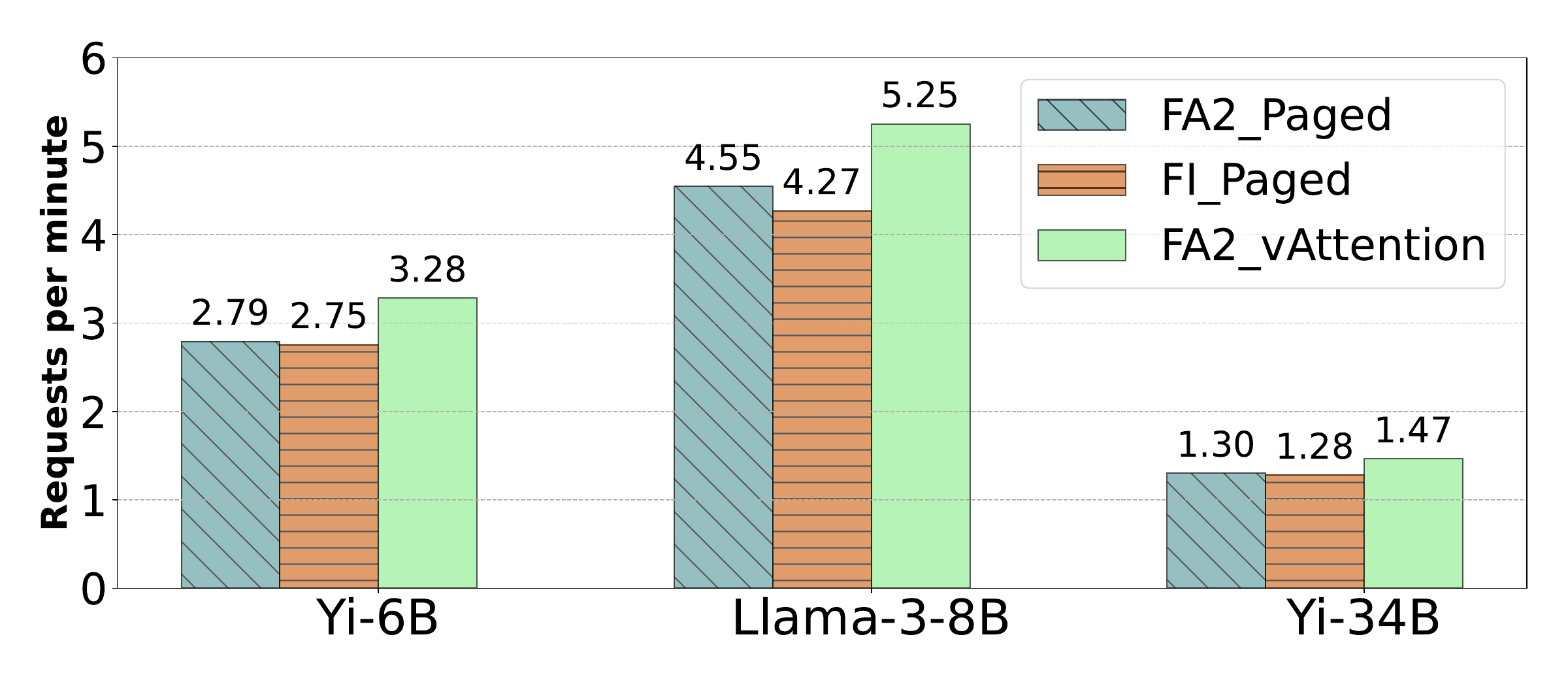}
    \caption{Offline inference throughput while serving long-context requests from the \arxivsummarization dataset~\cite{arxiv}.}
    \label{fig:eval:offline:arxiv}
\end{figure}

\subsection{End-to-end Performance: Offline Scenarios}
\label{sec:eval:makespan}

We measure the end-to-end system throughput in an offline scenario as the number of requests completed per minute. We evaluate a total of 427 long-context requests from the \arxivsummarization workload trace wherein the total context length per-request varies from 64K tokens to 192K tokens and the number of output tokens per-request varies from 17 to 5153. The mean prefill to decode token ratio is 356.~\autoref{fig:eval:offline:arxiv} shows throughput for different attention backends for all three models. \favattention outperforms \fapaged by $1.18\times$, $1.15\times$ and $1.13\times$, and \fipaged by $1.19\times$, $1.23\times$, and $1.14\times$ for \yismall, \llamasmall and \yimedium, respectively. In general, \sysname's performance gains are proportional to the context length and the prefill to decode token ratio (P:D ratio); these factors determine how much prefill attention contributes to the total runtime. A higher P:D ratio as well as longer context lengths indicate that the workload is more prefill bound. \sysname provides higher gains in such cases. 

\subsection{End-to-end Performance: Online Scenarios}
\label{sec:eval:online}

\begin{figure}[t!]
    \centering
    \begin{subfigure}[b]{0.49\columnwidth}
    \centering
        \includegraphics[width=\textwidth]{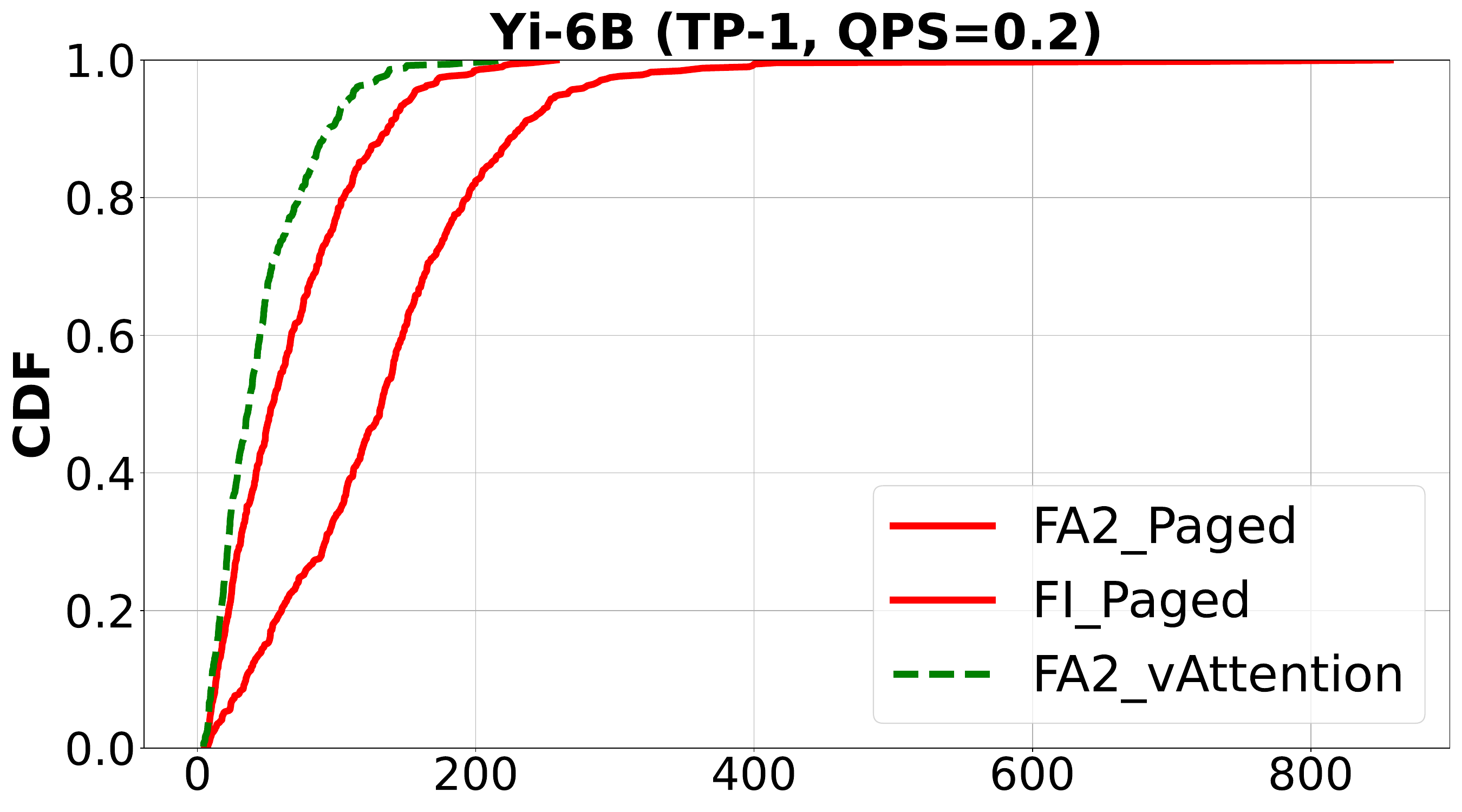}
    \end{subfigure}
    \begin{subfigure}[b]{0.49\columnwidth}
    \centering
        \includegraphics[width=\textwidth]{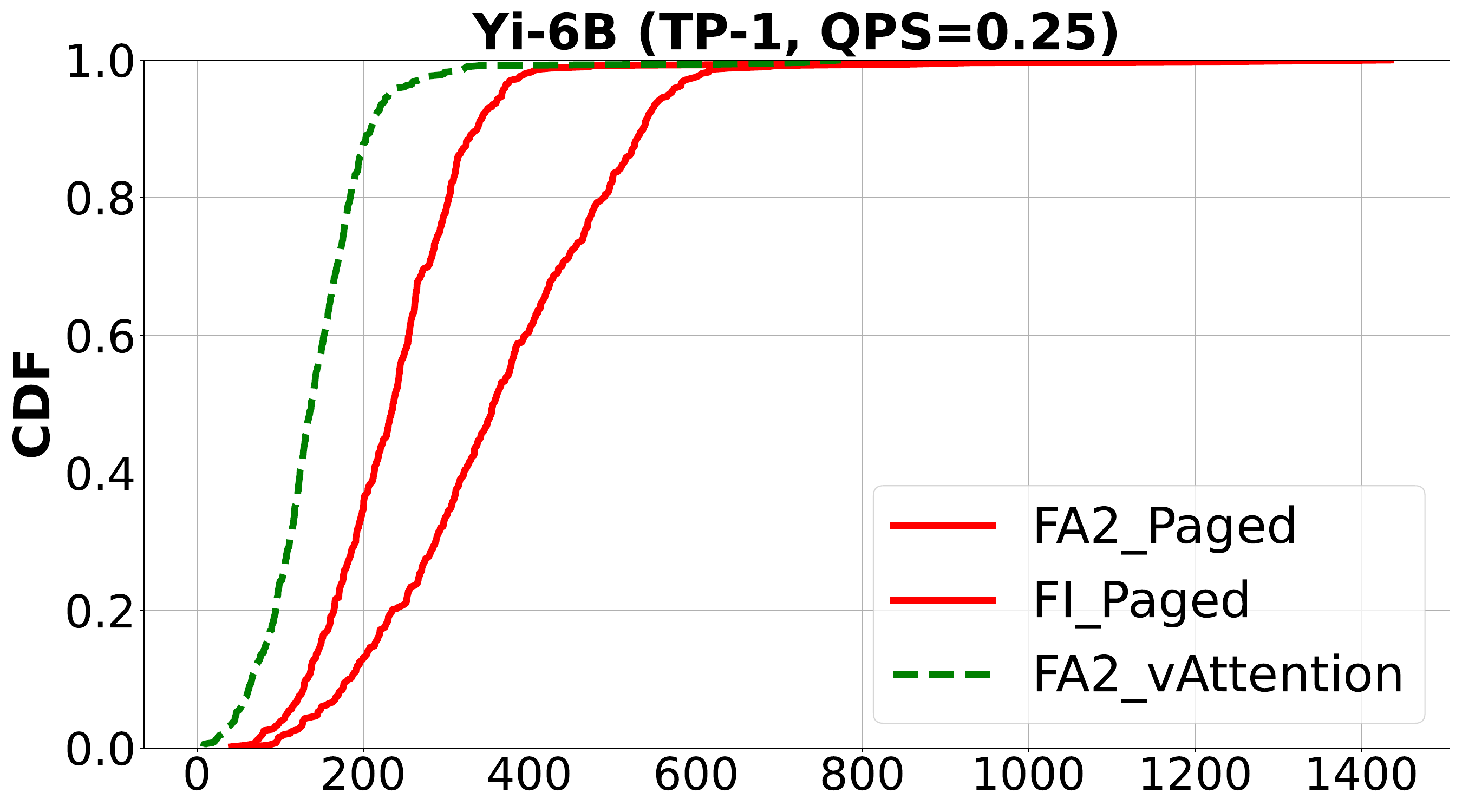}
    \end{subfigure}
    \\
    \begin{subfigure}[b]{0.49\columnwidth}
    \centering
        \includegraphics[width=\textwidth]{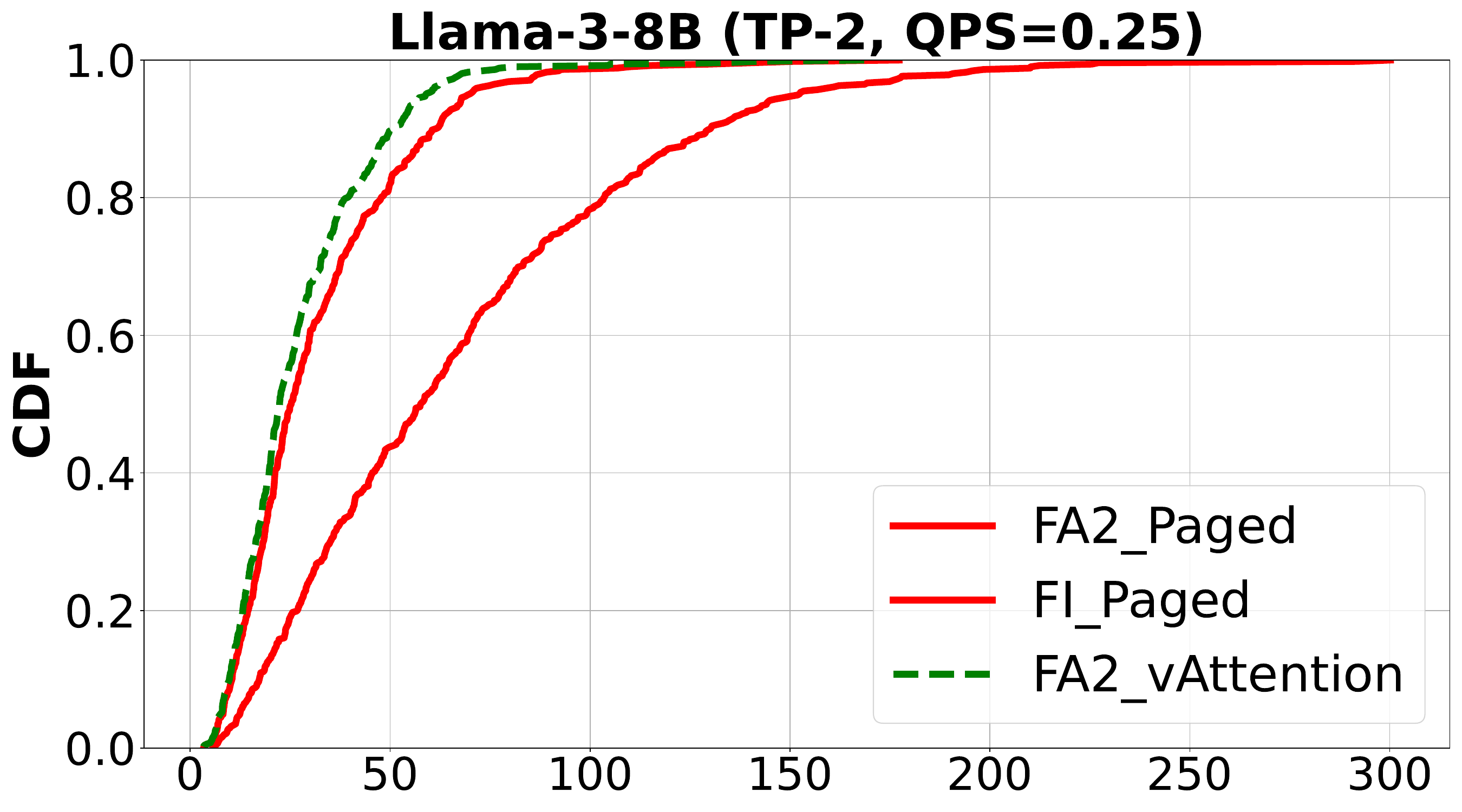}
    \end{subfigure}
    \begin{subfigure}[b]{0.49\columnwidth}
    \centering
        \includegraphics[width=\textwidth]{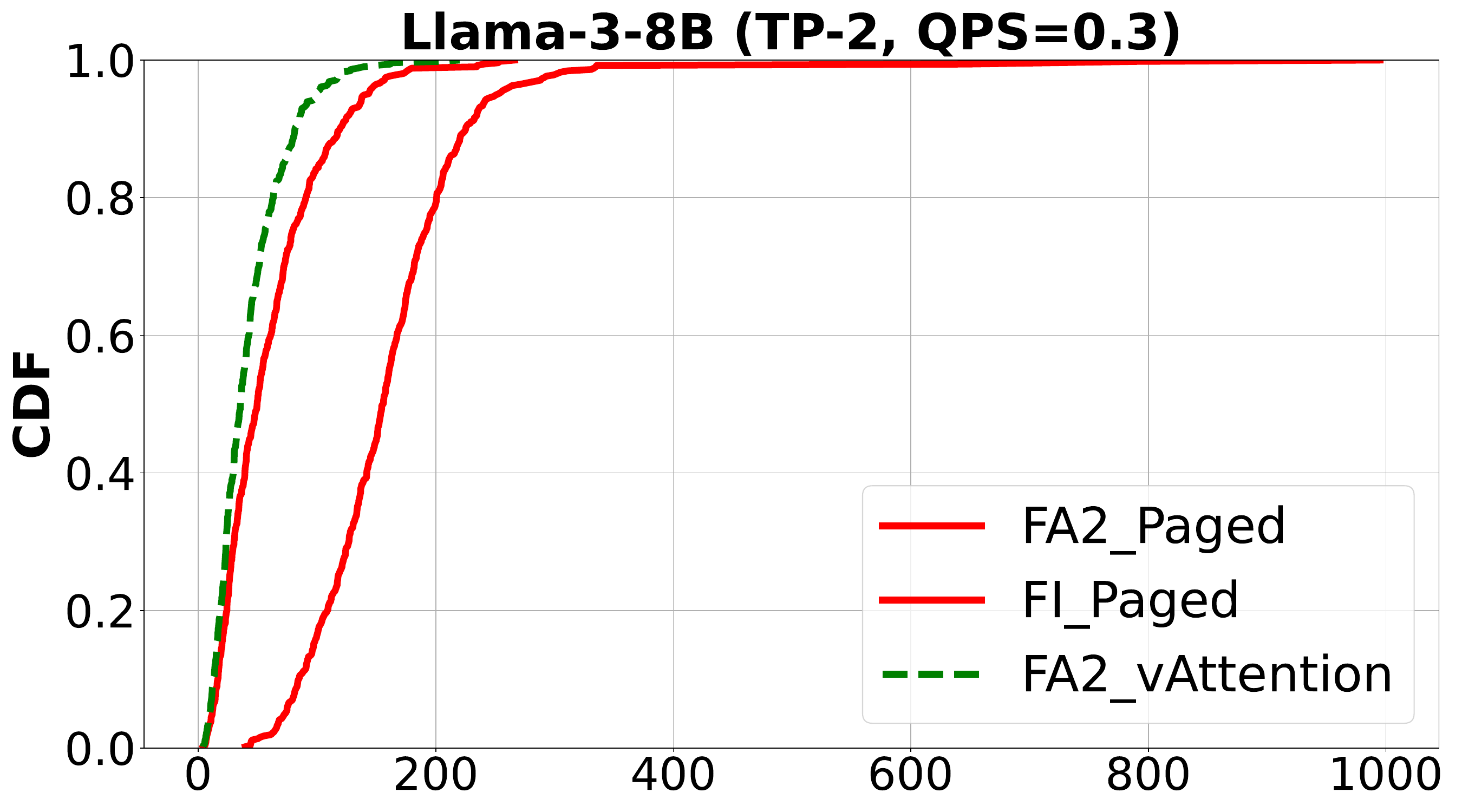}
    \end{subfigure}
    \\
        \begin{subfigure}[b]{0.49\columnwidth}
    \centering
        \includegraphics[width=\textwidth]{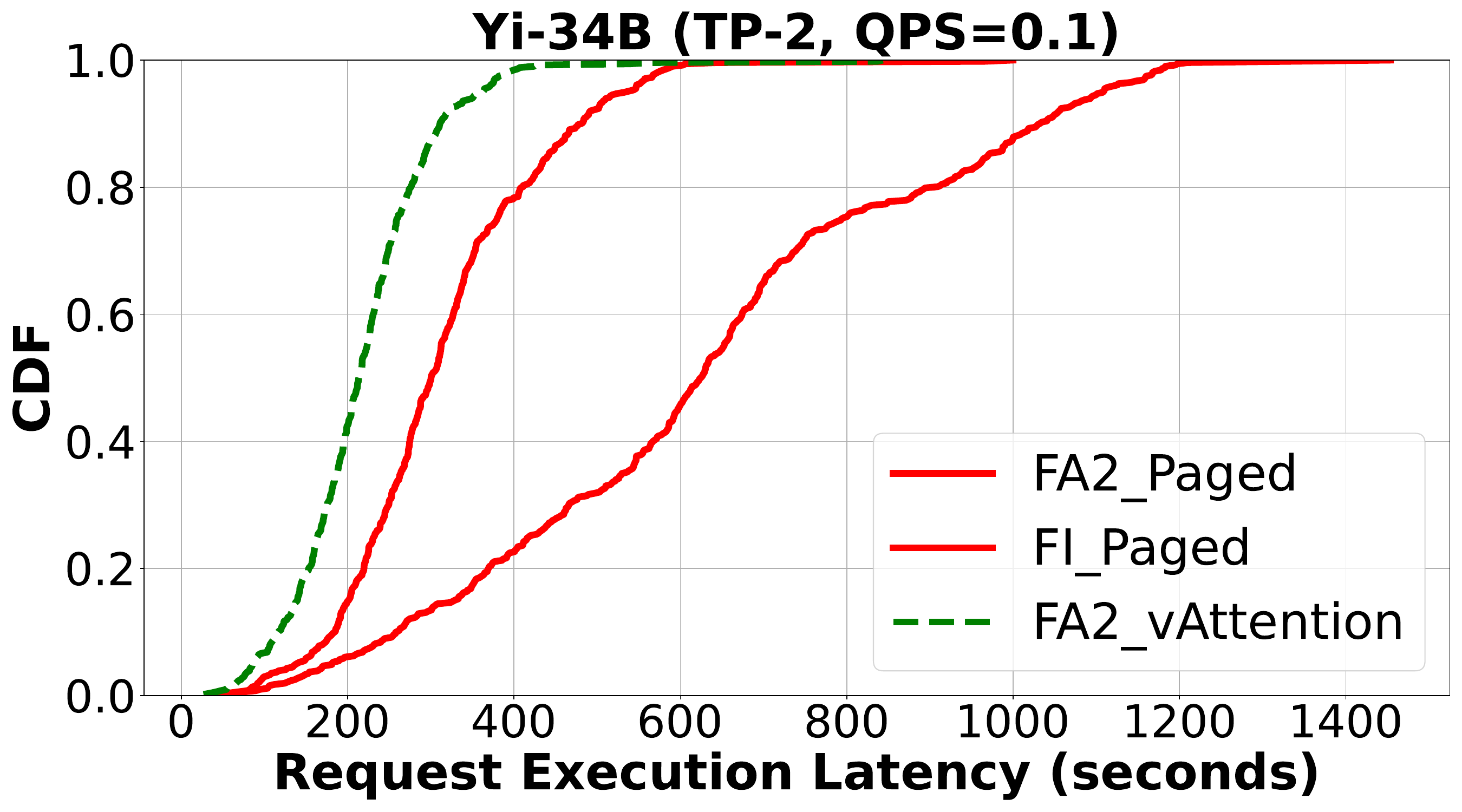}
    \end{subfigure}
    \begin{subfigure}[b]{0.49\columnwidth}
    \centering
        \includegraphics[width=\textwidth]{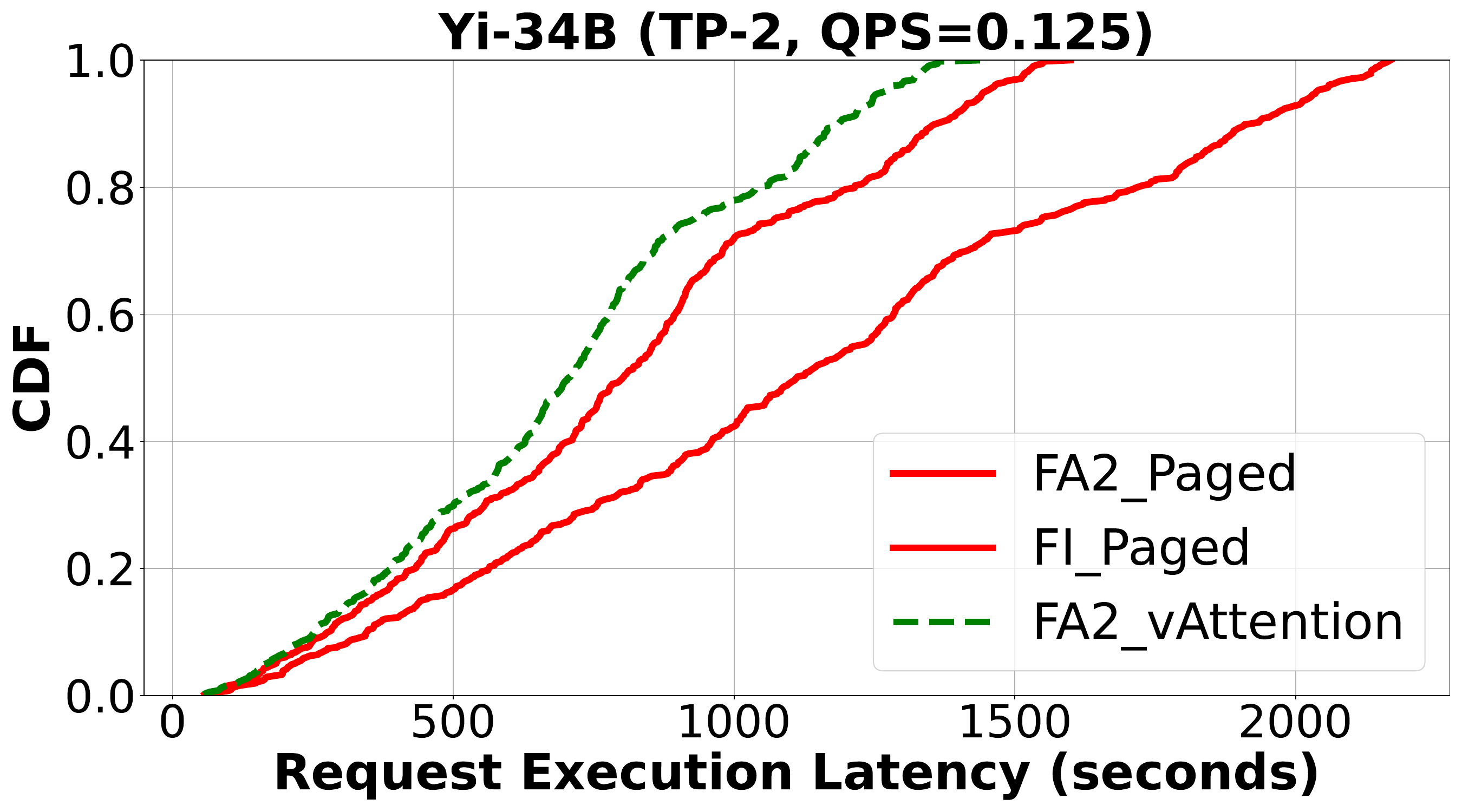}
    \end{subfigure}
    \caption{CDF of end-to-end request execution latency in online inference under varying load.}
    \label{fig:eval:online}
\end{figure}

We evaluate an online inference scenario serving long-context requests from \arxivsummarization~\cite{arxiv}. In total, we run 512 requests wherein the per-request input context length varies from 22K to 45K tokens (mean 29K), the number of decode tokens varies from 6 to 3250 (mean 348) and the mean P:D ratio is 129. We evaluate performance near the serving capacity of the system which denotes the maximum load a system can handle without incurring high queuing delays~\cite{vllmsosp,sarathiserve2024}. We vary input load (queries-per-second or QPS) based on Poisson distribution and schedule requests in first-come-first-serve order.~\autoref{fig:eval:online} shows the CDF of the end-to-end request execution latency for this experiment. We see that \sysname consistently outperforms both baselines. For example, \favattention reduces the median request execution latency over \fapaged by up to 42\% (QPS 0.25) for \yismall, by up to 28\% (QPS 0.3) for \llamasmall and  by up to 29\% for \yimedium (QPS 0.1).  The primary reason for \sysname's performance gain is that it can compute the prefill phase of new requests faster than the \pa-based systems which substantially reduces queuing delays. Consistent with our prior results, \fipaged is slower than \fapaged and hence our gains our \fipaged are relatively higher compared to our gains over \fapaged.

\begin{figure}[t!]
    \centering
    \includegraphics[trim={0 35 0 0}, clip=True, width=0.95\columnwidth]{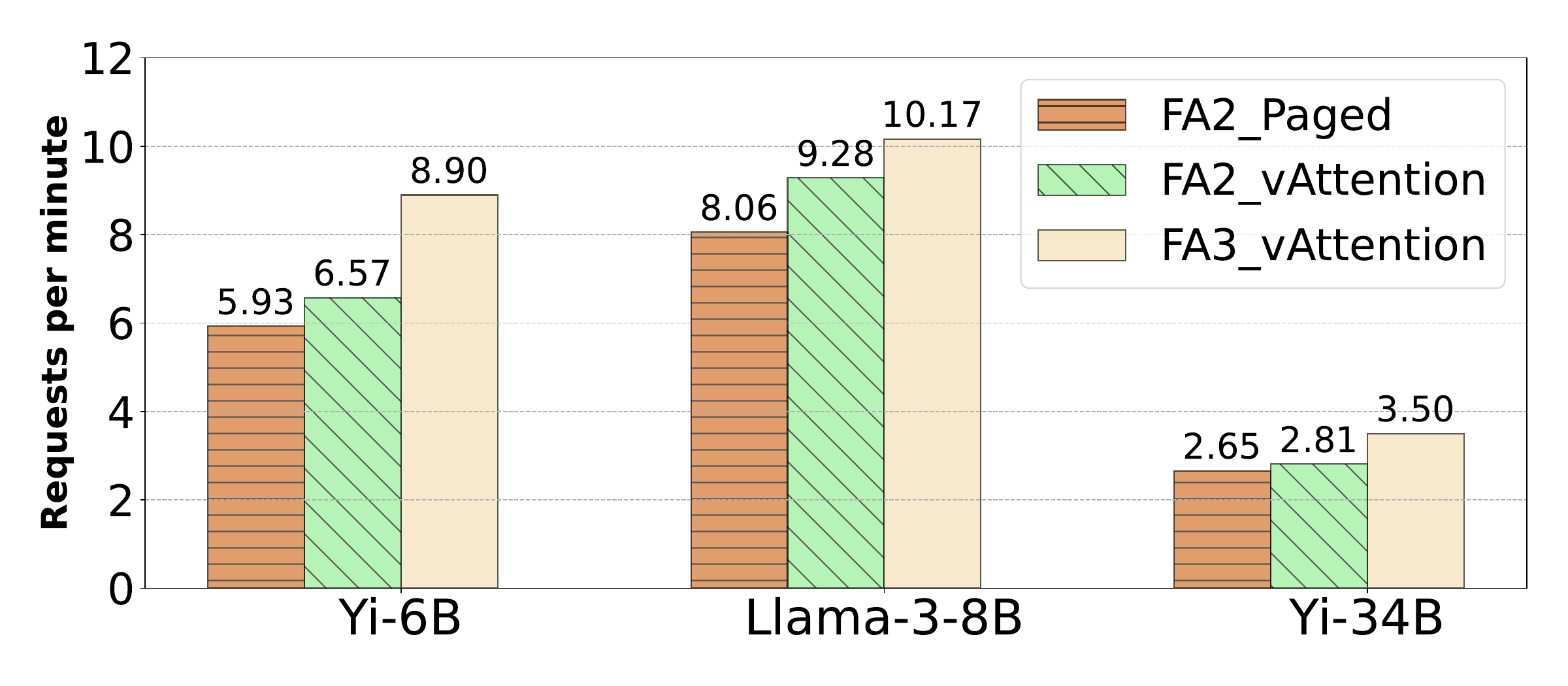}
    \caption{Offline inference throughput on H100 GPUs.}
    \label{fig:eval:offline:arxiv-fa3}
\end{figure}

\subsection{Exemplifying Portability: FlashAttention-3}
\label{sec:eval:fa3}

We demonstrate the portability benefit of \sysname  with the recently released FA3 kernel~\cite{flash-attention-3}. FA3 is optimized for the NVIDIA Hopper architecture and did not support \pa when released. Therefore, dynamic memory allocation via \pa is not feasible for FA3 at the time of writing this paper (integration of FA3 into \vllm is also a work in progress~\cite{fa3-vllm-integration-issue, fa3-vllm-integration-roadmap}). \sysname not only enables dynamic memory allocation with FA3, it also requires no code changes to deploy FA3.~\autoref{fig:eval:offline:arxiv-fa3} shows the offline inference throughput of our models with 1--2 H100 GPUs (\yismall deployed on a single GPU and the other models on two GPUs each) on the same \arxivsummarization-based workload as evaluated in~\autoref{sec:eval:makespan}. FA3 with \sysname provides an additional speedup of up to $1.35\times$ (\yismall) over \favattention which is already up to $1.15\times$ faster than \fapaged.

\subsection{Ablation Studies}
\label{sec:eval:ablation}

\begin{figure}[t!]
    \centering
    \includegraphics[trim={0 20 0 20}, clip=True, width=0.95\columnwidth]{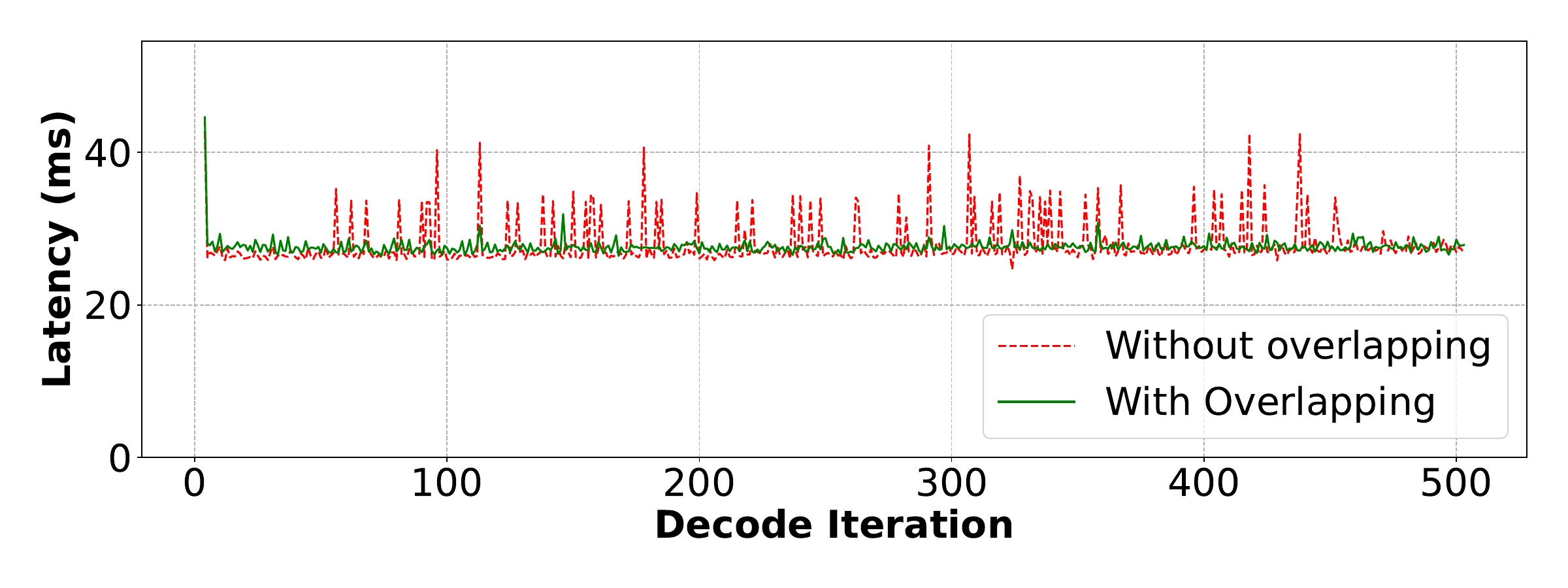}
    \caption{Latency of decode iterations with and without overlapping memory allocation with compute (batch size=32, context length=4K--8K, model: \llamasmall)}
    \label{fig:eval:decode-latency}
\end{figure}

\subsubsection{Hiding allocation latency} \autoref{fig:eval:decode-latency} shows the latency of more than 500 decode iterations of \llamasmall. For this experiment, we used a batch of 32 requests with per request prefill context length varying between 2K to 3K tokens. It is evident that overlapping memory allocation with compute effectively hides the latency of  CUDA VMM APIs. We used 2MB pages for this experiment to show that even the worst case memory allocation latency can be hidden by overlapping it with compute (\autoref{tab:design:cudaapis} shows that 2MB pages incur highest latency). In contrast, allocating memory synchronously via CUDA APIs leads to frequent latency spikes of 5 to 15 milliseconds, depending on how many requests require physical memory allocation in a given iteration.

\subsubsection{Deferred reclamation} 
\begin{figure}[t!]
    \centering
    \includegraphics[trim={0 50 0 0}, clip=True, width=0.9\columnwidth]{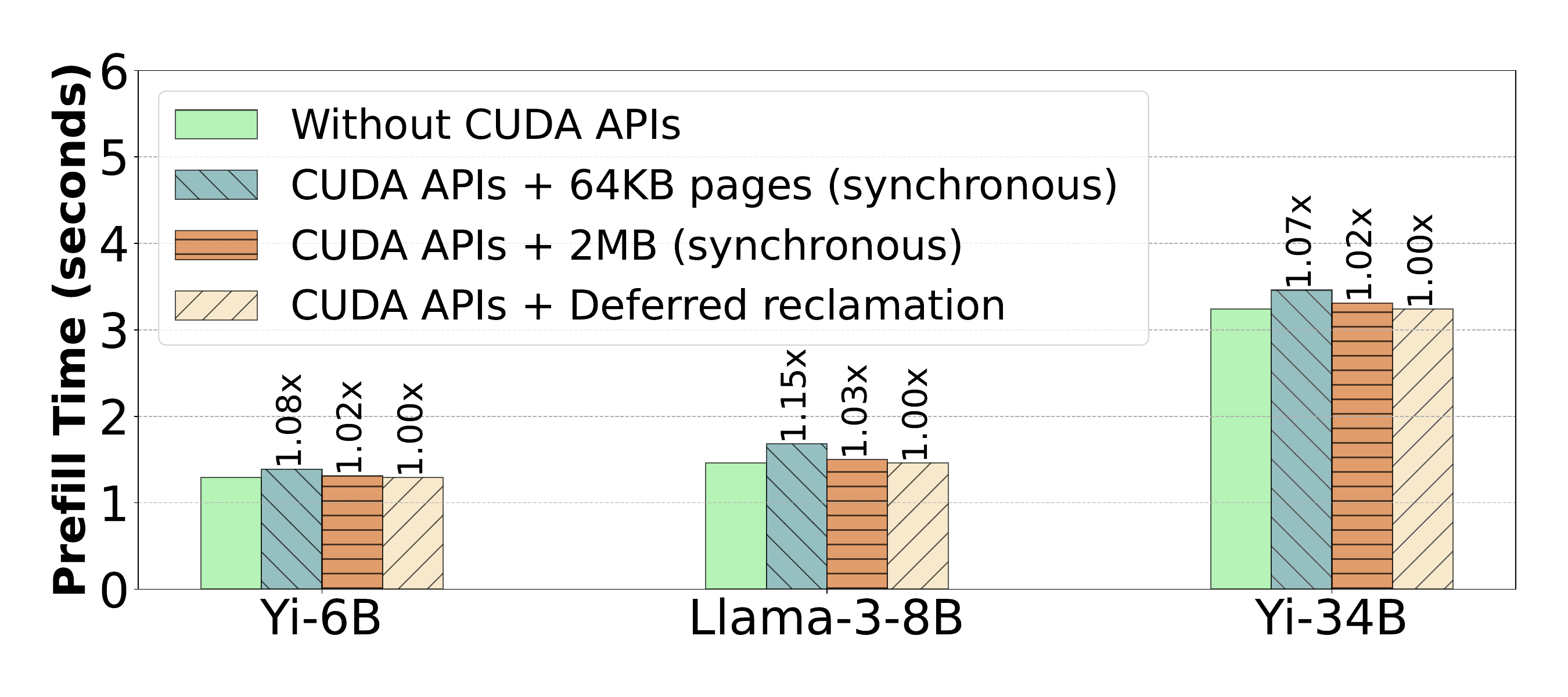}
    \caption{Prefill completion time of a single prompt of 16K tokens with different memory allocation strategies.}
    \label{fig:eval:prefill-time}
\end{figure}

\autoref{fig:eval:prefill-time} shows that synchronous memory allocation incurs prefill overhead of up to $1.15\times$ with 64KB pages and up to $1.03\times$ with 2MB pages. In most cases, deferred reclamation eliminates the need to invoke CUDA VMM APIs for prefills because a newly arrived request can simply re-use physical memory allocated to a prior request. This way, deferred reclamation ensures that prefill latency is not affected by memory allocation.

\begin{figure}[t!]
    \centering
    \begin{subfigure}[b]{0.48\columnwidth}
    \centering
        \includegraphics[trim={0 25 0 0}, clip=True, width=\textwidth]{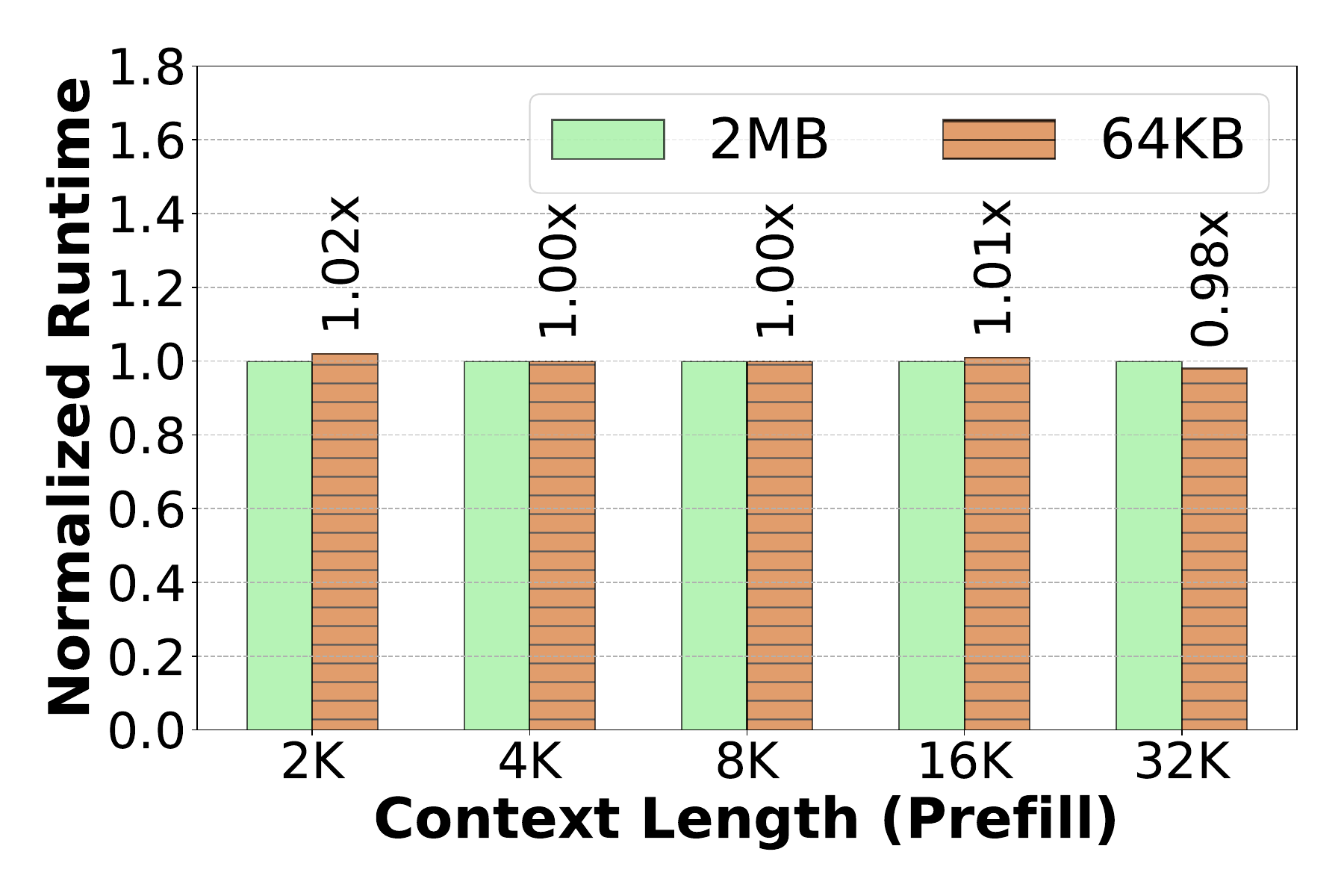}
    \end{subfigure}
    \begin{subfigure}[b]{0.48\columnwidth}
    \centering
        \includegraphics[trim={0 25 0 0}, clip=True,  width=\textwidth]{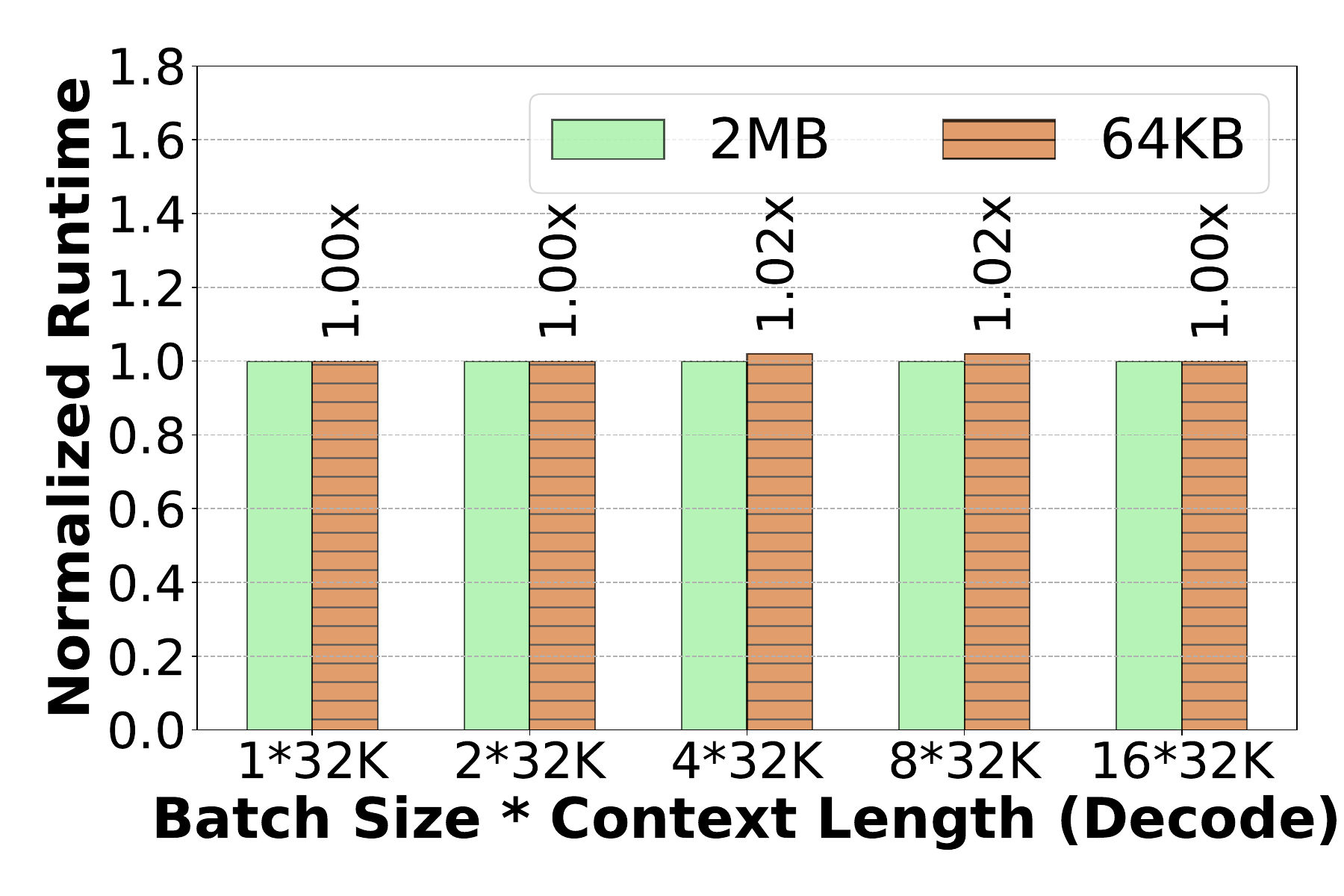}
    \end{subfigure}
    \caption{Effect of page size on the runtime of \flashattention's prefill (left) and decode (right) attention kernels (model: \llamasmall). }
    \label{fig:eval:micro:pagesize}
\end{figure}

\subsubsection{Effect of page size.} In many applications, use of smaller pages can potentially degrade performance due to TLB thrashing~\cite{neighbourhoodawareat,ingens,hawkeye,architecturalsupportforgpuat}. We find that this is not the case for LLM inference. For example, \autoref{fig:eval:micro:pagesize} shows that the execution latency of an attention kernel remains largely unaffected when the \kvcache is allocated using 64KB pages, compared to using 2MB pages. In a separate experiment, we find that these results are also consistent with very large models, e.g., Llama-3-70B and GPT-3-175B. We attribute this to the regular computation pattern of the attention operator as well as the hand-tuned implementations that explicitly try to avoid irregular memory accesses.

\begin{table}[t!]
\scalebox{0.9}{
\color{black}\begin{tabular}{l|rrrc}
\multirow{2}{*}{\textbf{Model}} & \multicolumn{4}{c}{\textbf{Block Size (\# Tokens in a page-group)}} \\ 
 & 64KB & 128KB & 256KB & 2MB \\ \toprule
\yismall (TP-1) & 64 & 128 & 256 & 2048  \\
\yismall (TP-2) & 128 & 256 & 512 & 4096  \\ \hline
\llamasmall (TP-1) & 32 & 64 & 128 & 1024  \\
\llamasmall (TP-2) & 64 & 128 & 256 & 2048  \\ \hline
\yimedium (TP-1) & 32 & 64 & 128 & 1024   \\
\yimedium (TP-2) & 64 & 128 & 256 & 2048 \\  \bottomrule
\end{tabular}}
\caption{\kvcache block size as a function of page-group size and degree of tensor parallelism.}
\label{table:eval:blocksize:fragmentation}
\end{table}

\begin{figure}[t!]
    \centering
    \includegraphics[trim={0 50 0 45}, clip=True, width=0.9\columnwidth]{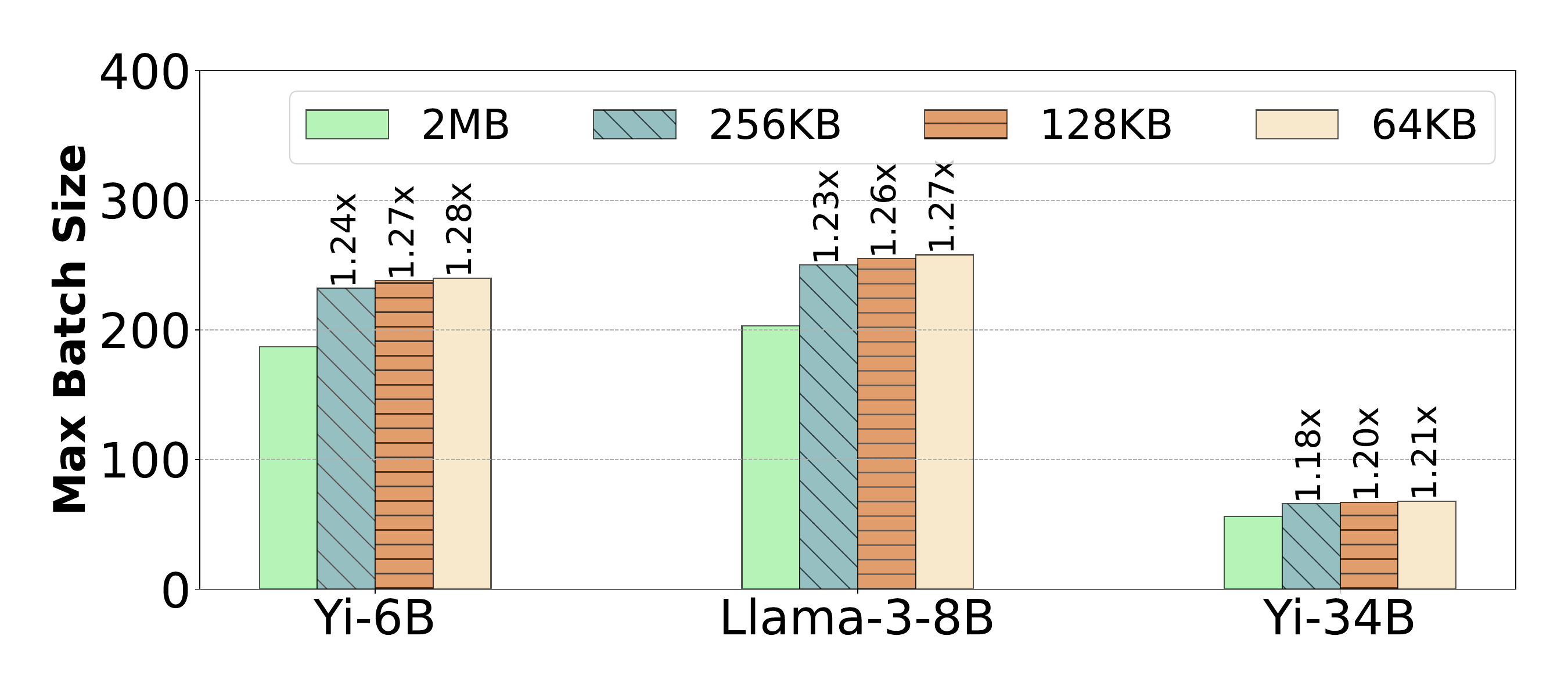}
    \caption{Maximum batch size obtained with different page-group sizes for a dynamic workload.}
    \label{fig:eval:batchsize}
\end{figure}

\autoref{table:eval:blocksize:fragmentation} shows the block size i.e., number of tokens per physical page-group. Smaller page-groups enable fine-grained allocation, approaching \vllm's recommended block size of 16--32 (the minimum block size in \flashattention is 256 -- higher than \sysname). This shows that \sysname is as effective in reducing fragmentation as \pa. In terms of end-to-end system performance, we find that 2MB pages are good enough for many online serving scenarios where latency constraints prevent use of very high batch sizes. In contrast, smaller page-groups are better for throughput-oriented scenarios where maximizing batch size is important to obtain peak performance. For example, using 2MB pages could serve batch sizes of up to 187, 203 and 56 for \yismall, \llamasmall and \yimedium on a dynamic trace (dataset: OpenChat~\cite{wang2023openchat}, load: 7 queries per seconds). In contrast, 64KB pages helped serve batch sizes of up to 240, 258 and 68 for our models as shown in~\autoref{fig:eval:batchsize}.

\begin{table}[t!]
    \centering
    \begin{tabular}{c|c|c|c|c}
      Config. & 64KB & 128KB & 256KB & 2MB \\ \toprule
       TP-1 & 7.59 & 14.56 & 27.04 & 35.17 \\
       TP-2 & 15.18 & 29.12 & 54.08 & 70.34 \\ \bottomrule
    \end{tabular}
    \caption{Physical memory allocation bandwidth (GB per second) with varying allocation granularity.}
    \label{tab:eval:allocation-bandwidth}
\end{table}

\subsubsection{Memory allocation bandwidth} \autoref{tab:eval:allocation-bandwidth} shows that even with 64KB pages (our smallest), \sysname can allocate as much as 7.6GB physical memory per second per GPU. This is more than an order of magnitude higher than the maximum memory allocation rate of 750MB per second of decodes (\autoref{fig:analysis:llm:memory}). Larger page-group sizes and higher TP dimensions increase the memory allocation rate proportionally. Therefore, memory allocation bandwidth of CUDA VMM APIs is more than sufficient for LLM inference.

\section{Discussion}
\label{sec:discussion}

\subsection{Managing \kvcache via Unified Memory}

To enable dynamic memory allocation, we also considered leveraging unified memory via \cudamallocmanaged~\cite{nvidia-cuda,gpu-mismanaged}. However, we find that unified memory support is currently not suitable for serving LLMs. First, it does not support partial freeing, preventing reclamation of physical memory of individual requests. Second, it lacks support for memory aliasing which prevents de-duplication of \kvcache content in physical memory (de-duplication is useful when requests share a common prefix~\cite{vllmsosp}), consequently limiting batch size. \cudamallocmanaged also allocates 2MB pages by default which can cause severe fragmentation. However, our changes in NVIDIA drivers are based on unified memory: we allocate virtual tensors using \cudamallocmanaged, enabling support for partial freeing and page sharing with additional APIs. In addition, we introduce latency hiding optimizations and support for smaller pages. Therefore, one could consider our extensions to NVIDIA drivers as ``unified memory optimized for LLM serving''.

\begin{table}[t!]
\scalebox{0.9}{
\color{black}\begin{tabular}{l@{\hspace{6pt}}|c@{\hspace{6pt}}|c@{\hspace{6pt}}} \\
\multirow{2}{*}{\textbf{Model}} & \multicolumn{2}{c}{\textbf{Block size (\# Tokens in a 2MB page)}} \\ 
 &  w/o Tensor Slicing & w/ Tensor Slicing \\ \toprule
\yismall (TP-1) & 2048 & 64  \\
\yismall (TP-2) & 4096 & 128 \\ \hline
\llamasmall (TP-1) & 1024 & 32  \\
\llamasmall (TP-2) & 2048 & 64 \\ \hline
\yimedium (TP-1) & 1024 & 18 \\
\yimedium (TP-2) & 2048 & 36\\  \bottomrule
\end{tabular}}
\caption{\kvcache block size with and without tensor slicing with 2MB pages.}
\label{table:eval:megacache:blocksize}
\end{table}

\subsection{Reducing Fragmentation via Tensor Slicing}
To reduce fragmentation caused by 2MB pages, we also provide an alternative method that does not require modifying  NVIDIA drivers. In this method, we use a single 2MB page to store the \kvcache tokens of all layers for a given request. This can be done by allocating one virtual tensor of shape $[B, L, N, H, D]$ for \kcache (and one for \vcache) and slicing it across all layers, instead of allocating $2\times N$ virtual tensors of shape $[B, L, H, D]$. This reduces fragmentation to $1/N$ of the prior design as shown in~\autoref{table:eval:megacache:blocksize}. In this design, the \kcache (or \vcache) of a particular layer $n$ is represented by slice $[B, L, n:n+1, H, D]$ of the original tensor, and can be passed to attention kernels for computation. However, note that with tensor slicing, the \kcache (and \vcache) of each layer is no longer contiguous. Computing attention over such a tensor slice is possible only if the the attention kernel supports memory addressing with strides. While many kernels (e.g., \flashattention) support strides out-of-the-box, the earlier versions of \flashinfer lacked such support~\cite{fi-strides}. Therefore, relying on tensor slicing as the primary method of reducing fragmentation would have prevented \sysname from supporting \flashinfer kernels. Hence, to support unmodified \flashinfer kernels, we chose to add support for smaller pages in NVIDIA drivers. Importantly, our two solutions are compatible with each other; one could deploy smaller pages with tensor slicing to further reduce fragmentation, if required.



\subsection{Programming Effort}
\begin{figure}[t!]
    \centering
    \includegraphics[width=0.85\columnwidth]{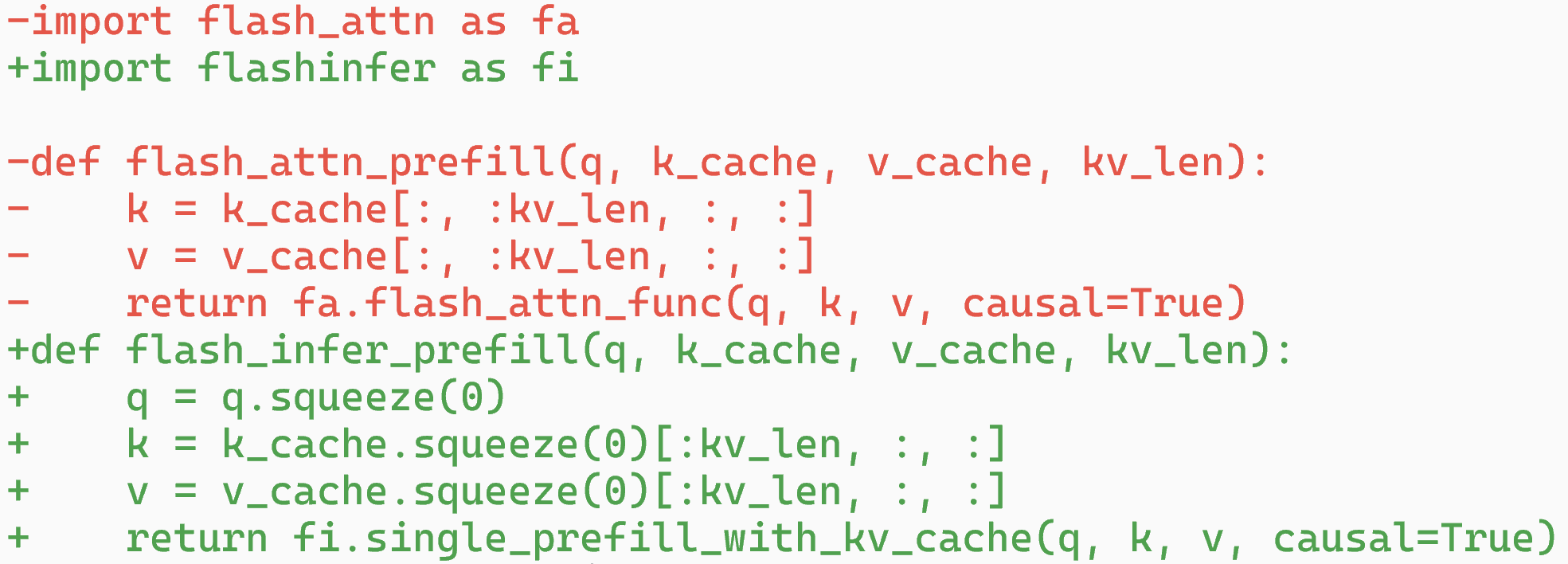}
    \caption{Illustration of code changes needed to replace the prefill attention kernel of \flashattention by \flashinfer when using \sysname for memory allocation.}
    \label{fig:eval:codediff}
\end{figure}

PagedAttention requires significant programming effort. For example, integrating \flashinfer decode kernels in \vllm required  more than 600 lines of code changes in over 15 files~\cite{vllmfi1,vllmfi2,vllmfi3}. Implementing the initial paging support in \flashattention kernel also required $\approx 280$ lines of code changes~\cite{fapaging} and additional efforts to support small block sizes~\cite{fasmall}. In contrast, \sysname makes it feasible to replace one attention kernel with another with only a few lines of code changes in the serving framework (\autoref{fig:eval:codediff}). Note that the example shown in~\autoref{fig:eval:codediff} has no interaction with memory management; this is precisely the goal of \sysname, i.e., when a slow kernel is replaced by a fast kernel, memory management should continue to work transparently.

\section{Related Work}
\label{sec:relatedwork}

Optimizing LLM inference is an active area of research ~\cite{helix, griggs2024melangecostefficientlarge, miao2023spotserveservinggenerativelarge, zhu2024nanoflowoptimallargelanguage, kossmann2024gpuhalfemptyhalffullpractical, strati2024dejavukvcachestreamingfast, 
lee2024infinigenefficientgenerativeinference, singhania2024lokilowrankkeysefficient, zheng2024sglangefficientexecutionstructured, ma2024compressingkvcachelongcontext,mahmud2024mapleframeworkactivepreference}. Various techniques have been proposed to improve different aspects of LLM serving like batching~\cite{orca, vllmsosp, sarathiserve2024}, disaggregation~\cite{patel2023splitwise, distserve2024, tetriinfer}, scheduling~\cite{flexgen, fastserve}. However, central to all these techniques is the need for efficient \kvcache memory management. Since \vllm, \pa has been adopted in various serving frameworks e.g., TensorRT-LLM~\cite{trtllmgithub}, LightLLM~\cite{lightllm:github}, and libraries e.g., \flashattention~\cite{fagithub} and \flashinfer~\cite{figithub}. Some other concurrent works have also proposed optimizations for attention kernels~\cite{flashattnhopper,mirage-tensor-compiler, thunderkittens, h2o2023, flashdecoding++,leanattention,flash-attention-3, pod-attn-2024}. Deploying new kernels for inference is hard with \pa.  \sysname offers an alternate -- which we believe is also a more principled -- approach to dynamic \kvcache memory management that makes it easier to deploy new kernels.

In a recent work, GMLake~\cite{gmlakeasplos24} showed that using CUDA virtual memory support can mitigate fragmentation in DNN training jobs, increasing training batch size. In particular, GMLake uses CUDA support to coalesce multiple smaller physical memory pages into a single virtually contiguous object that can prevent out-of-memory errors. In contrast, \sysname targets optimizing inference workloads. 

\section{Conclusion}
\label{sec:conclusion}

In this paper, we propose \sysname for dynamic \kvcache memory management in LLM serving systems. The key highlight of \sysname is that it mitigates fragmentation in physical memory while retaining the contiguity of \kvcache in virtual memory. We present various examples to highlight that \sysname  reduces programming burden while improving portability and performance compared to the popular \pa approach. 

\begin{acks}
We thank our shepherd Sam Ainsworth, along with the anonymous ASPLOS reviewers, for their valuable feedback. We thank NVIDIA for useful discussions on CUDA VMM APIs and for sharing their experiences with \sysname. We also thank members of the open-source community (GitHub user names izhuhaoran and tongping) for trying to upstream the \sysname approach to \vllm~\cite{vattn-vllm-1,vattn-vllm-2}, and Zihao Ye for clarifying the role of register spilling in the runtime overhead of \pa.  Ajay Nayak is supported by the Prime Minister’s Fellowship Scheme for Doctoral Research, co-sponsored by the Confederation of Indian Industry, the Government of India, and Microsoft Research India.
\end{acks}

\bibliographystyle{ACM-Reference-Format}
\balance
\bibliography{main}


\begin{thebibliography}{83}


\ifx \showCODEN    \undefined \def \showCODEN     #1{\unskip}     \fi
\ifx \showDOI      \undefined \def \showDOI       #1{#1}\fi
\ifx \showISBNx    \undefined \def \showISBNx     #1{\unskip}     \fi
\ifx \showISBNxiii \undefined \def \showISBNxiii  #1{\unskip}     \fi
\ifx \showISSN     \undefined \def \showISSN      #1{\unskip}     \fi
\ifx \showLCCN     \undefined \def \showLCCN      #1{\unskip}     \fi
\ifx \shownote     \undefined \def \shownote      #1{#1}          \fi
\ifx \showarticletitle \undefined \def \showarticletitle #1{#1}   \fi
\ifx \showURL      \undefined \def \showURL       {\relax}        \fi
\providecommand\bibfield[2]{#2}
\providecommand\bibinfo[2]{#2}
\providecommand\natexlab[1]{#1}
\providecommand\showeprint[2][]{arXiv:#2}

\bibitem[fag(2022)]%
        {fagithub}
 \bibinfo{year}{2022}\natexlab{}.
\newblock \bibinfo{title}{{FlashAttention}}.
\newblock \bibinfo{howpublished}{\url{https://github.com/Dao-AILab/flash-attention}}.
\newblock


\bibitem[fla(2023)]%
        {flashdecoding}
 \bibinfo{year}{2023}\natexlab{}.
\newblock \bibinfo{title}{{Flash-Decoding for long-context inference}}.
\newblock \bibinfo{howpublished}{\url{https://crfm.stanford.edu/2023/10/12/flashdecoding.html}}.
\newblock


\bibitem[fig(2023)]%
        {figithub}
 \bibinfo{year}{2023}\natexlab{}.
\newblock \bibinfo{title}{{FlashInfer: Kernel Library for LLM Serving}}.
\newblock \bibinfo{howpublished}{\url{https://github.com/flashinfer-ai/flashinfer}}.
\newblock


\bibitem[lig(2023)]%
        {lightllm:github}
 \bibinfo{year}{2023}\natexlab{}.
\newblock \bibinfo{title}{LightLLM: A Light and Fast Inference Service for LLM}.
\newblock \bibinfo{howpublished}{\url{https://github.com/ModelTC/lightllm}}.
\newblock


\bibitem[ten(2023)]%
        {tensorrt-perf-decay}
 \bibinfo{year}{2023}\natexlab{}.
\newblock \bibinfo{title}{{Performance decay when using paged attention}}.
\newblock \bibinfo{howpublished}{\url{https://github.com/NVIDIA/TensorRT-LLM/issues/75}}.
\newblock


\bibitem[trt(2023)]%
        {trtllmgithub}
 \bibinfo{year}{2023}\natexlab{}.
\newblock \bibinfo{title}{TensorRT-LLM: A TensorRT Toolbox for Optimized Large Language Model Inference}.
\newblock \bibinfo{howpublished}{\url{https://github.com/NVIDIA/TensorRT-LLM}}.
\newblock


\bibitem[fa-(2023)]%
        {fa-paged-crash}
 \bibinfo{year}{2023}\natexlab{}.
\newblock \bibinfo{title}{{Use optimized kernels for MQA/GQA}}.
\newblock \bibinfo{howpublished}{\url{https://github.com/vllm-project/vllm/issues/1880}}.
\newblock


\bibitem[fas(2024a)]%
        {fasmall}
 \bibinfo{year}{2024}\natexlab{a}.
\newblock \bibinfo{title}{{Add support for small page sizes}}.
\newblock \bibinfo{howpublished}{\url{https://github.com/Dao-AILab/flash-attention/pull/824}}.
\newblock


\bibitem[ama(2024)]%
        {amazoncodewhisperer}
 \bibinfo{year}{2024}\natexlab{}.
\newblock \bibinfo{title}{Amazon CodeWhisperer}.
\newblock \bibinfo{howpublished}{\url{https://aws.amazon.com/codewhisperer/}}.
\newblock


\bibitem[bin(2024)]%
        {bingai}
 \bibinfo{year}{2024}\natexlab{}.
\newblock \bibinfo{title}{Bing AI}.
\newblock \bibinfo{howpublished}{\url{https://www.bing.com/chat}}.
\newblock


\bibitem[arx(2024)]%
        {arxiv}
 \bibinfo{year}{2024}\natexlab{}.
\newblock \bibinfo{title}{ccdv/arxiv-summarization}.
\newblock \bibinfo{howpublished}{\url{https://huggingface.co/datasets/ccdv/arxiv-summarization}}.
\newblock


\bibitem[cud(2024)]%
        {cudavirtualmemory}
 \bibinfo{year}{2024}\natexlab{}.
\newblock \bibinfo{title}{{CUDA Toolkit Documentation: Virtual Memory Management}}.
\newblock \bibinfo{howpublished}{\url{https://docs.nvidia.com/cuda/cuda-driver-api/group__CUDA__VA.html}}.
\newblock


\bibitem[vll(2024a)]%
        {vllm-copy-kernels}
 \bibinfo{year}{2024}\natexlab{a}.
\newblock \bibinfo{title}{{Custom CUDA kernels for KV cache copy operations}}.
\newblock \bibinfo{howpublished}{\url{https://github.com/vllm-project/vllm/blob/main/csrc/cache_kernels.cu}}.
\newblock


\bibitem[fi-(2024)]%
        {fi-strides}
 \bibinfo{year}{2024}\natexlab{}.
\newblock \bibinfo{title}{Custom strides to support non-contiguous KV cache}.
\newblock \bibinfo{howpublished}{\url{https://github.com/flashinfer-ai/flashinfer/commit/85b1878996a29814f674ee5000facb1e2e763d9a}}.
\newblock


\bibitem[fas(2024b)]%
        {fastertransformer}
 \bibinfo{year}{2024}\natexlab{b}.
\newblock \bibinfo{title}{{Faster Transformer}}.
\newblock \bibinfo{howpublished}{\url{https://github.com/NVIDIA/FasterTransformer}}.
\newblock


\bibitem[fa3(2024a)]%
        {fa3-vllm-integration-issue}
 \bibinfo{year}{2024}\natexlab{a}.
\newblock \bibinfo{title}{[Feature]: FlashAttention 3 support}.
\newblock \bibinfo{howpublished}{\url{https://github.com/vllm-project/vllm/issues/6348\#issuecomment-2540969988}}.
\newblock


\bibitem[vll(2024b)]%
        {vllm-blocktable-issue}
 \bibinfo{year}{2024}\natexlab{b}.
\newblock \bibinfo{title}{{Fix eager mode performance}}.
\newblock \bibinfo{howpublished}{\url{https://github.com/vllm-project/vllm/pull/2377}}.
\newblock


\bibitem[git(2024)]%
        {githubcopilot}
 \bibinfo{year}{2024}\natexlab{}.
\newblock \bibinfo{title}{Github Copilot}.
\newblock \bibinfo{howpublished}{\url{https://github.com/features/copilot}}.
\newblock


\bibitem[bar(2024)]%
        {bard}
 \bibinfo{year}{2024}\natexlab{}.
\newblock \bibinfo{title}{Google Bard}.
\newblock \bibinfo{howpublished}{\url{https://bard.google.com}}.
\newblock


\bibitem[fap(2024)]%
        {fapaging}
 \bibinfo{year}{2024}\natexlab{}.
\newblock \bibinfo{title}{{Implement Page KV Cache}}.
\newblock \bibinfo{howpublished}{\url{https://github.com/Dao-AILab/flash-attention/commit/54e80a3829c6d2337570d01e78ebd9529c02d342}}.
\newblock


\bibitem[lla(2024)]%
        {llama-8b-hf}
 \bibinfo{year}{2024}\natexlab{}.
\newblock \bibinfo{title}{{Meta-Llama-3-8B}}.
\newblock \bibinfo{howpublished}{\url{https://huggingface.co/meta-llama/Meta-Llama-3-8B}}.
\newblock


\bibitem[pas(2024)]%
        {pascalmmu}
 \bibinfo{year}{2024}\natexlab{}.
\newblock \bibinfo{title}{{Pascal MMU Format Changes}}.
\newblock \bibinfo{howpublished}{\url{https://nvidia.github.io/open-gpu-doc/pascal/gp100-mmu-format.pdf}}.
\newblock


\bibitem[vat(2024a)]%
        {vattn-vllm-2}
 \bibinfo{year}{2024}\natexlab{a}.
\newblock \bibinfo{title}{PoC of dAttention support (based on vAttention)}.
\newblock \bibinfo{howpublished}{\url{https://github.com/vllm-project/vllm/pull/9078}}.
\newblock


\bibitem[vll(2024c)]%
        {vllmfi1}
 \bibinfo{year}{2024}\natexlab{c}.
\newblock \bibinfo{title}{{Refactor Attention Take 2}}.
\newblock \bibinfo{howpublished}{\url{https://github.com/vllm-project/vllm/pull/3462}}.
\newblock


\bibitem[rep(2024)]%
        {replitghostwriter}
 \bibinfo{year}{2024}\natexlab{}.
\newblock \bibinfo{title}{Replit Ghostwriter}.
\newblock \bibinfo{howpublished}{\url{https://replit.com/site/ghostwriter}}.
\newblock


\bibitem[fa3(2024b)]%
        {fa3-vllm-integration-roadmap}
 \bibinfo{year}{2024}\natexlab{b}.
\newblock \bibinfo{title}{[Roadmap] vLLM Roadmap Q4 2024 \#9006}.
\newblock \bibinfo{howpublished}{\url{https://github.com/vllm-project/vllm/issues/9006\#issue-2559831134}}.
\newblock


\bibitem[vll(2024d)]%
        {vllmfi3}
 \bibinfo{year}{2024}\natexlab{d}.
\newblock \bibinfo{title}{{Separate attention backends}}.
\newblock \bibinfo{howpublished}{\url{https://github.com/vllm-project/vllm/pull/3005/}}.
\newblock


\bibitem[hft(2024)]%
        {hftgi}
 \bibinfo{year}{2024}\natexlab{}.
\newblock \bibinfo{title}{Text Generation Inference}.
\newblock \bibinfo{howpublished}{\url{https://huggingface.co/text-generation-inference}}.
\newblock


\bibitem[thu(2024)]%
        {thunderkittens}
 \bibinfo{year}{2024}\natexlab{}.
\newblock \bibinfo{title}{{Tile primitives for speedy kernels}}.
\newblock \bibinfo{howpublished}{\url{https://github.com/HazyResearch/ThunderKittens}}.
\newblock


\bibitem[vll(2024e)]%
        {vllmfi2}
 \bibinfo{year}{2024}\natexlab{e}.
\newblock \bibinfo{title}{{Use FlashInfer for Decoding}}.
\newblock \bibinfo{howpublished}{\url{https://github.com/vllm-project/vllm/pull/4353}}.
\newblock


\bibitem[vat(2024b)]%
        {vattn-vllm-1}
 \bibinfo{year}{2024}\natexlab{b}.
\newblock \bibinfo{title}{VMM KV cache for NVIDIA GPUs}.
\newblock \bibinfo{howpublished}{\url{https://github.com/vllm-project/vllm/pull/6102}}.
\newblock


\bibitem[yi-(2024a)]%
        {yi-34b-200k-hf}
 \bibinfo{year}{2024}\natexlab{a}.
\newblock \bibinfo{title}{{Yi-34B-200K}}.
\newblock \bibinfo{howpublished}{\url{https://huggingface.co/01-ai/Yi-34B-200K}}.
\newblock


\bibitem[yi-(2024b)]%
        {yi-6b-200k-hf}
 \bibinfo{year}{2024}\natexlab{b}.
\newblock \bibinfo{title}{{Yi-6B-200K}}.
\newblock \bibinfo{howpublished}{\url{https://huggingface.co/01-ai/Yi-6B-200K}}.
\newblock


\bibitem[Agrawal et~al\mbox{.}(2024a)]%
        {vidur}
\bibfield{author}{\bibinfo{person}{Amey Agrawal}, \bibinfo{person}{Nitin Kedia}, \bibinfo{person}{Jayashree Mohan}, \bibinfo{person}{Ashish Panwar}, \bibinfo{person}{Nipun Kwatra}, \bibinfo{person}{Bhargav~S Gulavani}, \bibinfo{person}{Ramachandran Ramjee}, {and} \bibinfo{person}{Alexey Tumanov}.} \bibinfo{year}{2024}\natexlab{a}.
\newblock \showarticletitle{Vidur: A Large-Scale Simulation Framework For LLM Inference}.
\newblock \bibinfo{journal}{\emph{Proceedings of The Seventh Annual Conference on Machine Learning and Systems, 2024, Santa Clara}} (\bibinfo{year}{2024}).
\newblock


\bibitem[Agrawal et~al\mbox{.}(2024b)]%
        {sarathiserve2024}
\bibfield{author}{\bibinfo{person}{Amey Agrawal}, \bibinfo{person}{Nitin Kedia}, \bibinfo{person}{Ashish Panwar}, \bibinfo{person}{Jayashree Mohan}, \bibinfo{person}{Nipun Kwatra}, \bibinfo{person}{Bhargav Gulavani}, \bibinfo{person}{Alexey Tumanov}, {and} \bibinfo{person}{Ramachandran Ramjee}.} \bibinfo{year}{2024}\natexlab{b}.
\newblock \showarticletitle{Taming {Throughput-Latency} Tradeoff in {LLM} Inference with {Sarathi-Serve}}. In \bibinfo{booktitle}{\emph{18th USENIX Symposium on Operating Systems Design and Implementation (OSDI 24)}}. \bibinfo{publisher}{USENIX Association}, \bibinfo{address}{Santa Clara, CA}, \bibinfo{pages}{117--134}.
\newblock
\showISBNx{978-1-939133-40-3}
\urldef\tempurl%
\url{https://www.usenix.org/conference/osdi24/presentation/agrawal}
\showURL{%
\tempurl}


\bibitem[Agrawal et~al\mbox{.}(2023)]%
        {sarathi2023}
\bibfield{author}{\bibinfo{person}{Amey Agrawal}, \bibinfo{person}{Ashish Panwar}, \bibinfo{person}{Jayashree Mohan}, \bibinfo{person}{Nipun Kwatra}, \bibinfo{person}{Bhargav~S. Gulavani}, {and} \bibinfo{person}{Ramachandran Ramjee}.} \bibinfo{year}{2023}\natexlab{}.
\newblock \bibinfo{title}{SARATHI: Efficient LLM Inference by Piggybacking Decodes with Chunked Prefills}.
\newblock
\newblock
\showeprint[arxiv]{2308.16369}~[cs.LG]


\bibitem[Ainslie et~al\mbox{.}(2023)]%
        {groupedqueryattention}
\bibfield{author}{\bibinfo{person}{Joshua Ainslie}, \bibinfo{person}{James Lee-Thorp}, \bibinfo{person}{Michiel de Jong}, \bibinfo{person}{Yury Zemlyanskiy}, \bibinfo{person}{Federico Lebrón}, {and} \bibinfo{person}{Sumit Sanghai}.} \bibinfo{year}{2023}\natexlab{}.
\newblock \bibinfo{title}{GQA: Training Generalized Multi-Query Transformer Models from Multi-Head Checkpoints}.
\newblock
\newblock
\showeprint[arxiv]{2305.13245}~[cs.CL]


\bibitem[Beltagy et~al\mbox{.}(2020)]%
        {longformer}
\bibfield{author}{\bibinfo{person}{Iz Beltagy}, \bibinfo{person}{Matthew~E. Peters}, {and} \bibinfo{person}{Arman Cohan}.} \bibinfo{year}{2020}\natexlab{}.
\newblock \bibinfo{title}{Longformer: The Long-Document Transformer}.
\newblock
\newblock
\showeprint[arxiv]{2004.05150}~[cs.CL]


\bibitem[Bikshandi and Shah(2023)]%
        {flashattnhopper}
\bibfield{author}{\bibinfo{person}{Ganesh Bikshandi} {and} \bibinfo{person}{Jay Shah}.} \bibinfo{year}{2023}\natexlab{}.
\newblock \bibinfo{title}{A Case Study in CUDA Kernel Fusion: Implementing FlashAttention-2 on NVIDIA Hopper Architecture using the CUTLASS Library}.
\newblock
\newblock
\showeprint[arxiv]{2312.11918}~[cs.LG]


\bibitem[Child et~al\mbox{.}(2019)]%
        {generatinglongsequence2019}
\bibfield{author}{\bibinfo{person}{Rewon Child}, \bibinfo{person}{Scott Gray}, \bibinfo{person}{Alec Radford}, {and} \bibinfo{person}{Ilya Sutskever}.} \bibinfo{year}{2019}\natexlab{}.
\newblock \bibinfo{title}{Generating Long Sequences with Sparse Transformers}.
\newblock
\newblock
\showeprint[arxiv]{1904.10509}~[cs.LG]


\bibitem[Chowdhery et~al\mbox{.}(2022)]%
        {chowdhery2022Palm}
\bibfield{author}{\bibinfo{person}{Aakanksha Chowdhery}, \bibinfo{person}{Sharan Narang}, \bibinfo{person}{Jacob Devlin}, \bibinfo{person}{Maarten Bosma}, \bibinfo{person}{Gaurav Mishra}, \bibinfo{person}{Adam Roberts}, \bibinfo{person}{Paul Barham}, \bibinfo{person}{Hyung~Won Chung}, \bibinfo{person}{Charles Sutton}, \bibinfo{person}{Sebastian Gehrmann}, \bibinfo{person}{Parker Schuh}, \bibinfo{person}{Kensen Shi}, \bibinfo{person}{Sasha Tsvyashchenko}, \bibinfo{person}{Joshua Maynez}, \bibinfo{person}{Abhishek Rao}, \bibinfo{person}{Parker Barnes}, \bibinfo{person}{Yi Tay}, \bibinfo{person}{Noam Shazeer}, \bibinfo{person}{Vinodkumar Prabhakaran}, \bibinfo{person}{Emily Reif}, \bibinfo{person}{Nan Du}, \bibinfo{person}{Ben Hutchinson}, \bibinfo{person}{Reiner Pope}, \bibinfo{person}{James Bradbury}, \bibinfo{person}{Jacob Austin}, \bibinfo{person}{Michael Isard}, \bibinfo{person}{Guy Gur{-}Ari}, \bibinfo{person}{Pengcheng Yin}, \bibinfo{person}{Toju Duke}, \bibinfo{person}{Anselm Levskaya},
  \bibinfo{person}{Sanjay Ghemawat}, \bibinfo{person}{Sunipa Dev}, \bibinfo{person}{Henryk Michalewski}, \bibinfo{person}{Xavier Garcia}, \bibinfo{person}{Vedant Misra}, \bibinfo{person}{Kevin Robinson}, \bibinfo{person}{Liam Fedus}, \bibinfo{person}{Denny Zhou}, \bibinfo{person}{Daphne Ippolito}, \bibinfo{person}{David Luan}, \bibinfo{person}{Hyeontaek Lim}, \bibinfo{person}{Barret Zoph}, \bibinfo{person}{Alexander Spiridonov}, \bibinfo{person}{Ryan Sepassi}, \bibinfo{person}{David Dohan}, \bibinfo{person}{Shivani Agrawal}, \bibinfo{person}{Mark Omernick}, \bibinfo{person}{Andrew~M. Dai}, \bibinfo{person}{Thanumalayan~Sankaranarayana Pillai}, \bibinfo{person}{Marie Pellat}, \bibinfo{person}{Aitor Lewkowycz}, \bibinfo{person}{Erica Moreira}, \bibinfo{person}{Rewon Child}, \bibinfo{person}{Oleksandr Polozov}, \bibinfo{person}{Katherine Lee}, \bibinfo{person}{Zongwei Zhou}, \bibinfo{person}{Xuezhi Wang}, \bibinfo{person}{Brennan Saeta}, \bibinfo{person}{Mark Diaz}, \bibinfo{person}{Orhan Firat},
  \bibinfo{person}{Michele Catasta}, \bibinfo{person}{Jason Wei}, \bibinfo{person}{Kathy Meier{-}Hellstern}, \bibinfo{person}{Douglas Eck}, \bibinfo{person}{Jeff Dean}, \bibinfo{person}{Slav Petrov}, {and} \bibinfo{person}{Noah Fiedel}.} \bibinfo{year}{2022}\natexlab{}.
\newblock \showarticletitle{PaLM: Scaling Language Modeling with Pathways}.
\newblock \bibinfo{journal}{\emph{CoRR}}  \bibinfo{volume}{abs/2204.02311} (\bibinfo{year}{2022}).
\newblock
\urldef\tempurl%
\url{https://doi.org/10.48550/arXiv.2204.02311}
\showDOI{\tempurl}
\showeprint[arXiv]{2204.02311}


\bibitem[Dao(2023)]%
        {flashattention2}
\bibfield{author}{\bibinfo{person}{Tri Dao}.} \bibinfo{year}{2023}\natexlab{}.
\newblock \bibinfo{title}{FlashAttention-2: Faster Attention with Better Parallelism and Work Partitioning}.
\newblock
\newblock
\showeprint[arxiv]{2307.08691}~[cs.LG]


\bibitem[Dao et~al\mbox{.}(2024)]%
        {flashattention}
\bibfield{author}{\bibinfo{person}{Tri Dao}, \bibinfo{person}{Daniel~Y. Fu}, \bibinfo{person}{Stefano Ermon}, \bibinfo{person}{Atri Rudra}, {and} \bibinfo{person}{Christopher R\'{e}}.} \bibinfo{year}{2024}\natexlab{}.
\newblock \showarticletitle{FLASHATTENTION: fast and memory-efficient exact attention with IO-awareness}. In \bibinfo{booktitle}{\emph{Proceedings of the 36th International Conference on Neural Information Processing Systems}} (New Orleans, LA, USA) \emph{(\bibinfo{series}{NIPS '22})}. \bibinfo{publisher}{Curran Associates Inc.}, \bibinfo{address}{Red Hook, NY, USA}, Article \bibinfo{articleno}{1189}, \bibinfo{numpages}{16}~pages.
\newblock
\showISBNx{9781713871088}


\bibitem[Griggs et~al\mbox{.}(2024)]%
        {griggs2024melangecostefficientlarge}
\bibfield{author}{\bibinfo{person}{Tyler Griggs}, \bibinfo{person}{Xiaoxuan Liu}, \bibinfo{person}{Jiaxiang Yu}, \bibinfo{person}{Doyoung Kim}, \bibinfo{person}{Wei-Lin Chiang}, \bibinfo{person}{Alvin Cheung}, {and} \bibinfo{person}{Ion Stoica}.} \bibinfo{year}{2024}\natexlab{}.
\newblock \bibinfo{title}{M\'elange: Cost Efficient Large Language Model Serving by Exploiting GPU Heterogeneity}.
\newblock
\newblock
\showeprint[arxiv]{2404.14527}~[cs.DC]
\urldef\tempurl%
\url{https://arxiv.org/abs/2404.14527}
\showURL{%
\tempurl}


\bibitem[Guo et~al\mbox{.}(2024)]%
        {gmlakeasplos24}
\bibfield{author}{\bibinfo{person}{Cong Guo}, \bibinfo{person}{Rui Zhang}, \bibinfo{person}{Jiale Xu}, \bibinfo{person}{Jingwen Leng}, \bibinfo{person}{Zihan Liu}, \bibinfo{person}{Ziyu Huang}, \bibinfo{person}{Minyi Guo}, \bibinfo{person}{Hao Wu}, \bibinfo{person}{Shouren Zhao}, \bibinfo{person}{Junping Zhao}, {and} \bibinfo{person}{Ke Zhang}.} \bibinfo{year}{2024}\natexlab{}.
\newblock \showarticletitle{GMLake: Efficient and Transparent GPU Memory Defragmentation for Large-scale DNN Training with Virtual Memory Stitching}. In \bibinfo{booktitle}{\emph{Proceedings of the 29th ACM International Conference on Architectural Support for Programming Languages and Operating Systems, Volume 2}} (La Jolla, CA, USA) \emph{(\bibinfo{series}{ASPLOS '24})}. \bibinfo{publisher}{Association for Computing Machinery}, \bibinfo{address}{New York, NY, USA}, \bibinfo{pages}{450–466}.
\newblock
\showISBNx{9798400703850}
\urldef\tempurl%
\url{https://doi.org/10.1145/3620665.3640423}
\showDOI{\tempurl}


\bibitem[Holmes et~al\mbox{.}(2024)]%
        {splitfuse2024}
\bibfield{author}{\bibinfo{person}{Connor Holmes}, \bibinfo{person}{Masahiro Tanaka}, \bibinfo{person}{Michael Wyatt}, \bibinfo{person}{Ammar~Ahmad Awan}, \bibinfo{person}{Jeff Rasley}, \bibinfo{person}{Samyam Rajbhandari}, \bibinfo{person}{Reza~Yazdani Aminabadi}, \bibinfo{person}{Heyang Qin}, \bibinfo{person}{Arash Bakhtiari}, \bibinfo{person}{Lev Kurilenko}, {and} \bibinfo{person}{Yuxiong He}.} \bibinfo{year}{2024}\natexlab{}.
\newblock \bibinfo{title}{DeepSpeed-FastGen: High-throughput Text Generation for LLMs via MII and DeepSpeed-Inference}.
\newblock
\newblock
\showeprint[arxiv]{2401.08671}~[cs.PF]


\bibitem[Hong et~al\mbox{.}(2024)]%
        {flashdecoding++}
\bibfield{author}{\bibinfo{person}{Ke Hong}, \bibinfo{person}{Guohao Dai}, \bibinfo{person}{Jiaming Xu}, \bibinfo{person}{Qiuli Mao}, \bibinfo{person}{Xiuhong Li}, \bibinfo{person}{Jun Liu}, \bibinfo{person}{kangdi chen}, \bibinfo{person}{Yuhan Dong}, {and} \bibinfo{person}{Yu Wang}.} \bibinfo{year}{2024}\natexlab{}.
\newblock \showarticletitle{FlashDecoding++: Faster Large Language Model Inference with Asynchronization, Flat GEMM Optimization, and Heuristics}. In \bibinfo{booktitle}{\emph{Proceedings of Machine Learning and Systems}}, \bibfield{editor}{\bibinfo{person}{P.~Gibbons}, \bibinfo{person}{G.~Pekhimenko}, {and} \bibinfo{person}{C.~De Sa}} (Eds.), Vol.~\bibinfo{volume}{6}. \bibinfo{pages}{148--161}.
\newblock
\urldef\tempurl%
\url{https://proceedings.mlsys.org/paper_files/paper/2024/file/5321b1dabcd2be188d796c21b733e8c7-Paper-Conference.pdf}
\showURL{%
\tempurl}


\bibitem[Hu et~al\mbox{.}(2024)]%
        {tetriinfer}
\bibfield{author}{\bibinfo{person}{Cunchen Hu}, \bibinfo{person}{Heyang Huang}, \bibinfo{person}{Liangliang Xu}, \bibinfo{person}{Xusheng Chen}, \bibinfo{person}{Jiang Xu}, \bibinfo{person}{Shuang Chen}, \bibinfo{person}{Hao Feng}, \bibinfo{person}{Chenxi Wang}, \bibinfo{person}{Sa Wang}, \bibinfo{person}{Yungang Bao}, {et~al\mbox{.}}} \bibinfo{year}{2024}\natexlab{}.
\newblock \showarticletitle{Inference without Interference: Disaggregate LLM Inference for Mixed Downstream Workloads}.
\newblock \bibinfo{journal}{\emph{arXiv preprint arXiv:2401.11181}} (\bibinfo{year}{2024}).
\newblock


\bibitem[Kamath et~al\mbox{.}(2024)]%
        {pod-attn-2024}
\bibfield{author}{\bibinfo{person}{Aditya~K Kamath}, \bibinfo{person}{Ramya Prabhu}, \bibinfo{person}{Jayashree Mohan}, \bibinfo{person}{Simon Peter}, \bibinfo{person}{Ramachandran Ramjee}, {and} \bibinfo{person}{Ashish Panwar}.} \bibinfo{year}{2024}\natexlab{}.
\newblock \bibinfo{title}{POD-Attention: Unlocking Full Prefill-Decode Overlap for Faster LLM Inference}.
\newblock
\newblock
\showeprint[arxiv]{2410.18038}~[cs.LG]
\urldef\tempurl%
\url{https://arxiv.org/abs/2410.18038}
\showURL{%
\tempurl}


\bibitem[Kossmann et~al\mbox{.}(2024)]%
        {kossmann2024gpuhalfemptyhalffullpractical}
\bibfield{author}{\bibinfo{person}{Ferdi Kossmann}, \bibinfo{person}{Bruce Fontaine}, \bibinfo{person}{Daya Khudia}, \bibinfo{person}{Michael Cafarella}, {and} \bibinfo{person}{Samuel Madden}.} \bibinfo{year}{2024}\natexlab{}.
\newblock \bibinfo{title}{Is the GPU Half-Empty or Half-Full? Practical Scheduling Techniques for LLMs}.
\newblock
\newblock
\showeprint[arxiv]{2410.17840}~[cs.LG]
\urldef\tempurl%
\url{https://arxiv.org/abs/2410.17840}
\showURL{%
\tempurl}


\bibitem[Kwon et~al\mbox{.}(2023)]%
        {vllmsosp}
\bibfield{author}{\bibinfo{person}{Woosuk Kwon}, \bibinfo{person}{Zhuohan Li}, \bibinfo{person}{Siyuan Zhuang}, \bibinfo{person}{Ying Sheng}, \bibinfo{person}{Lianmin Zheng}, \bibinfo{person}{Cody~Hao Yu}, \bibinfo{person}{Joseph Gonzalez}, \bibinfo{person}{Hao Zhang}, {and} \bibinfo{person}{Ion Stoica}.} \bibinfo{year}{2023}\natexlab{}.
\newblock \showarticletitle{Efficient Memory Management for Large Language Model Serving with PagedAttention}. In \bibinfo{booktitle}{\emph{Proceedings of the 29th Symposium on Operating Systems Principles}} (Koblenz, Germany) \emph{(\bibinfo{series}{SOSP '23})}. \bibinfo{publisher}{Association for Computing Machinery}, \bibinfo{address}{New York, NY, USA}, \bibinfo{pages}{611–626}.
\newblock
\showISBNx{9798400702297}
\urldef\tempurl%
\url{https://doi.org/10.1145/3600006.3613165}
\showDOI{\tempurl}


\bibitem[Kwon et~al\mbox{.}(2016)]%
        {ingens}
\bibfield{author}{\bibinfo{person}{Youngjin Kwon}, \bibinfo{person}{Hangchen Yu}, \bibinfo{person}{Simon Peter}, \bibinfo{person}{Christopher~J. Rossbach}, {and} \bibinfo{person}{Emmett Witchel}.} \bibinfo{year}{2016}\natexlab{}.
\newblock \showarticletitle{Coordinated and Efficient Huge Page Management with Ingens}. In \bibinfo{booktitle}{\emph{12th USENIX Symposium on Operating Systems Design and Implementation (OSDI 16)}}. \bibinfo{publisher}{USENIX Association}, \bibinfo{address}{Savannah, GA}, \bibinfo{pages}{705--721}.
\newblock
\showISBNx{978-1-931971-33-1}
\urldef\tempurl%
\url{https://www.usenix.org/conference/osdi16/technical-sessions/presentation/kwon}
\showURL{%
\tempurl}


\bibitem[Lee et~al\mbox{.}(2024)]%
        {lee2024infinigenefficientgenerativeinference}
\bibfield{author}{\bibinfo{person}{Wonbeom Lee}, \bibinfo{person}{Jungi Lee}, \bibinfo{person}{Junghwan Seo}, {and} \bibinfo{person}{Jaewoong Sim}.} \bibinfo{year}{2024}\natexlab{}.
\newblock \bibinfo{title}{InfiniGen: Efficient Generative Inference of Large Language Models with Dynamic KV Cache Management}.
\newblock
\newblock
\showeprint[arxiv]{2406.19707}~[cs.LG]
\urldef\tempurl%
\url{https://arxiv.org/abs/2406.19707}
\showURL{%
\tempurl}


\bibitem[Ma et~al\mbox{.}(2024)]%
        {ma2024compressingkvcachelongcontext}
\bibfield{author}{\bibinfo{person}{Da Ma}, \bibinfo{person}{Lu Chen}, \bibinfo{person}{Situo Zhang}, \bibinfo{person}{Yuxun Miao}, \bibinfo{person}{Su Zhu}, \bibinfo{person}{Zhi Chen}, \bibinfo{person}{Hongshen Xu}, \bibinfo{person}{Hanqi Li}, \bibinfo{person}{Shuai Fan}, \bibinfo{person}{Lei Pan}, {and} \bibinfo{person}{Kai Yu}.} \bibinfo{year}{2024}\natexlab{}.
\newblock \bibinfo{title}{Compressing KV Cache for Long-Context LLM Inference with Inter-Layer Attention Similarity}.
\newblock
\newblock
\showeprint[arxiv]{2412.02252}~[cs.CL]
\urldef\tempurl%
\url{https://arxiv.org/abs/2412.02252}
\showURL{%
\tempurl}


\bibitem[Mahmud et~al\mbox{.}(2024)]%
        {mahmud2024mapleframeworkactivepreference}
\bibfield{author}{\bibinfo{person}{Saaduddin Mahmud}, \bibinfo{person}{Mason Nakamura}, {and} \bibinfo{person}{Shlomo Zilberstein}.} \bibinfo{year}{2024}\natexlab{}.
\newblock \bibinfo{title}{MAPLE: A Framework for Active Preference Learning Guided by Large Language Models}.
\newblock
\newblock
\showeprint[arxiv]{2412.07207}~[cs.LG]
\urldef\tempurl%
\url{https://arxiv.org/abs/2412.07207}
\showURL{%
\tempurl}


\bibitem[Mei et~al\mbox{.}(2024)]%
        {helix}
\bibfield{author}{\bibinfo{person}{Yixuan Mei}, \bibinfo{person}{Yonghao Zhuang}, \bibinfo{person}{Xupeng Miao}, \bibinfo{person}{Juncheng Yang}, \bibinfo{person}{Zhihao Jia}, {and} \bibinfo{person}{Rashmi Vinayak}.} \bibinfo{year}{2024}\natexlab{}.
\newblock \bibinfo{title}{Helix: Distributed Serving of Large Language Models via Max-Flow on Heterogeneous GPUs}.
\newblock
\newblock
\showeprint[arxiv]{2406.01566}~[cs.LG]
\urldef\tempurl%
\url{https://arxiv.org/abs/2406.01566}
\showURL{%
\tempurl}


\bibitem[Miao et~al\mbox{.}(2023)]%
        {miao2023spotserveservinggenerativelarge}
\bibfield{author}{\bibinfo{person}{Xupeng Miao}, \bibinfo{person}{Chunan Shi}, \bibinfo{person}{Jiangfei Duan}, \bibinfo{person}{Xiaoli Xi}, \bibinfo{person}{Dahua Lin}, \bibinfo{person}{Bin Cui}, {and} \bibinfo{person}{Zhihao Jia}.} \bibinfo{year}{2023}\natexlab{}.
\newblock \bibinfo{title}{SpotServe: Serving Generative Large Language Models on Preemptible Instances}.
\newblock
\newblock
\showeprint[arxiv]{2311.15566}~[cs.DC]
\urldef\tempurl%
\url{https://arxiv.org/abs/2311.15566}
\showURL{%
\tempurl}


\bibitem[Nayak et~al\mbox{.}(2021)]%
        {gpu-mismanaged}
\bibfield{author}{\bibinfo{person}{Ajay Nayak}, \bibinfo{person}{Pratheek B.}, \bibinfo{person}{Vinod Ganapathy}, {and} \bibinfo{person}{Arkaprava Basu}.} \bibinfo{year}{2021}\natexlab{}.
\newblock \showarticletitle{(Mis)managed: A Novel TLB-based Covert Channel on GPUs}. In \bibinfo{booktitle}{\emph{Proceedings of the 2021 ACM Asia Conference on Computer and Communications Security}} (Virtual Event, Hong Kong) \emph{(\bibinfo{series}{ASIA CCS '21})}. \bibinfo{publisher}{Association for Computing Machinery}, \bibinfo{address}{New York, NY, USA}, \bibinfo{pages}{872–885}.
\newblock
\showISBNx{9781450382878}
\urldef\tempurl%
\url{https://doi.org/10.1145/3433210.3453077}
\showDOI{\tempurl}


\bibitem[Nicely and NVIDIA(2024)]%
        {cudnn-9}
\bibfield{author}{\bibinfo{person}{Matthew Nicely} {and} \bibinfo{person}{NVIDIA}.} \bibinfo{year}{2024}\natexlab{}.
\newblock \bibinfo{title}{Accelerating Transformers with NVIDIA cuDNN 9}.
\newblock
\newblock
\urldef\tempurl%
\url{https://developer.nvidia.com/blog/accelerating-transformers-with-nvidia-cudnn-9/}
\showURL{%
\tempurl}


\bibitem[NVIDIA(2024)]%
        {nvidia-cuda}
\bibfield{author}{\bibinfo{person}{NVIDIA}.} \bibinfo{year}{2024}\natexlab{}.
\newblock \bibinfo{title}{CUDA C++ Programming Guide}.
\newblock
\newblock
\urldef\tempurl%
\url{https://docs.nvidia.com/cuda/cuda-c-programming-guide/index.html}
\showURL{%
\tempurl}


\bibitem[OpenAI(2023)]%
        {openai2022gpt4techreport}
\bibfield{author}{\bibinfo{person}{OpenAI}.} \bibinfo{year}{2023}\natexlab{}.
\newblock \showarticletitle{{GPT-4} Technical Report}.
\newblock \bibinfo{journal}{\emph{CoRR}}  \bibinfo{volume}{abs/2303.08774} (\bibinfo{year}{2023}).
\newblock
\urldef\tempurl%
\url{https://doi.org/10.48550/arXiv.2303.08774}
\showDOI{\tempurl}
\showeprint[arXiv]{2303.08774}


\bibitem[Panwar et~al\mbox{.}(2019)]%
        {hawkeye}
\bibfield{author}{\bibinfo{person}{Ashish Panwar}, \bibinfo{person}{Sorav Bansal}, {and} \bibinfo{person}{K. Gopinath}.} \bibinfo{year}{2019}\natexlab{}.
\newblock \showarticletitle{HawkEye: Efficient Fine-Grained OS Support for Huge Pages}. In \bibinfo{booktitle}{\emph{Proceedings of the Twenty-Fourth International Conference on Architectural Support for Programming Languages and Operating Systems}} (Providence, RI, USA) \emph{(\bibinfo{series}{ASPLOS '19})}. \bibinfo{publisher}{Association for Computing Machinery}, \bibinfo{address}{New York, NY, USA}, \bibinfo{pages}{347–360}.
\newblock
\showISBNx{9781450362405}
\urldef\tempurl%
\url{https://doi.org/10.1145/3297858.3304064}
\showDOI{\tempurl}


\bibitem[Patel et~al\mbox{.}(2024)]%
        {patel2023splitwise}
\bibfield{author}{\bibinfo{person}{Pratyush Patel}, \bibinfo{person}{Esha Choukse}, \bibinfo{person}{Chaojie Zhang}, \bibinfo{person}{Aashaka Shah}, \bibinfo{person}{Íñigo Goiri}, \bibinfo{person}{Saeed Maleki}, {and} \bibinfo{person}{Ricardo Bianchini}.} \bibinfo{year}{2024}\natexlab{}.
\newblock \showarticletitle{Splitwise: Efficient Generative LLM Inference Using Phase Splitting}. In \bibinfo{booktitle}{\emph{2024 ACM/IEEE 51st Annual International Symposium on Computer Architecture (ISCA)}}. \bibinfo{pages}{118--132}.
\newblock
\urldef\tempurl%
\url{https://doi.org/10.1109/ISCA59077.2024.00019}
\showDOI{\tempurl}


\bibitem[Pichai et~al\mbox{.}(2014)]%
        {architecturalsupportforgpuat}
\bibfield{author}{\bibinfo{person}{Bharath Pichai}, \bibinfo{person}{Lisa Hsu}, {and} \bibinfo{person}{Abhishek Bhattacharjee}.} \bibinfo{year}{2014}\natexlab{}.
\newblock \showarticletitle{Architectural support for address translation on GPUs: designing memory management units for CPU/GPUs with unified address spaces}. In \bibinfo{booktitle}{\emph{Proceedings of the 19th International Conference on Architectural Support for Programming Languages and Operating Systems}} (Salt Lake City, Utah, USA) \emph{(\bibinfo{series}{ASPLOS '14})}. \bibinfo{publisher}{Association for Computing Machinery}, \bibinfo{address}{New York, NY, USA}, \bibinfo{pages}{743–758}.
\newblock
\showISBNx{9781450323055}
\urldef\tempurl%
\url{https://doi.org/10.1145/2541940.2541942}
\showDOI{\tempurl}


\bibitem[Pratheek et~al\mbox{.}(2022)]%
        {mcmgpus}
\bibfield{author}{\bibinfo{person}{B Pratheek}, \bibinfo{person}{Neha Jawalkar}, {and} \bibinfo{person}{Arkaprava Basu}.} \bibinfo{year}{2022}\natexlab{}.
\newblock \showarticletitle{Designing Virtual Memory System of MCM GPUs}. In \bibinfo{booktitle}{\emph{2022 55th IEEE/ACM International Symposium on Microarchitecture (MICRO)}}. \bibinfo{pages}{404--422}.
\newblock
\urldef\tempurl%
\url{https://doi.org/10.1109/MICRO56248.2022.00036}
\showDOI{\tempurl}


\bibitem[Sanovar et~al\mbox{.}(2024)]%
        {leanattention}
\bibfield{author}{\bibinfo{person}{Rya Sanovar}, \bibinfo{person}{Srikant Bharadwaj}, \bibinfo{person}{Renee~St. Amant}, \bibinfo{person}{Victor Rühle}, {and} \bibinfo{person}{Saravan Rajmohan}.} \bibinfo{year}{2024}\natexlab{}.
\newblock \bibinfo{title}{Lean Attention: Hardware-Aware Scalable Attention Mechanism for the Decode-Phase of Transformers}.
\newblock
\newblock
\showeprint[arxiv]{2405.10480}~[cs.AR]
\urldef\tempurl%
\url{https://arxiv.org/abs/2405.10480}
\showURL{%
\tempurl}


\bibitem[Shah et~al\mbox{.}(2024)]%
        {flash-attention-3}
\bibfield{author}{\bibinfo{person}{Jay Shah}, \bibinfo{person}{Ganesh Bikshandi}, \bibinfo{person}{Ying Zhang}, \bibinfo{person}{Vijay Thakkar}, \bibinfo{person}{Pradeep Ramani}, {and} \bibinfo{person}{Tri Dao}.} \bibinfo{year}{2024}\natexlab{}.
\newblock \showarticletitle{FlashAttention-3: Fast and Accurate Attention with Asynchrony and Low-precision}.
\newblock  (\bibinfo{year}{2024}).
\newblock


\bibitem[Shazeer(2019)]%
        {multiqueryattention}
\bibfield{author}{\bibinfo{person}{Noam Shazeer}.} \bibinfo{year}{2019}\natexlab{}.
\newblock \bibinfo{title}{Fast Transformer Decoding: One Write-Head is All You Need}.
\newblock
\newblock
\showeprint[arxiv]{1911.02150}~[cs.NE]


\bibitem[Sheng et~al\mbox{.}(2023)]%
        {flexgen}
\bibfield{author}{\bibinfo{person}{Ying Sheng}, \bibinfo{person}{Lianmin Zheng}, \bibinfo{person}{Binhang Yuan}, \bibinfo{person}{Zhuohan Li}, \bibinfo{person}{Max Ryabinin}, \bibinfo{person}{Beidi Chen}, \bibinfo{person}{Percy Liang}, \bibinfo{person}{Christopher R\'{e}}, \bibinfo{person}{Ion Stoica}, {and} \bibinfo{person}{Ce Zhang}.} \bibinfo{year}{2023}\natexlab{}.
\newblock \showarticletitle{FlexGen: high-throughput generative inference of large language models with a single GPU}. In \bibinfo{booktitle}{\emph{Proceedings of the 40th International Conference on Machine Learning}} (Honolulu, Hawaii, USA) \emph{(\bibinfo{series}{ICML'23})}. \bibinfo{publisher}{JMLR.org}, Article \bibinfo{articleno}{1288}, \bibinfo{numpages}{23}~pages.
\newblock


\bibitem[Shin et~al\mbox{.}(2018)]%
        {neighbourhoodawareat}
\bibfield{author}{\bibinfo{person}{Seunghee Shin}, \bibinfo{person}{Michael LeBeane}, \bibinfo{person}{Yan Solihin}, {and} \bibinfo{person}{Arkaprava Basu}.} \bibinfo{year}{2018}\natexlab{}.
\newblock \showarticletitle{Neighborhood-Aware Address Translation for Irregular GPU Applications}. In \bibinfo{booktitle}{\emph{2018 51st Annual IEEE/ACM International Symposium on Microarchitecture (MICRO)}}. \bibinfo{pages}{352--363}.
\newblock
\urldef\tempurl%
\url{https://doi.org/10.1109/MICRO.2018.00036}
\showDOI{\tempurl}


\bibitem[Singhania et~al\mbox{.}(2024)]%
        {singhania2024lokilowrankkeysefficient}
\bibfield{author}{\bibinfo{person}{Prajwal Singhania}, \bibinfo{person}{Siddharth Singh}, \bibinfo{person}{Shwai He}, \bibinfo{person}{Soheil Feizi}, {and} \bibinfo{person}{Abhinav Bhatele}.} \bibinfo{year}{2024}\natexlab{}.
\newblock \bibinfo{title}{Loki: Low-rank Keys for Efficient Sparse Attention}.
\newblock
\newblock
\showeprint[arxiv]{2406.02542}~[cs.LG]
\urldef\tempurl%
\url{https://arxiv.org/abs/2406.02542}
\showURL{%
\tempurl}


\bibitem[Strati et~al\mbox{.}(2024)]%
        {strati2024dejavukvcachestreamingfast}
\bibfield{author}{\bibinfo{person}{Foteini Strati}, \bibinfo{person}{Sara Mcallister}, \bibinfo{person}{Amar Phanishayee}, \bibinfo{person}{Jakub Tarnawski}, {and} \bibinfo{person}{Ana Klimovic}.} \bibinfo{year}{2024}\natexlab{}.
\newblock \bibinfo{title}{D\'ej\`aVu: KV-cache Streaming for Fast, Fault-tolerant Generative LLM Serving}.
\newblock
\newblock
\showeprint[arxiv]{2403.01876}~[cs.DC]
\urldef\tempurl%
\url{https://arxiv.org/abs/2403.01876}
\showURL{%
\tempurl}


\bibitem[Vaswani et~al\mbox{.}(2017)]%
        {attentionpaper}
\bibfield{author}{\bibinfo{person}{Ashish Vaswani}, \bibinfo{person}{Noam Shazeer}, \bibinfo{person}{Niki Parmar}, \bibinfo{person}{Jakob Uszkoreit}, \bibinfo{person}{Llion Jones}, \bibinfo{person}{Aidan~N Gomez}, \bibinfo{person}{\L~ukasz Kaiser}, {and} \bibinfo{person}{Illia Polosukhin}.} \bibinfo{year}{2017}\natexlab{}.
\newblock \showarticletitle{Attention is All you Need}. In \bibinfo{booktitle}{\emph{Advances in Neural Information Processing Systems}}, \bibfield{editor}{\bibinfo{person}{I.~Guyon}, \bibinfo{person}{U.~Von Luxburg}, \bibinfo{person}{S.~Bengio}, \bibinfo{person}{H.~Wallach}, \bibinfo{person}{R.~Fergus}, \bibinfo{person}{S.~Vishwanathan}, {and} \bibinfo{person}{R.~Garnett}} (Eds.), Vol.~\bibinfo{volume}{30}. \bibinfo{publisher}{Curran Associates, Inc.}
\newblock
\urldef\tempurl%
\url{https://proceedings.neurips.cc/paper_files/paper/2017/file/3f5ee243547dee91fbd053c1c4a845aa-Paper.pdf}
\showURL{%
\tempurl}


\bibitem[Wang et~al\mbox{.}(2023)]%
        {wang2023openchat}
\bibfield{author}{\bibinfo{person}{Guan Wang}, \bibinfo{person}{Sijie Cheng}, \bibinfo{person}{Xianyuan Zhan}, \bibinfo{person}{Xiangang Li}, \bibinfo{person}{Sen Song}, {and} \bibinfo{person}{Yang Liu}.} \bibinfo{year}{2023}\natexlab{}.
\newblock \bibinfo{title}{OpenChat: Advancing Open-source Language Models with Mixed-Quality Data}.
\newblock
\newblock
\showeprint[arxiv]{2309.11235}~[cs.CL]


\bibitem[Wu et~al\mbox{.}(2023)]%
        {fastserve}
\bibfield{author}{\bibinfo{person}{Bingyang Wu}, \bibinfo{person}{Yinmin Zhong}, \bibinfo{person}{Zili Zhang}, \bibinfo{person}{Gang Huang}, \bibinfo{person}{Xuanzhe Liu}, {and} \bibinfo{person}{Xin Jin}.} \bibinfo{year}{2023}\natexlab{}.
\newblock \bibinfo{title}{Fast Distributed Inference Serving for Large Language Models}.
\newblock
\newblock
\showeprint[arxiv]{2305.05920}~[cs.LG]


\bibitem[Wu et~al\mbox{.}(2024)]%
        {mirage-tensor-compiler}
\bibfield{author}{\bibinfo{person}{Mengdi Wu}, \bibinfo{person}{Xinhao Cheng}, \bibinfo{person}{Oded Padon}, {and} \bibinfo{person}{Zhihao Jia}.} \bibinfo{year}{2024}\natexlab{}.
\newblock \bibinfo{title}{A Multi-Level Superoptimizer for Tensor Programs}.
\newblock
\newblock
\showeprint[arxiv]{2405.05751}


\bibitem[Ye et~al\mbox{.}(2024)]%
        {flashinfer}
\bibfield{author}{\bibinfo{person}{Zihao Ye}, \bibinfo{person}{Lequn Chen}, \bibinfo{person}{Ruihang Lai}, \bibinfo{person}{Yilong Zhao}, \bibinfo{person}{Size Zheng}, \bibinfo{person}{Junru Shao}, \bibinfo{person}{Bohan Hou}, \bibinfo{person}{Hongyi Jin}, \bibinfo{person}{Yifei Zuo}, \bibinfo{person}{Liangsheng Yin}, \bibinfo{person}{Tianqi Chen}, {and} \bibinfo{person}{Luis Ceze}.} \bibinfo{year}{2024}\natexlab{}.
\newblock \bibinfo{title}{Accelerating Self-Attentions for LLM Serving with FlashInfer}.
\newblock
\newblock
\urldef\tempurl%
\url{https://flashinfer.ai/2024/02/02/introduce-flashinfer.html}
\showURL{%
\tempurl}


\bibitem[Yu et~al\mbox{.}(2022)]%
        {orca}
\bibfield{author}{\bibinfo{person}{Gyeong-In Yu}, \bibinfo{person}{Joo~Seong Jeong}, \bibinfo{person}{Geon-Woo Kim}, \bibinfo{person}{Soojeong Kim}, {and} \bibinfo{person}{Byung-Gon Chun}.} \bibinfo{year}{2022}\natexlab{}.
\newblock \showarticletitle{Orca: A Distributed Serving System for {Transformer-Based} Generative Models}. In \bibinfo{booktitle}{\emph{16th USENIX Symposium on Operating Systems Design and Implementation (OSDI 22)}}. \bibinfo{publisher}{USENIX Association}, \bibinfo{address}{Carlsbad, CA}, \bibinfo{pages}{521--538}.
\newblock
\showISBNx{978-1-939133-28-1}
\urldef\tempurl%
\url{https://www.usenix.org/conference/osdi22/presentation/yu}
\showURL{%
\tempurl}


\bibitem[Zhang et~al\mbox{.}(2023a)]%
        {gpu-reverse-engg-tlb1}
\bibfield{author}{\bibinfo{person}{Zhenkai Zhang}, \bibinfo{person}{Tyler Allen}, \bibinfo{person}{Fan Yao}, \bibinfo{person}{Xing Gao}, {and} \bibinfo{person}{Rong Ge}.} \bibinfo{year}{2023}\natexlab{a}.
\newblock \showarticletitle{TunneLs for Bootlegging: Fully Reverse-Engineering GPU TLBs for Challenging Isolation Guarantees of NVIDIA MIG}. In \bibinfo{booktitle}{\emph{Proceedings of the 2023 ACM SIGSAC Conference on Computer and Communications Security}} (Copenhagen, Denmark) \emph{(\bibinfo{series}{CCS '23})}. \bibinfo{publisher}{Association for Computing Machinery}, \bibinfo{address}{New York, NY, USA}, \bibinfo{pages}{960–974}.
\newblock
\showISBNx{9798400700507}
\urldef\tempurl%
\url{https://doi.org/10.1145/3576915.3616672}
\showDOI{\tempurl}


\bibitem[Zhang et~al\mbox{.}(2023b)]%
        {h2o2023}
\bibfield{author}{\bibinfo{person}{Zhenyu Zhang}, \bibinfo{person}{Ying Sheng}, \bibinfo{person}{Tianyi Zhou}, \bibinfo{person}{Tianlong Chen}, \bibinfo{person}{Lianmin Zheng}, \bibinfo{person}{Ruisi Cai}, \bibinfo{person}{Zhao Song}, \bibinfo{person}{Yuandong Tian}, \bibinfo{person}{Christopher R\'{e}}, \bibinfo{person}{Clark Barrett}, \bibinfo{person}{Zhangyang~"Atlas" Wang}, {and} \bibinfo{person}{Beidi Chen}.} \bibinfo{year}{2023}\natexlab{b}.
\newblock \showarticletitle{H2O: Heavy-Hitter Oracle for Efficient Generative Inference of Large Language Models}. In \bibinfo{booktitle}{\emph{Advances in Neural Information Processing Systems}}, \bibfield{editor}{\bibinfo{person}{A.~Oh}, \bibinfo{person}{T.~Naumann}, \bibinfo{person}{A.~Globerson}, \bibinfo{person}{K.~Saenko}, \bibinfo{person}{M.~Hardt}, {and} \bibinfo{person}{S.~Levine}} (Eds.), Vol.~\bibinfo{volume}{36}. \bibinfo{publisher}{Curran Associates, Inc.}, \bibinfo{pages}{34661--34710}.
\newblock
\urldef\tempurl%
\url{https://proceedings.neurips.cc/paper_files/paper/2023/file/6ceefa7b15572587b78ecfcebb2827f8-Paper-Conference.pdf}
\showURL{%
\tempurl}


\bibitem[Zheng et~al\mbox{.}(2024)]%
        {zheng2024sglangefficientexecutionstructured}
\bibfield{author}{\bibinfo{person}{Lianmin Zheng}, \bibinfo{person}{Liangsheng Yin}, \bibinfo{person}{Zhiqiang Xie}, \bibinfo{person}{Chuyue Sun}, \bibinfo{person}{Jeff Huang}, \bibinfo{person}{Cody~Hao Yu}, \bibinfo{person}{Shiyi Cao}, \bibinfo{person}{Christos Kozyrakis}, \bibinfo{person}{Ion Stoica}, \bibinfo{person}{Joseph~E. Gonzalez}, \bibinfo{person}{Clark Barrett}, {and} \bibinfo{person}{Ying Sheng}.} \bibinfo{year}{2024}\natexlab{}.
\newblock \bibinfo{title}{SGLang: Efficient Execution of Structured Language Model Programs}.
\newblock
\newblock
\showeprint[arxiv]{2312.07104}~[cs.AI]
\urldef\tempurl%
\url{https://arxiv.org/abs/2312.07104}
\showURL{%
\tempurl}


\bibitem[Zhong et~al\mbox{.}(2024)]%
        {distserve2024}
\bibfield{author}{\bibinfo{person}{Yinmin Zhong}, \bibinfo{person}{Shengyu Liu}, \bibinfo{person}{Junda Chen}, \bibinfo{person}{Jianbo Hu}, \bibinfo{person}{Yibo Zhu}, \bibinfo{person}{Xuanzhe Liu}, \bibinfo{person}{Xin Jin}, {and} \bibinfo{person}{Hao Zhang}.} \bibinfo{year}{2024}\natexlab{}.
\newblock \showarticletitle{{DistServe}: Disaggregating Prefill and Decoding for Goodput-optimized Large Language Model Serving}. In \bibinfo{booktitle}{\emph{18th USENIX Symposium on Operating Systems Design and Implementation (OSDI 24)}}. \bibinfo{publisher}{USENIX Association}, \bibinfo{address}{Santa Clara, CA}, \bibinfo{pages}{193--210}.
\newblock
\showISBNx{978-1-939133-40-3}
\urldef\tempurl%
\url{https://www.usenix.org/conference/osdi24/presentation/zhong-yinmin}
\showURL{%
\tempurl}


\bibitem[Zhu et~al\mbox{.}(2024)]%
        {zhu2024nanoflowoptimallargelanguage}
\bibfield{author}{\bibinfo{person}{Kan Zhu}, \bibinfo{person}{Yilong Zhao}, \bibinfo{person}{Liangyu Zhao}, \bibinfo{person}{Gefei Zuo}, \bibinfo{person}{Yile Gu}, \bibinfo{person}{Dedong Xie}, \bibinfo{person}{Yufei Gao}, \bibinfo{person}{Qinyu Xu}, \bibinfo{person}{Tian Tang}, \bibinfo{person}{Zihao Ye}, \bibinfo{person}{Keisuke Kamahori}, \bibinfo{person}{Chien-Yu Lin}, \bibinfo{person}{Stephanie Wang}, \bibinfo{person}{Arvind Krishnamurthy}, {and} \bibinfo{person}{Baris Kasikci}.} \bibinfo{year}{2024}\natexlab{}.
\newblock \bibinfo{title}{NanoFlow: Towards Optimal Large Language Model Serving Throughput}.
\newblock
\newblock
\showeprint[arxiv]{2408.12757}~[cs.DC]
\urldef\tempurl%
\url{https://arxiv.org/abs/2408.12757}
\showURL{%
\tempurl}


\end{thebibliography}

\clearpage
\appendix
\section{Artifact Appendix}

\subsection{Abstract}

\sysname is a simpler, portable and performant alternative to \pa for dynamic memory management in LLM serving systems. This artifact contains instructions to install \sysname and scripts to reproduce key results of our paper (Figures 2, 3, 4, 6, 7, 8, 9, 11). The scripts are configured to run three large language models: \yismall, \llamasmall and \yimedium. 

\subsection{Artifact check-list (meta-information)}

{\small
\begin{itemize}
  \item {\bf Algorithm: } Dynamic memory management for serving LLMs.
  \item {\bf Run-time environment: } CUDA 12.1, Python 3.10, PyTorch 2.3.0.
  \item {\bf Hardware: } Two NVLink connected NVIDIA A100 GPUs with 80GB memory each.
  \item {\bf Disk space required: } 200GB.
  \item {\bf Time needed to prepare workflow: } 1 hour.
  \item {\bf Time needed to complete experiments: } 1-2 days.
  \item {\bf Publicly available?: } Yes
  \item {\bf Archived  DOI?: } https://doi.org/10.5281/zenodo.14048692
\end{itemize}
}

\subsection{Description}

\subsubsection{How to access}

Clone the artifact as follows:

\begin{Verbatim}[frame=single,rulecolor=\color{black},fontfamily=zi4,fontsize=\small]
$ git clone https://github.com/microsoft/vattention
\end{Verbatim}

\subsubsection{Hardware dependencies}
Running \yismall requires one NVIDIA A100 GPU with 80GB memory while the other two models require two NVLink-connected A100 GPUs with 80GB memory each.

\subsubsection{Software dependencies}
Requires PyTorch 2.3.0 and CUDA 12.1 (later CUDA versions may or may not work).

\subsubsection{Expected runtime} Figures 2, 3, and 11 should take a few minutes each. Figures 4, 7 should take about 1 hour each. Figure 6 requires approximately 3 hours. Figures 8 and 9 may consume more than 12 hours each.

\subsection{Installation}

Move to the directory containing artifact scripts, create and activate a conda environment, then install the artifact as follows:
\begin{Verbatim}[frame=single,rulecolor=\color{black},fontfamily=zi4,fontsize=\small]
$ cd vattention/scripts/artifact_asplos25/
$ conda create -n vattn python=3.10
$ conda activate vattn
$ (vattn) ./install.sh
\end{Verbatim}

 \verb|vattention/scripts/artifact_asplos25/README.md| file provides detailed installation instructions.
\subsection{Alternate Setup: Docker Image}

We also provide a docker image for \sysname with all its dependencies pre-installed. To access the docker image, you need to have \href{https://docs.docker.com/engine/installation/}{Docker} and \href{https://github.com/NVIDIA/nvidia-docker/}{NVIDIA Docker} installed on your system. You can then launch the docker container and navigate to the folder containing \sysname artifact, as follows:
\begin{Verbatim}[frame=single,rulecolor=\color{black},fontfamily=zi4,fontsize=\small,xrightmargin=-5mm]
$ docker run --gpus all -it \
  -p 8181:8181 --rm --ipc=host --cap-add=SYS_ADMIN \
  rnp1910/vattention:asplos_25_pytorch_run
$ cd /workspace/vattention/scripts/artifact_asplos25  
\end{Verbatim}

Then follow the experiment workflow detailed in \ref{workflow}

\subsection{Accessing \llamasmall}

Accessing \llamasmall requires logging into huggingface with the user's private token (\verb|HF_TOKEN| below). Login as follows before any experiment:
\begin{Verbatim}[frame=single,rulecolor=\color{black},fontfamily=zi4,fontsize=\small]
$ (vattn) huggingface-cli login --token HF_TOKEN
\end{Verbatim}

\subsection{Experiment workflow} \label{workflow}
You can launch all experiments at once or individually as:

\begin{Verbatim}[frame=single,rulecolor=\color{black},fontfamily=zi4,fontsize=\small]
$ (vattn) ./run_all.sh
          OR
$ (vattn) ./run_figure_2.sh
$ (vattn) ./run_figure_3.sh 
\end{Verbatim}

\subsection{Evaluation and expected results}
The raw output logs and the final plots will be redirected to \verb|./logs|.  and \verb|./plots/| subdirectories within the main artifact directory \verb|vattention/scripts/artifact_asplos25/|. The plots are in the same format as the paper and can be compared directly. The expected result is that non-paged or \sysname based configurations would perform better than the \pa counterparts, especially at longer context lengths. However, the exact numbers may differ from the paper depending on GPUs and software libraries installed on the experiment system.

\subsection{Experiment customization}
Reproducing Figures 8 and 9 of the paper requires running a large number of requests and hence these experiments can take more than 24 hours to complete. To reduce the experimental time, we have configured their scripts to run with a smaller number of requests (100 each) by default. If you want to run the full trace, execute these scripts with the \verb|--full| argument, i.e., \verb|./run_figure_8.sh --full| and \verb|./run_figure_9.sh --full|. If two NVLink connected GPUs aren't available, update run scripts to avoid running \llamasmall and \yimedium. You can also configure \llamasmall to run on a single GPU by setting \verb|--model_tensor_parallel_degree| to 1 in the \verb|./helpers/common.sh| file. Note that such changes would likely impact the absolute performance numbers considerably but \sysname is still expected to perform better than \pa.

\end{document}